\documentclass[lettersize,journal]{IEEEtran}
\usepackage{amsmath,amsfonts}
\usepackage{algorithmic}
\usepackage{array}
\usepackage[font=normalsize]{subfig}
\usepackage{textcomp}
\usepackage{stfloats}
\usepackage{url}
\usepackage{verbatim}
\usepackage{graphicx}

\usepackage{graphics} 
\usepackage{epsfig}   
\usepackage{mathptmx} 
\usepackage{times}    
\usepackage{amssymb}  
\usepackage{mathtools}
\usepackage{verbatim}
\usepackage{cite}
\usepackage{arydshln}
\usepackage{booktabs}
\usepackage[ruled,linesnumbered]{algorithm2e}
\usepackage{tikz}
\usepackage{scalefnt}
\usetikzlibrary{positioning,shapes}
\SetKwComment{Comment}{}{}

\newtheorem{Def}{Definition}

\DeclareMathAlphabet\mathcal{OMS}{cmsy}{m}{n}
\SetMathAlphabet\mathcal{bold}{OMS}{cmsy}{b}{n}

\usepackage{longtable}
\newsavebox\ltmcbox
\usepackage{booktabs}
\usepackage{tabularx}
\usepackage{multirow}
\usepackage{bm}
\usepackage{caption}

\usepackage{pgf,tikz,pgfplots}
\pgfplotsset{compat=1.15}
\usepackage{mathrsfs}
\usetikzlibrary{arrows}
\usetikzlibrary{decorations.markings}

\usepackage{wasysym}

\newcommand{\plucker}{Pl$\mathrm{\ddot{u}}$cker }

\makeatletter
\let\NAT@parse\undefined
\makeatother
\usepackage{hyperref}
\usepackage[square,numbers]{natbib}

\usepackage{amsmath,amssymb,graphicx}
\renewcommand{\parallel}{\mathrel{/\mkern-5mu/}}
\makeatletter
\newcommand{\notparallel}{%
  \mathrel{\mathpalette\not@parallel\relax}%
}
\newcommand{\not@parallel}[2]{%
  \ooalign{\reflectbox{$\m@th#1\smallsetminus$}\cr\hfil$\m@th#1\parallel$\cr}%
}
\makeatother

\def\BibTeX{{\rm B\kern-.05em{\sc i\kern-.025em b}\kern-.08em
    T\kern-.1667em\lower.7ex\hbox{E}\kern-.125emX}}
\usepackage{balance}
\begin{document}
\title{A Fingertip Sensor and Algorithms for Pre-touch Distance Ranging and Material Detection in Robotic Grasping}
\author{Cheng~Fang, \emph{Student Member, IEEE}, Di~Wang, \emph{Student Member, IEEE}, Fengzhi~Guo, \emph{Student Member, IEEE},  Jun~Zou, \emph{Senior Member, IEEE}, and Dezhen~Song, \emph{Senior Member, IEEE}
\thanks{This research is supported in part by National Science Foundation under NRI-1925037 and Amazon Research Award 2020.}
\thanks{C. Fang and J. Zou are with the Department of Electrical and Computer Engineering Department, Texas A\&M University, College Station, TX 77843, USA, Email: \texttt{junzou@tamu.edu}.}
\thanks{D. Wang, F. Guo, and D. Song are with the Department of Computer Science and Engineering of Texas A\&M University, College Station, TX 77843, USA, Email: \texttt{dzsong@cs.tamu.edu}.}
\thanks{C. Fang and D. Wang are co-first authors of this paper. Co-corresponding authors: D. Song and J. Zou.}
}


\maketitle
\begin{abstract}
To enhance robotic grasping capabilities, we are developing new contactless fingertip sensors to measure distance in close proximity and simultaneously detect the type of material and the interior structure. These sensors are referred to as pre-touch dual-modal and dual-mechanism (PDM$^2$) sensors, and they operate using both pulse-echo ultrasound (US) and optoacoustic (OA) modalities. We present the design of a PDM$^2$ sensor that utilizes a pulsed laser beam and a customized  ultrasound transceiver with a wide acoustic bandwidth for ranging and sensing. Both US and OA signals are collected simultaneously, triggered by the same laser pulse. To validate our design, we have fabricated a prototype of the PDM$^2$ sensor and integrated it into an object scanning system. We have also developed algorithms to enable the sensor, including time-of-flight (ToF) auto estimation, ranging rectification, sensor and system calibration, distance ranging, material/structure detection, and object contour detection and reconstruction. The experimental results demonstrate that the new PDM$^2$ sensor and its algorithms effectively enable the object scanning system to achieve satisfactory ranging and contour reconstruction performances, along with satisfying material/structure detection capabilities. In conclusion, the PDM$^2$ sensor offers a practical and powerful solution to improve grasping of unknown objects with the robotic gripper by providing advanced perception capabilities.

\vspace{.1in}
\emph{Note-to-Practitioner} -- The grasping of unknown objects poses a significant challenge for robotic material handling. The presence of unfamiliar materials and unknown interior structures makes it difficult to accurately determine factors such as friction coefficient or contact force. To overcome this challenge, we have developed a novel PDM$^2$ sensor capable of providing essential information about these objects before physical contact occurs. By supplying this information in advance, the sensor enables the planning algorithm to generate more informed and optimized grasping strategies, thereby enhancing the success rate of grasping tasks. We envision that the integration of this sensor into robotic systems can substantially broaden the application domain of robots, extending beyond the familiar factory floor to the unexplored service market. This expansion is possible because the PDM$^2$ sensor empowers robots to effectively handle previously unknown objects, thus increasing their versatility and adaptability in real-world scenarios.
\end{abstract}

\begin{IEEEkeywords}
Near-distance ranging, material detection, pre-touch dual-modal and dual-mechanism sensing, robotic grasping
\end{IEEEkeywords}

\section{Introduction and Related Works}

\IEEEPARstart{R}OBUSTLY grasping unknown objects poses a significant challenge in the field of robotics \cite{mason2001mechanics,ciocarlie2014towards}. As robots transition from industrial floors to a wide range of domestic applications, prior knowledge of the objects with which they interact is often unavailable, rendering sensorless grasping methods ineffective \cite{goldberg1993orienting,erdmann1988exploration}. Therefore, sensor-based approaches have emerged as a potential solution, capable of providing crucial information about an object's relative pose, as well as its material type and interior structure.  Ideally, through the integration of near-distance (e.g., $< 1.0$ cm) ranging capabilities, robotic fingers can dynamically adjust their grasping strategy in response to subtle changes in the object's pose just before contact. By incorporating this real-time ranging information, the motion planner can anticipate force distribution, impact characteristics, and friction coefficients, facilitating the formulation of a more robust grasping plan.  Furthermore, the sensors employed should also be able to provide detailed information on the material type and interior structure. These additional data enables the motion planner to make more accurate predictions regarding the object's properties, aiding in the anticipation of force distributions and the determination of optimal grasping strategies.

Regrettably, current sensor technologies face significant challenges in meeting these requirements to a large extent. For example, cameras and laser range finders face occlusion problems caused by robotic grippers themselves \cite{smith1996vision}, or may have a blind zone at close proximity \cite{wehr1999airborne,lu2020sharing,amann2001laser,stelzer2008precise}. Tactile sensors \cite{howe1993tactile,romano2011human} and force sensors \cite{xu2015design} require physical contact, which can potentially lead to the shift or damage of the target, resulting in slow or ill-posed grasping results. Although there have been recent advancements in the development of contactless proximity/pre-touch sensors based on optical, electric field, and acoustic signals, these sensors still face limitations in their sensing modalities and their compatibility with various target materials. For example, electric field sensors often encounter challenges when dealing with targets that possess dielectric constants similar to those of air \cite{smith2007electric,wistort2008electric,mayton2010electric,mayton2009electric}. Optical sensors lack sufficient lateral resolution and struggle to handle optically transparent or highly reflective targets \cite{hsiao2009reactive,yang2017pre,hasegawa2010development,maldonado2012improving}. Acoustic sensors fail when faced with porous or sound-absorbing materials \cite{jiang2013unified,guglielmelli1993avoiding,jiang2012seashell,jiang2012pretouch}.

\begin{figure}[!htbp]
    \centering
    \subfloat[]{\includegraphics[height=1.26in]{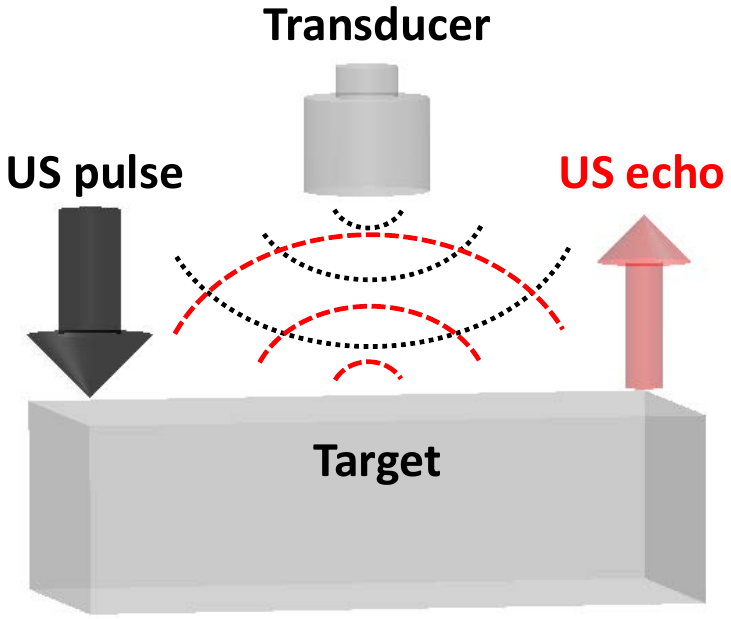}\label{fig:sensing-mechanisms-US}} 
    \hspace{.1in}
    \subfloat[]{\includegraphics[height=1.26in]{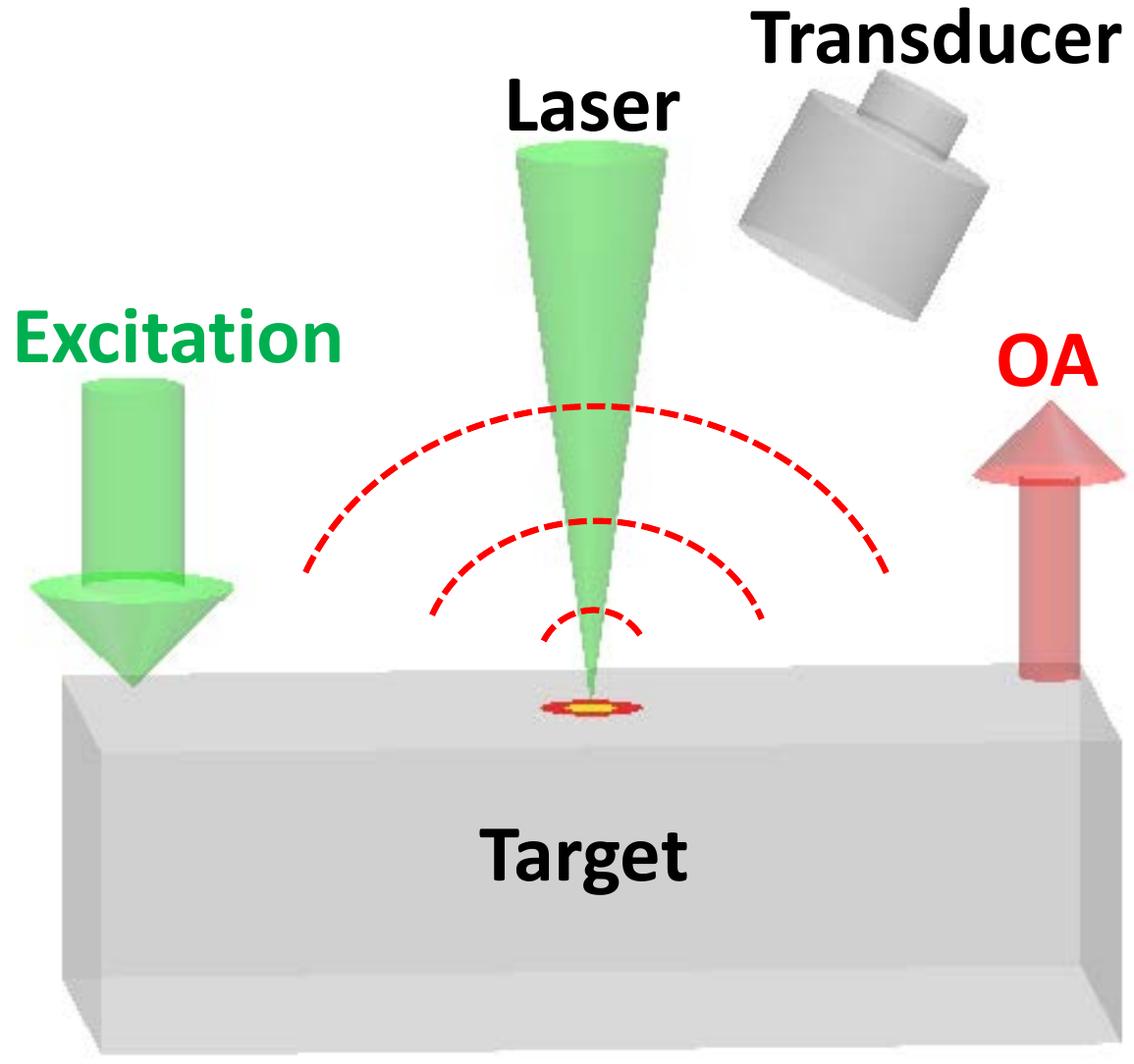}\label{fig:sensing-mechanisms-OA}}
    \caption{Conceptual illustrations of the two modalities and sensing mechanisms: (a) pulse-echo ultrasound (US), (b) laser and induced optoacoustics (OA).} 
\end{figure}

To address these challenges, we designed and designed a pre-touch sensing setup mounted on the fingertip, which incorporates both pulse-echo ultrasound (US) and optoacoustic (OA) modalities for distance ranging and material/structure sensing \cite{fang2021fingertip}. This sensor configuration is referred to as the pre-touch dual-modal and dual sensing mechanisms (PDM$^2$) design. 
The US modality (Fig. \ref{fig:sensing-mechanisms-US}) relies on the reception of acoustic signals reflected by the target, transmitted by an ultrasound transducer. These signals provide information about the distance to the target and its material/structure properties. The OA modality (Fig. \ref{fig:sensing-mechanisms-OA}) is based on the direct generation of acoustic signals within the target due to incident laser pulses. The acoustic wave is induced when a short laser pulse is absorbed by the target, resulting in a temperature pulse and subsequent mechanical expansion and contraction of the target material. In both modalities, the temporal delay of the received signals is utilized to determine the distance between the target and the sensor, enabling accurate ranging. Furthermore, the frequency spectra of the US and OA signals are used to extract distinctive features to differentiate target materials or interior structures \cite{fang2019toward,fang2020fingertip}.

\begin{figure}[!t]
\centering
\includegraphics[width=2.5in]{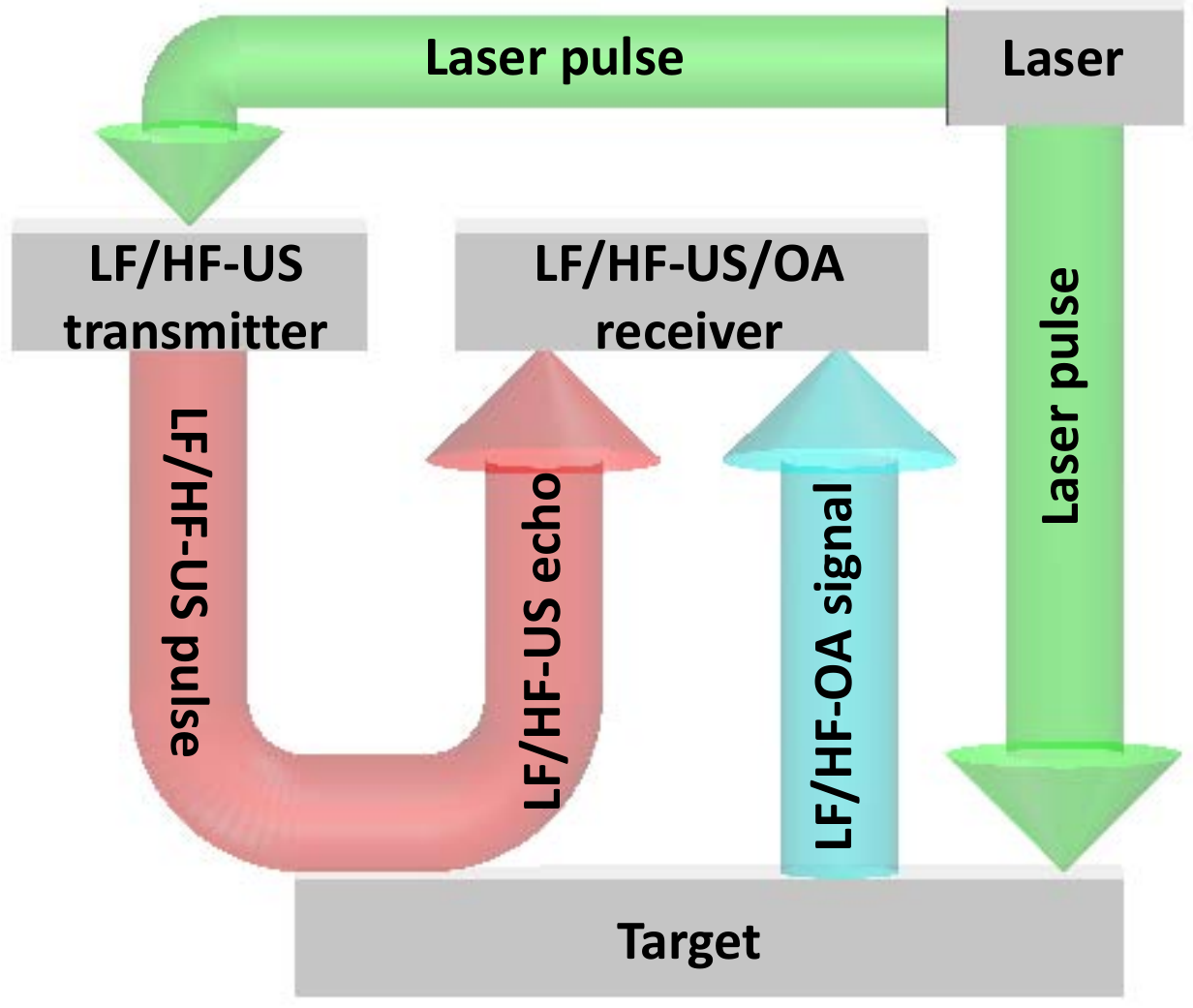}
\caption{The flow chart showing the working principles of the new PDM$^2$ sensor. HF: high-frequency; LF: low-frequency; US: ultrasound; OA: optoacoustic.}
\label{fig:working-principle}
\end{figure}

This paper makes a threefold contribution. First, we introduce a novel PDM$^2$ sensor that utilizes a single ultrasound transmitter and receiver with a wide acoustic bandwidth to achieve the transmission and reception of low- and high-frequency US) and OA signals (Fig. \ref{fig:working-principle}). With this configuration, both range and sensing capabilities are achieved using a single ultrasound transmitter and receiver. Moreover, the collection of both US and OA signals is triggered by the same laser pulse, which is delivered through an optical fiber. This design choice simplifies system construction and operation, making it straightforward and user-friendly. Second, we have developed signal processing algorithms for the sensor, which serve two main purposes: 1) the automatic estimation of time-of-flight (ToF), and 2) ranging rectification to compensate for nonlinearity within the system. Lastly, we have designed an object scanning system based on the PDM$^2$ sensor and proposed a calibration algorithm that addresses the issue of sensor installation. This algorithm improves the accuracy of ranging and material detection by facilitating sensor installation in the application.

To validate our design, we have manufactured a prototype PDM$^2$ sensor and constructed an object scanning system. We applied our algorithms, and the results obtained have demonstrated satisfactory performance in both ranging and material/structure sensing. The maximum deviation in ranging measurements is less than 0.2 mm, indicating high accuracy. In terms of material detection, the sensor successfully differentiates all eight types of objects that present optical and acoustic challenges. In conclusion, the newly developed PDM$^2$ sensor offers a practical and powerful solution to assist with robotic grasping of unknown objects. Its successful validation through the prototype and object-scanning system confirms its potential for real-world applications.

This paper builds on two earlier conference papers~\cite{fang2022second,wang2022design} by adding substantial new developments. We present new signal processing algorithms and a calibration method for general mounting platforms. We also investigate whether our PDM$^2$ sensor is sensitive to surface normal deviations in experiments in addition to the new experiments to validate ranging algorithms.

The remainder of the paper is structured as follows. In Sec.~\ref{sec:sensor_design}, we present the design of the new PDM$^2$ sensor and its scanning system. Sect.~\ref{sec:alg} delves into the algorithm design, specifically addressing automatic estimation of time-of-flight and ranging rectification issues. The design of the object scanning system, along with its calibration algorithm, is discussed in Sec.~\ref{sec:scanning}. Sec.~\ref{sec:exp} outlines the experimental design and presents the corresponding results. Finally, we conclude the paper in Sec.~\ref{sec:conlusion}.

\section{Sensor and Scanning System}\label{sec:sensor_design}

We have designed and fabricated a new type of sensor with a new mounting platform to investigate how to calibrate and test the new sensor. We begin with the new sensor design. 

\subsection{Sensor design and testing}

\begin{figure}[!t]
\centering
\includegraphics[width=2.5in]{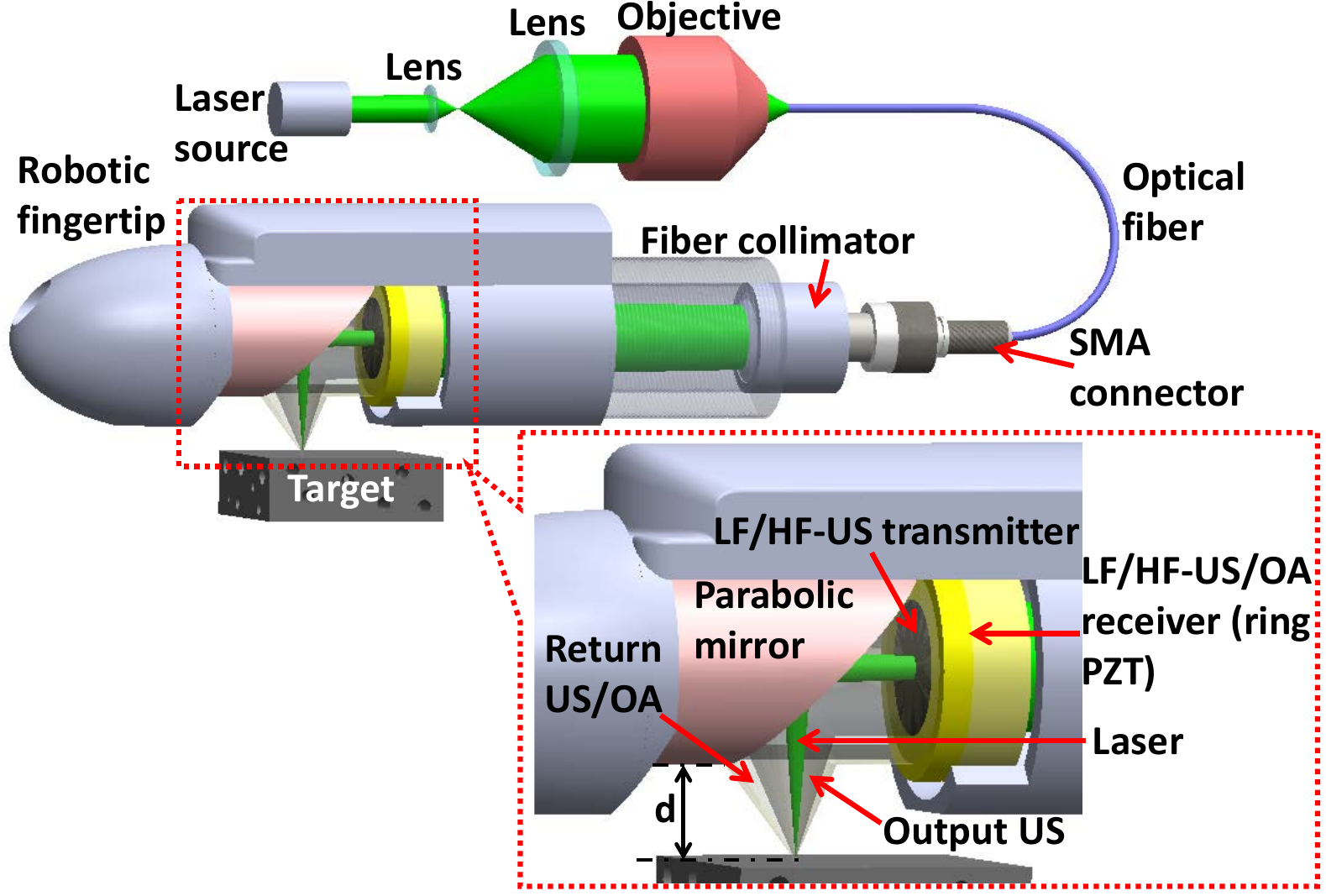}
\caption{Schematic design of the new PDM$^2$ sensor. HF: high-frequency; LF: low-frequency; US: ultrasound; OA: optoacoustic.}
\label{fig:schematic-design}
\end{figure}

The schematic design of the new PDM$^2$ sensor is shown in Fig. \ref{fig:schematic-design}. It comprises a single (optoacoustic) ultrasound transmitter and receiver arranged in a co-centered and co-axial configuration. With the laser pulses delivered through the optical fiber, the ultrasound transmitter emits both low- and high-frequency ultrasound pulses towards the target. The echo signals are then captured by the ring-shaped transducer. For ultrasound (US) ranging, the time-of-flights (ToFs) of the echo signals are utilized, while the frequency spectra of the echoes are employed for US sensing. 
Regarding optoacoustic (OA) ranging and sensing, the central region of the laser pulses passes through the center hole of the optoacoustic ultrasound transmitter, reaching the target surface via parabolic mirror reflection. The resulting OA signals are also received by the ring-shaped transducer. OA distance ranging is accomplished by analyzing the ToFs of the signals, while material/structure sensing is performed by examining their frequency spectra. 
In particular, the same ultrasound receiver is responsible for collecting both US and OA signals, all triggered by a single laser pulse. This streamlined approach simplifies sensor operation and data acquisition. Importantly, the single triggering scheme ensures that US and OA signals remain separate. This is due to the fact that the US signal undergoes a round trip (transmitter-target-receiver) with a ToF twice as long as that of the OA signal, which completes a single trip (target-receiver). Consequently, the temporal disparity between the two signals is significantly greater than their individual durations.

The proposed PDM$^2$ sensor represents a notable advancement compared to our initial prototype \cite{fang2021fingertip}. Our new PDM$^2$ sensor design emphasizes simplicity. The first-generation prototype consisted of two ultrasound transmitters, driven by a laser pulse and a pulser-receiver, along with two ultrasound receivers: a microphone and a lead zirconate titanate (PZT) ring, which were designed to handle low- and high-frequency ultrasound signals, respectively. Additionally, these two ultrasound receivers were used to capture low- and high-frequency optoacoustic (OA) signals. However, the use of multiple transmitters and receivers in system design contributes to higher complexity in both construction and operation. Consequently, this complexity limits the practical applications of the system in real grasping scenarios.

\begin{figure}[!t]
\centering
\includegraphics[width=2.5in]{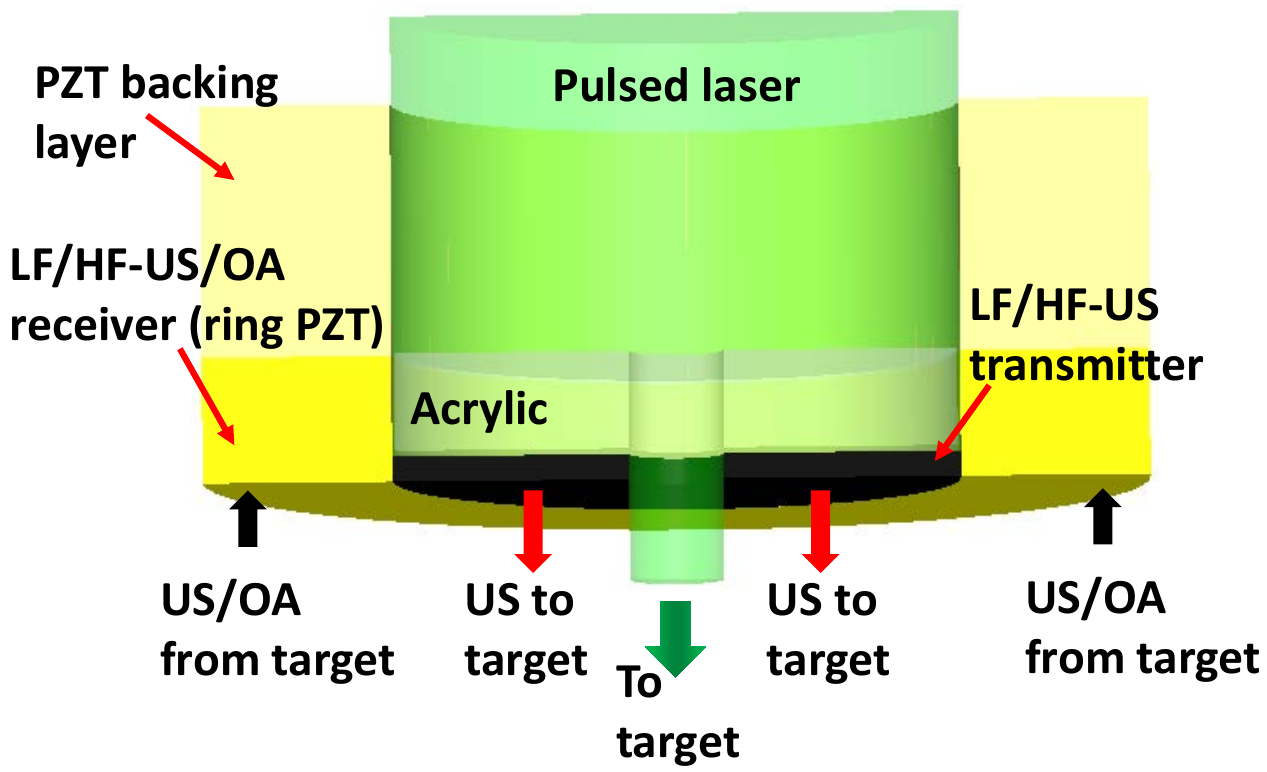}
\caption{A zoom-in diagram of the cross-section of the designed optoacoustic ultrasound transmitter integrated with the ring PZT transducer under pulsed laser illumination, which form a co-centered and co-axial arrangement. HF: high-frequency; LF: low-frequency; US: ultrasound; OA: optoacoustic.}
\label{fig:transmitter-with-PZT}
\end{figure}

To generate low- and high-frequency ultrasound signals for US distance ranging and material/structure sensing, a single optoacoustic ultrasound transmitter is used. To overcome this challenge, the optoacoustic ultrasound transmitter (Fig. \ref{fig:transmitter-with-PZT}) is designed with an optically transparent acrylic substrate as a supportive base and a layer of black vinyl electrical tape as the laser-absorption material. Both acrylic and vinyl possess low Young's modulus and high internal damping characteristics, enabling the effective generation of wideband ultrasound signals encompassing low-frequency (kHz) flexural and high-frequency (MHz) thickness modes. The acrylic substrate has dimensions of 1.6 mm in thickness and 9 mm in diameter, with a central hole measuring 1.5 mm in diameter to facilitate the passage of laser pulses for OA ranging and sensing. 
Similarly, the detection of both low- and high-frequency US/OA signals for ranging and sensing purposes is accomplished using a single-ring PZT transducer. To address this challenge, the ring PZT transducer (Fig. \ref{fig:transmitter-with-PZT}) is designed with a thick backing layer to dampen acoustic resonance and a large inner diameter to induce multiple modes of vibration, such as radial (kHz) and thickness (MHz) modes. This design enables the ring PZT transducer to possess a wide bandwidth that matches that of the optoacoustic ultrasound transmitter.

\begin{figure}[!htbp]
    \centering
    \subfloat[]{\includegraphics[width=3in]{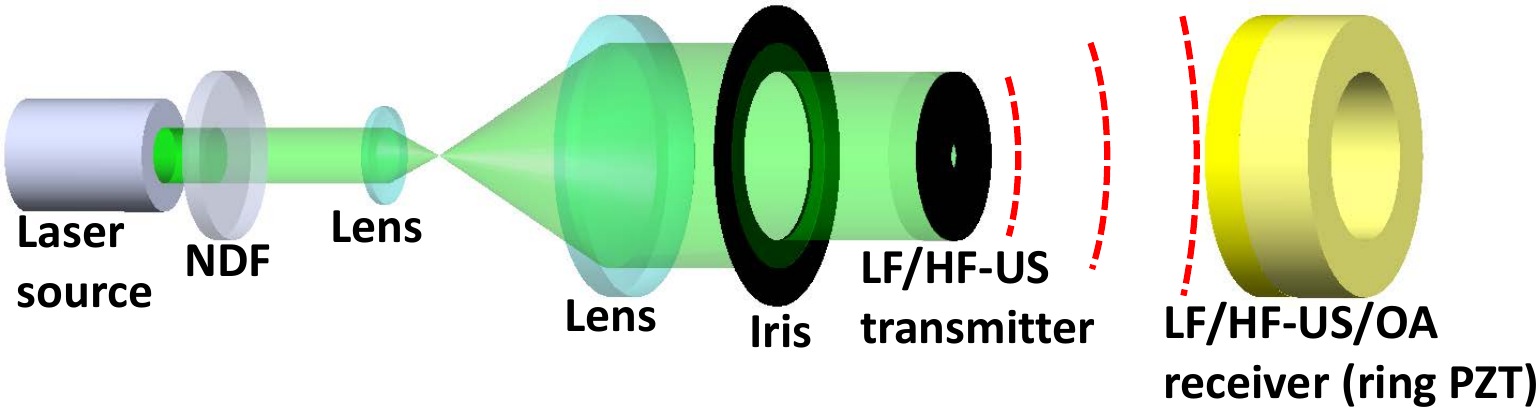}\label{fig:sensor-setup}}\\
    \subfloat[]{\includegraphics[height=1.26in]{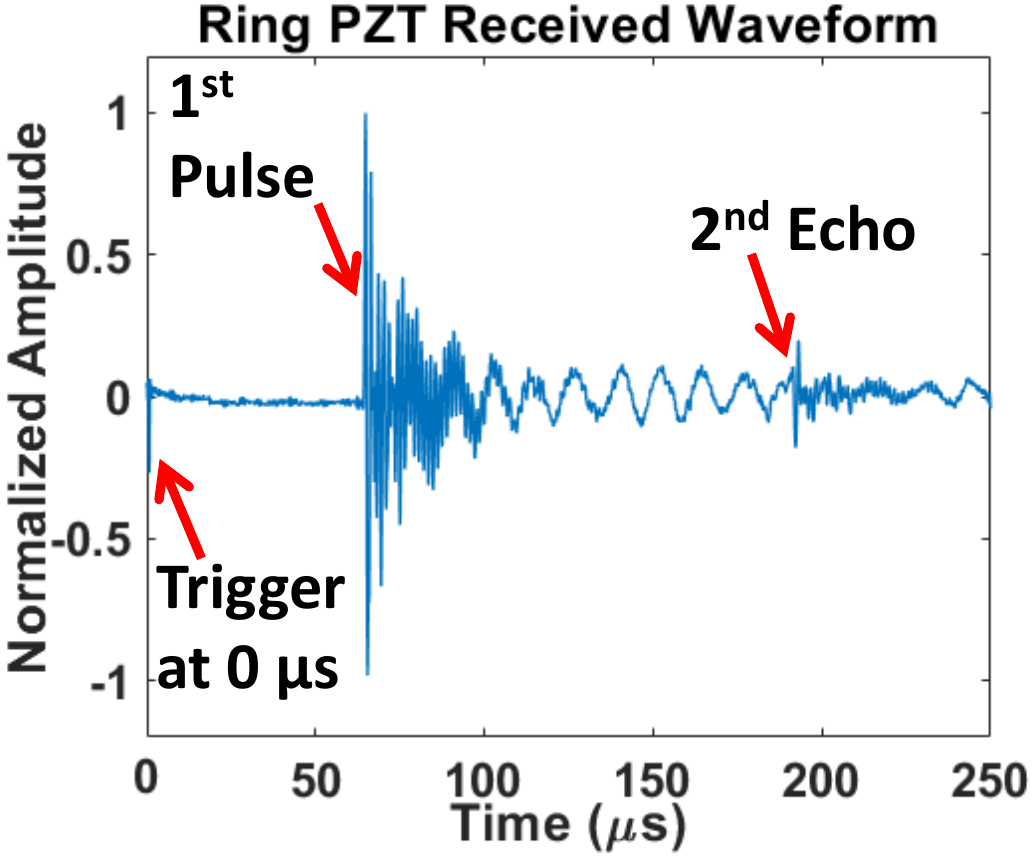}\label{fig:waveform-time-domain}} \hspace*{.1in}
    \subfloat[]{\includegraphics[height=1.26in]{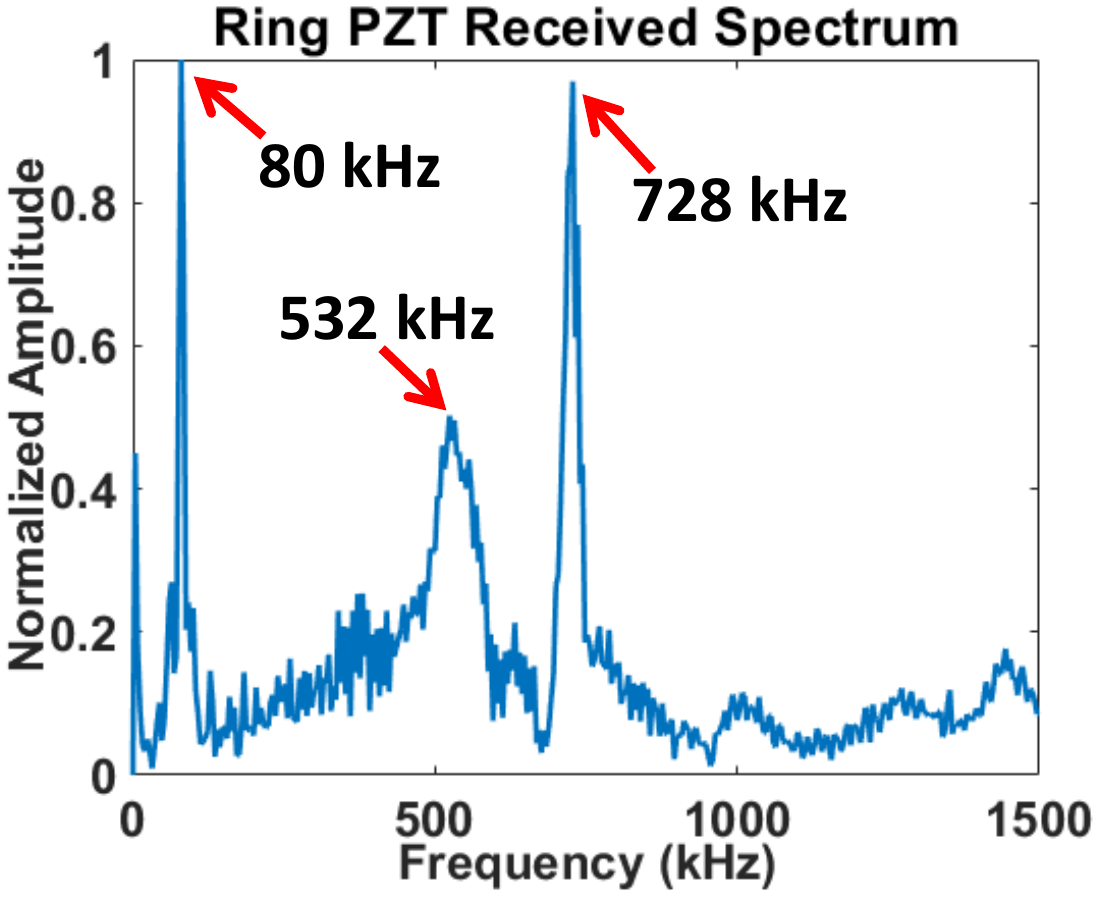}\label{fig:waveform-frequency-domain}}
    \caption{(a) Diagram of the setup in air to characterize the collective bandwidth of the optoacoustic ultrasound transmitter and ring PZT transducer in new PDM$^2$ sensor. NDF: neutral density filter; HF: high-frequency; LF: low-frequency; US: ultrasound; OA: optoacoustic. Representative (b) waveform and (c) frequency spectrum of the received ultrasound signals. }
    \label{sensor-setup-eg-waveform}
\end{figure}

The collective bandwidth of the optoacoustic ultrasound transmitter and the ring PZT transducer was characterized as shown in Fig.~\ref{fig:sensor-setup}. A Q-switched 532-nm Nd:YAG pulsed laser has been utilized as the light source, operating at a repetition rate of 10 Hz with a pulse duration of 8 ns and an average pulse energy of 20 mJ/pulse. To protect the optoacoustic transmitter from potential damage, a neutral density filter (NDF) has been employed to reduce the laser pulse energy to approximately 2.5 mJ/pulse. The laser beam is expanded using two lenses and further filtered through an iris. A photodetector is employed to detect the laser pulse and generate a trigger signal for data acquisition synchronization. The received signals are amplified by a preamplifier and recorded using an oscilloscope. Figs. \ref{fig:waveform-time-domain} and \ref{fig:waveform-frequency-domain} present a representative waveform and its frequency spectrum, respectively. The time-domain waveform exhibits two distinct pulses resulting from multiple acoustic reflections between the ultrasound transmitter and the ring PZT transducer. The acoustic frequency spectrum indicates the collective bandwidth of the optoacoustic ultrasound transmitter and ring PZT transducer, with center frequencies approximately around 80 kHz, 532 kHz, and 728 kHz.

\begin{figure}[!htbp]
    \centering
    \subfloat[]{\includegraphics[width=3in]{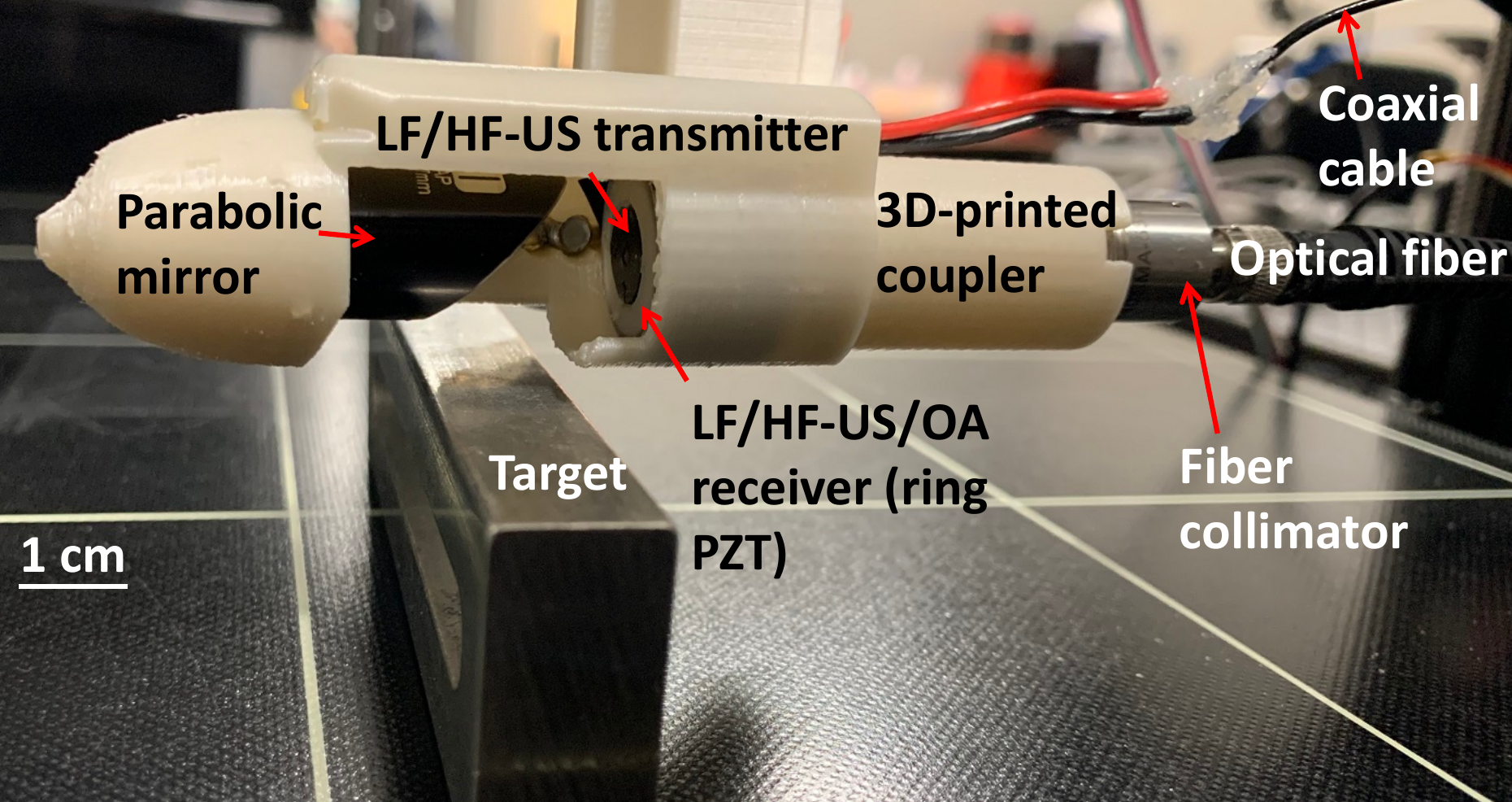}\label{fig:prototype-img}}\\
    \subfloat[]{\includegraphics[height=2in]{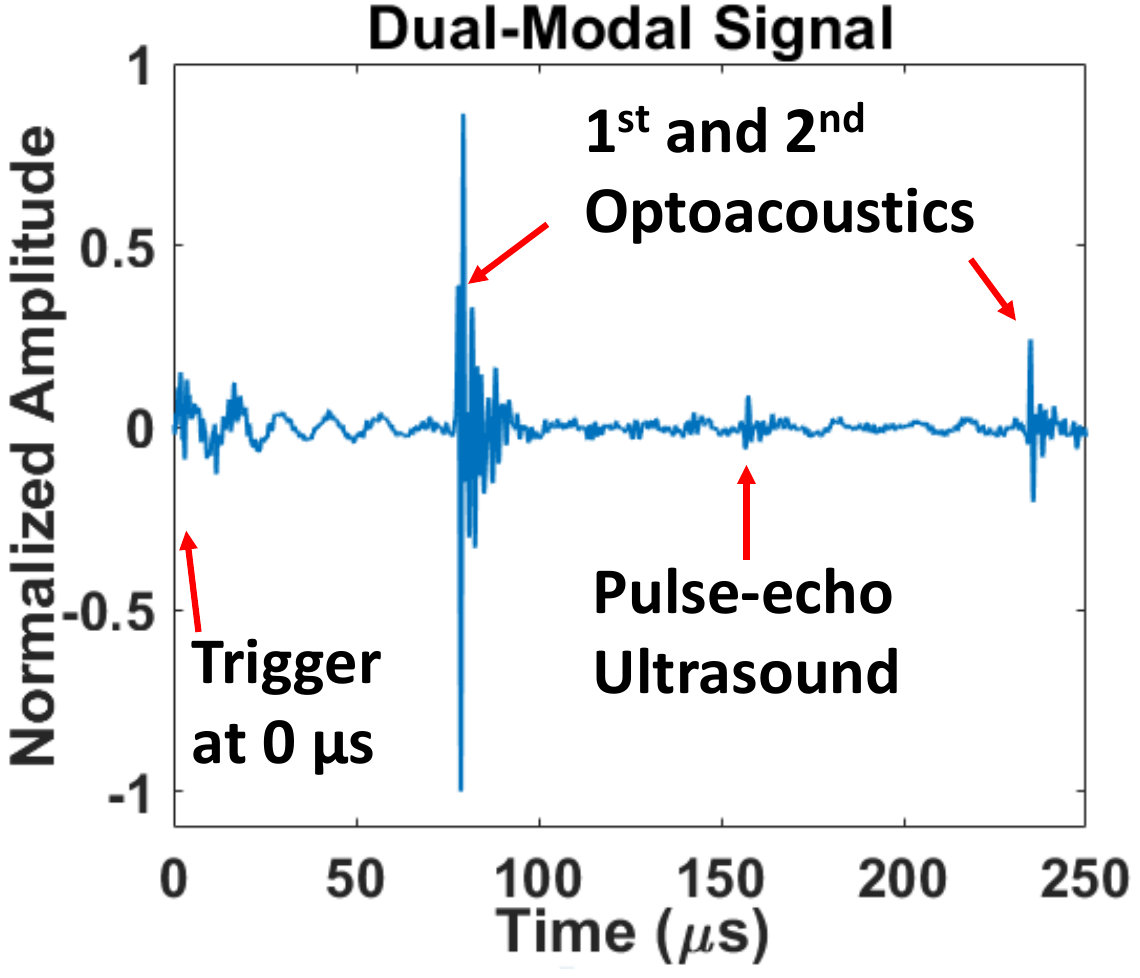}\label{fig:representative-waveform-US-OA}}
    \caption{(a) A close-up photograph of a fabricated prototype of the new PDM$^2$ sensor. HF: high-frequency; LF: low-frequency; US: ultrasound; OA: optoacoustic. (b) Representative waveform including the received US and OA signals from an aluminum block target through air. } 
    \label{fig:prototype}
\end{figure}

Fig.~\ref{fig:prototype-img} shows the fabricated prototype of the new PDM$^2$ sensor, which comprises a 3D printed housing, a 90 degree parabolic mirror, an optoacoustic ultrasound transmitter, a ring PZT transducer, a 3D printed coupler, a fiber collimator, and an optical fiber. The construction of the prototype is in agreement with the schematics shown in Fig.~\ref{fig:schematic-design}. The same pulsed laser illustrated in Fig.~\ref{fig:sensor-setup} is used, with a single laser pulse triggering simultaneous data acquisition for both US and OA signals (refer to Fig.~\ref{fig:representative-waveform-US-OA}). The time-of-flights (ToF) for the first OA signal, the US signal, and the second OA signal (echo after a round trip) measure approximately 77 µs, 154 µs, and 231 µs, respectively. These temporal separations are sufficiently long to distinguish and separate the US and OA signals from one another, facilitating ranging and material/structure sensing.

\subsection{Scanning System Design}
The new PDM$^2$ sensor is a unique device that requires a thorough study to effectively leverage its capabilities in object scanning and grasping. Consequently, we are tasked with developing a novel scanning system and implementing suitable calibration schemes. In this section, we will elaborate on the process of scanning system development, while deferring the discussion of calibration algorithm design to the dedicated algorithm section. 

As we mentioned in our early conference version~\cite{wang2022design}, the object scanning system with the new PDM$^2$ sensor serves two purposes: i) it allows us to study the algorithmic issues of deploying the new PDM$^2$ sensor and ii) we can use the system to scan many household items to establish an object/material database of common household items to facilitate recognition tasks in applications. 

\begin{figure}[h]
\centering
\subfloat[]{\includegraphics[width=1.5in]{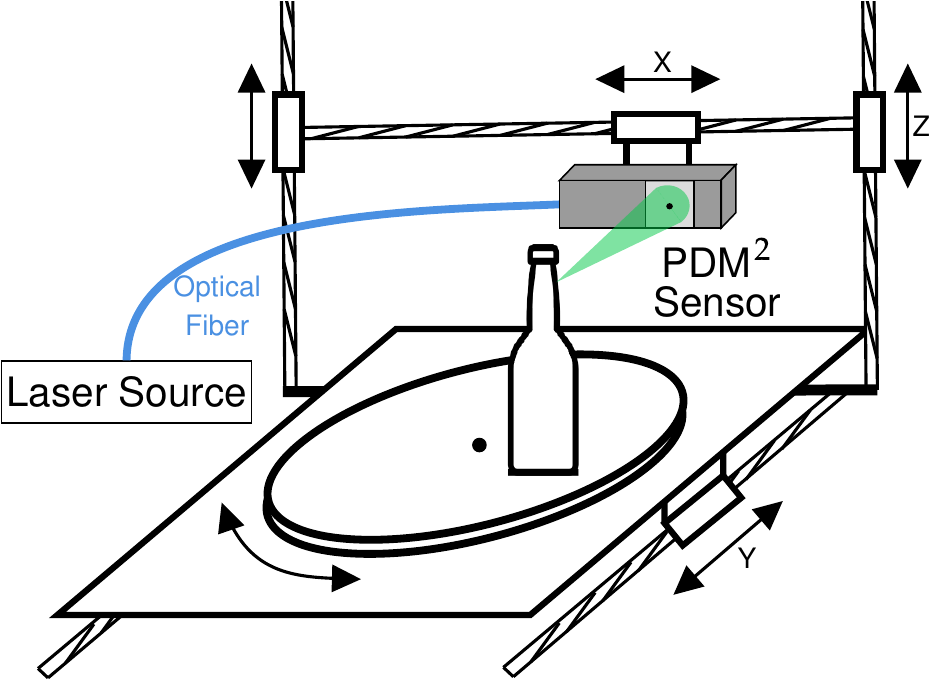}\label{fig_MD}\label{fig:scanning-system-illustration}}\hspace{0.1in}
\subfloat[]{\includegraphics[width=1.7in]{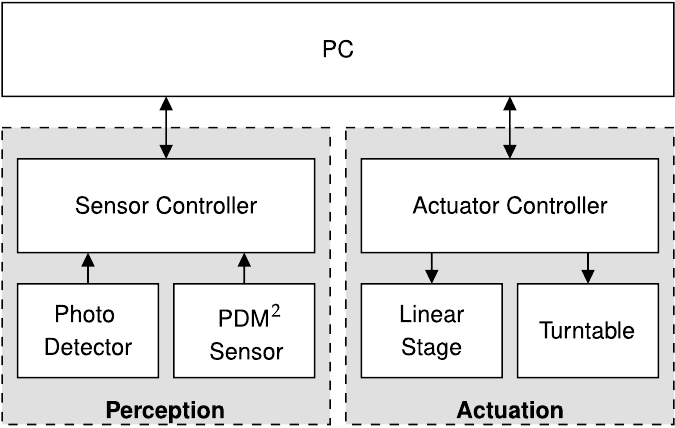}\label{fig:scanning-system-block-diagram}}
\caption{Schematic diagram of hardware components in (a)  and communications of the object scanning system in (b).}
\end{figure}

The system is composed of a refitted 3D printer (Anycubic™ Chiron) with a motorized turntable (TBVECHI™ HT03RA100) mounted on its printing stage (Fig. \ref{fig:scanning-system-illustration} and \ref{fig:scanning-system-block-diagram}). The 3D printer nozzle is replaced by the PDM$^2$ sensor to perform 3D translation, and the target of interest is supported and rotated by the turntable for distance range, material / structure detection, or full body scan. The scanning system sensor controller (STM32™ NUCLEO-H743ZI) and actuator controller (Atmel™ ATmega2560) communicate with the PC through a serial port to report the signals from the PDM$^2$ sensor and adjust the position of the nozzle and the angle of the turntable. 

\section{Signal Processing Algorithms}\label{sec:alg}

\begin{figure}[!t]
\centering
\includegraphics[width=3.5in]{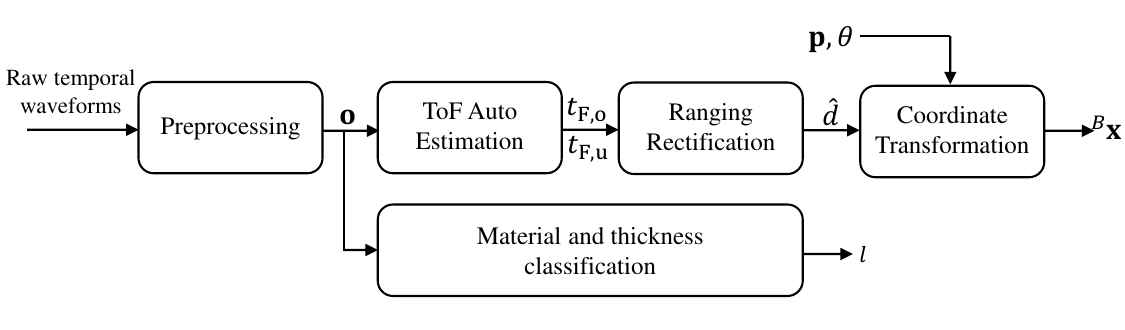}
\caption{An overview of data processing pipeline.}
\label{fig:software-system-diagram2}
\end{figure}

The overview of the signal processing pipeline is presented in Fig.~\ref{fig:software-system-diagram2}. Initially, the raw temporal waveforms undergo preprocessing, including averaging and band-pass filtering, to enhance the signal-to-noise ratio (SNR). The processed results, denoted as $\mathbf{o}$, are utilized to automatically estimate ToFs for both US and OA signals, represented as $t_{\mathrm{F,u}}$ and $t_\mathrm{F,o}$, respectively.

Following ranging rectification, the pre-touch distance, denoted as $d$, is estimated using nonlinear mapping based on ToFs. The scanning point position $\mathbf{x}$ can then be determined using $d$, in combination with the coordinate transformation of the position reading and the orientation of the sensor mounting platform $\mathbf{p}$ and $\theta$. This process allows for the reconstruction of the object contour as a point cloud using the readings obtained from the scanning points.

Concurrently, the waveforms $\mathbf{o}$ undergo processing by the bag-of-SFA symbols (BOSS) classifier~\cite{schafer2015boss, bagnall2017great} to differentiate the material or structure of the targets.

Now, we will provide detailed explanations for each building block, starting with the ToF estimation.

\subsection{Automatic estimation of Time-of-flight (ToF)}
\begin{figure}[!t]
\centering
\includegraphics[width=3.5in]{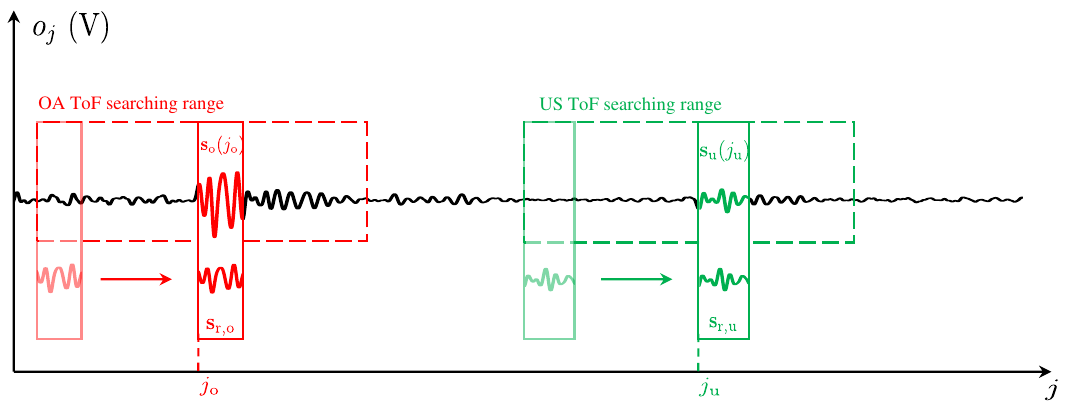}
\caption{ToF auto estimation which searches OA and US windows of signals. The search range and the sliding window are indicated by dashed and solid boxes, and the OA and US modality are colored red and green, respectively.}
\label{fig:ToF_est_theorical}
\end{figure}

ToFs of US and OA signals can be used to obtain distance ranging if provided with a known sound speed. Since a large number of raw temporal waveforms are collected and preprocessed, the ToFs are expected to be automatically extracted from each $\mathbf{o}$ for the pre-touch distance estimation $d$. Due to the relatively similar frequency spectra of the acquired US and OA signals, a cross-correlation (CC) method \cite{simkin1974measurements} is selected for the estimation of ToF, which calculates the correlation between a transmitted signal $\mathbf{s}_\mathrm{r}$ with length of n and the corresponding received signal $\mathbf{s}$ collected in a discrete system. Because the procedure of auto-estimation of ToF is almost the same for both US and OA signals, we only detail the procedure of US ToF estimation here. A special pattern $\mathbf{s}_\mathrm{r,u}$ is manually pre-selected from the historical pre-processed US waveform as the reference ($\mathbf{s}_\mathrm{r,o}$ for OA). A simplified diagram of the search process for ToF autoestimation is shown in Fig. \ref{fig:ToF_est_theorical}. The cross-correlation value $c\left(j\right)$ is defined as
\begin{equation}
    c(j) = \mathbf{s}_\mathrm{r,u}^{\mathsf{T}}\mathbf{s}(i),
\end{equation}
where $\mathbf{s}(i)$ is a sliding window through $\mathbf{o}$,
$\left\lbrack o_{j},\ldots,o_{j + n - 1} \right\rbrack^{\mathsf{T}}$, $i \in \{j_\mathrm{u,min},\ldots,j_\mathrm{u,max}\}$,
and the minimum and maximum US
time indices in any $\mathbf{o}$ is denoted by $j_\mathrm{u,min}$ and $j_\mathrm{u,max}$, respectively.  As it sliding through $\mathbf{o}$, the window with
maximum $c(j)$ is the most
correlated to the referenced $\mathbf{s}_\mathrm{r,u}$, which is identified as
the US signal in $\mathbf{o}$. Then the US ToF ($t_\mathrm{F,u}$) can be calculated by
\begin{equation}
    t_\mathrm{F,u} = \arg\max_{i}\frac{c(j)}{2f_s},
\end{equation}
where $f_{s}$ is the data acquisition sampling frequency and the half division is because the US signal travels a round trip between sensor and target. Similarly, the OA ToF
($t_\mathrm{F,o}$) is $arg\max_{i}c(j)/f_s$, because the OA
signal travels one way from the target to the sensor.

\subsection{Ranging Rectification}\label{ssc:ranging-rect-alg}
\begin{figure}[!t]
\centering
\includegraphics[width=2.5in]{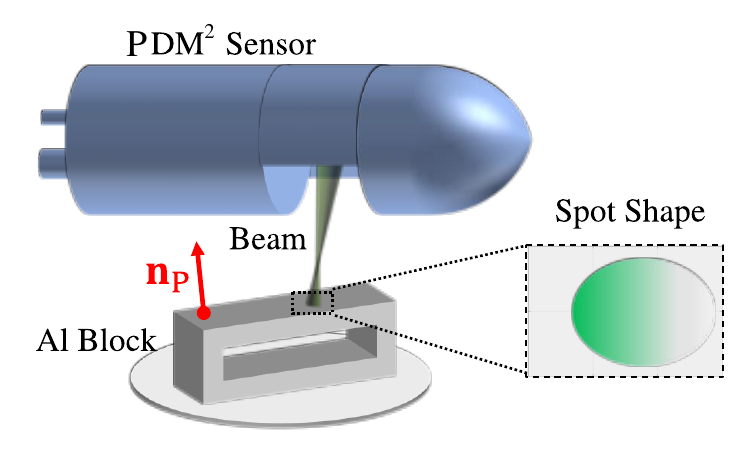}
\caption{A simplified diagram of a representative aluminum block target illuminated by the laser and ultrasound beams from the PDM$^2$ sensor, whose diameter varies along the transmission and direction deviates from the normal vector $\mathbf{n}_{p}$ of target surface.}
\label{fig:Ranging_Rectification}
\end{figure}
Ideally, the scanning spot on target should be infinitely small, and the  sensor-spot distance should be directly proportional to the acquired ToF. Unfortunately, both the laser and ultrasound beams have varied diameters along the transmission direction (Fig. \ref{fig:Ranging_Rectification}), which enlarges the spot into a target region. Consequently,  the distance of the targeted region, instead of an ideally-small spot, is indicated by the US/OA ToF, which results in ranging nonlinearity. 
To address this issue and minimize the range deviation, a second-order polynomial regression model is utilized to approximate the ToF range function, whose optimization is addressed by maximum likelihood estimation (MLE) as below.
\begin{equation}
\min_{\beta_{2},\beta_{1},\beta_{0}}{\sum_{i = 1}^{N}\left\| \beta_{2}t_{F,m,i}^{2} + \beta_{1}t_{F,m,i} + \beta_{0} - d_{i} \right\|_{\Sigma_{i}}^{2}}  
\end{equation}
where $\beta_{2},\beta_{1},\beta_{0} \in R$ are the model parameters,
$t_{F,m,i}$ represents the \emph{i}-th ToF reading of $m$ modalities in
the $N$-sample training dataset, $m \in \text{\{}o,u\text{\}}$, and
$\Sigma_{i}$ is the uncertainty of \emph{i}-th measurement. Then the
rectified ranging distance can be calculated by the following.
\begin{equation}
    \widehat{d} = \beta_{2}t_{F,m}^{2} + \beta_{1}t_{F,m} + \beta_{0}.
\end{equation}

After obtaining $\widehat{d}$, the subsequent task involves deriving the 3D position of the scanned point $\mathbf{x}$, using information from the coordinate system of the mounting platform, as illustrated in Fig.~\ref{fig:software-system-diagram2}. This information can be acquired through the calibration of the system and sensor in the sensor installation stage, which will be explained next.

\section{Sensor and System Calibration}\label{sec:scanning}

The calibration of the sensor and the scanning system is particularly important when the new PDM$^2$ sensor is integrated into a robotic manipulation platform. 

Sensor calibration is necessary because
the output laser and ultrasound beams may not be fully coincident with the desired direction, which depends on the fabrication and assembly accuracy of the sensor components. 

Scanning system calibration is also necessary because precise coordinating system transformation cannot be obtained without a proper calibration procedure considering the uniqueness of the new sensor. In our object scanning system, the direction of the axis and the center location of the turntable may not align well with the anticipated specifications. Similar issues may exist in other mounting platforms. Using our object scanning system as an example, we develop and demonstrate the sensing system calibration procedure and also show that it can be generalized to other platforms. 

\subsection{Calibration Parameters}

For the 1D scanning PDM$^2$ sensor, the key parameter to be calibrated is the direction vector of the laser and ultrasound beams in the referenced coordinates of the 3D linear stages, which is also the world reference in the object scanning system. The direction vector is denoted as $\mathbf{v}$ and is indicated by the yellow arrows in Fig.~\ref{fig:in-frame-motion-and-beam}. Because OA and US signals are codirectional and coaxial based on sensor design (Fig.~\ref{fig:schematic-design}), we only need to calibrate the $\mathbf{v}$ of the OA modality (laser beam) because the OA signal has a higher lateral resolution than the US signals.

The calibration of the scanning system materializes itself as the turntable's pose calibration in world reference frame which includes its rotation axis vector (normal to the plane surface) and its center location, which is denoted as the yellow vector $\mathbf{n}$ and the position $\mathrm{X_R}$, respectively (Fig.~\ref{fig:bw-frame-rot-platform-pose}). Later, we will explain how to expand the scanning system calibration from the turntable to a generic manipulator-based sensor mounting. The common notations used in the calibration are defined below.

\subsection{\textit{Nomenclature}}
 $\mathbb{S}^2$ is the unit 2-sphere in 3D Euclidean coordinate system, $\mathrm{T}_{\mathbf{q}}\mathbb{S}^2$ is the tangent space at point $\mathbf{q}\in\mathbb{S}^2$.

\begin{itemize}
\item[$\{\mathbf{0}\}$] represents sensor initial frame, it is a right-handed 3D Euclidean system defined by sensor initial position. Its origin is at the intersection point of the output laser beam and the outer surface of the robotic fingertip. Its $X$-, $Y$- and $Z$-axes are parallel to the $X$, $Y$, and $Z$ directions of the 3D linear stages, respectively. All variables are defaulted to $\{\mathbf{0}\}$.
\item[$\mathbf{n}$] is the normal vector of the turntable surface, $\mathbf{n} \in \mathbb{S}^2$. 
\item[$\mathbf{v}$] is the direction vector of the laser beam output of the sensor, $\mathbf{v} \in \mathbb{S}^2$.
\item[$\mathrm{S}$] is a reading of the sensor position from the linear stage, $\mathrm{S}\in \mathbb{R}^3$.
\item[$\mathrm{X}$] is a point in the 3D Euclidean space, $\mathrm{X} \in \mathbb{R}^3$.
\item[$\mathbf{E}$] is an edge in 3D Euclidean space. $\mathbf{E} = [[\mathbf{q}]_\times \textrm{  }  \mathbf{m}]$, where $\mathbf{q}\in\mathbb{S}^2$ is its unit length direction vector and $\mathbf{m}\in\mathrm{T}_{\mathbf{q}}\mathbb{S}^2$ is its moment vector in \plucker coordinate \cite{mason2001mechanics}. $[\cdot]_\times$ denotes the skew-symmetric matrix.
\end{itemize}

\subsection{Calibration Rig Design and Measurements} \label{subsec:Measurements}
A customized calibration rig is designed to facilitate  the calibration, which consists of a thin straight rigid graphite filament (0.5-mm-diameter pencil lead) mounted on a 3D-printed base. The thin graphite filament is chosen because of its small diameter and strong OA response. The rig is placed on the flat surface of the turntable. 
The OA data point perceived from the end of the horizontal filament is called the tip point (Fig.~\ref{fig:in-frame-motion-and-beam}), while that from the sidewall of the vertical filament is called the edge point (Figs.~\ref{fig:bw-frame-rot-platform-pose} and \ref{fig:cali-blowup}). Their 3D coordinates are recovered from the depth reading of the sensor $d$, the position of the sensor $\mathrm{S}$, and the direction of the beam $\mathbf{v}$. 
By scanning the $X$-$Z$ planar linear stages (in-frame motion) along the green dashed arrows in Figs.~\ref{fig:in-frame-motion-and-beam} and \ref{fig:bw-frame-rot-platform-pose}, multiple raw tip or edge points are combined into 2D frames, providing the OA amplitude heatmaps located in the bottom right black boxes.
The center points of the filament end or sidewall, which are the key input measurements for calibration, are determined by the position with the maximum amplitude in the heatmap.

The frame and center point indices are denoted by the subscripts $i$ and $j$, respectively, and $j$ is omitted when there is only one center point in a frame. For example, the center point $\mathrm{X}_{ij}$ denotes the $j$-th center point scanned in the $i$-th frame. In a tip scan, the tip center point (shown as the red dot in the bottom-right black box of Fig.~\ref{fig:in-frame-motion-and-beam}) is determined by averaging all thresholded tip points scanned using $X$-$Z$ planar motion. In an edge scan, the center points of the edge (red dots in the bottom right black box of Fig.~\ref{fig:bw-frame-rot-platform-pose}) are determined by averaging the thresholded edge points scanned using the $X$-axis motion. 
According to the central limit theorem, when a center point $\mathrm{X}_{ij}$ is computed from $n$-raw points, the noise associated with its corresponding sensor position reading $\mathrm{S}_{ij}$ in the linear stage can be described by a Gaussian distribution $\mathcal{N}(0,\Sigma_{ij})$, where $\Sigma_{ij} = \frac{1}{n}\Sigma_{\mathrm{P}ij}$ is the noise covariance matrix and $\Sigma_{\mathrm{P}ij}$ is the covariance matrix of the sensor position readings of the raw points. Noise in the depth measurement of the sensor $d_{ij}$ of a center point follows a similar derivation and has a variance $\sigma_{ij} = \frac{1}{n}\sigma_{\mathrm{P}ij}$, where $\sigma_{\mathrm{P}ij}$ is the variance of distance readings of the raw points.


Multiple frames of the rig at different distances or perspectives are captured by moving the sensor with the linear stages and rotating the rig with the turntable. The motion in Y axis is called between-frame motion (shown as the black solid arrows in Figs.~\ref{fig:in-frame-motion-and-beam} and \ref{fig:bw-frame-rot-platform-pose}), where the rotary movement  between the $i$-th and $k$-th frames is determined by the angular readings $\theta_{ik}$, the normal vector $\mathbf{n}$, and the center of the turntable $\mathrm{X}_{R}$. The noise of $\theta_{ik}$ follows a Gaussian distribution $\mathcal{N}(0,\sigma_{ik})$, where $\sigma_{ik}$ is the precision of the turntable.

\subsection{Calibration Procedure}\label{subsec:Procedure}
The calibration starts with estimating the direction vector $\mathbf{v}$ of the laser and ultrasound beams, followed by the rotation axis direction $\mathbf{n}$ and the center location $\mathrm{X_R}$ of the turntable (Fig. \ref{fig:obj-and-procedure}). In- and between-frame motions are operated by the 3D linear stages. The rig is firmly mounted on the flat surface of the turntable, whose base is robustly bonded with the flat printing platform of the 3D printer. 

\subsubsection{Sensor Parameter}
As shown in Fig. \ref{fig:in-frame-motion-and-beam}, the beam direction $\mathbf{v}$ is calibrated by the graphite filament, which is held horizontally with the tip pointing toward the laser beam. The calibration rig remains stationary during the entire process. After an in-frame motion, a frame of the tip is captured. The center point in the c-th frame, denoted as $\mathrm{X}_{c}$, is determined by the procedure described in Sec.~\ref{subsec:Measurements}. The sensor depth readings and position readings associated with $\mathrm{X}_{c}$, denoted as $(d_{c},\mathrm{S}_{c})$, are then recorded. A representative frame is acquired in Fig. \ref{fig:in-frame-motion-and-beam}, which is displayed as a heat map in the black box located in the bottom right corner. The center point of the tip is highlighted with a red dot. 
Once a frame is acquired, the between-frame translational motion of the sensor is applied along the Y-axis, followed by another round of in-frame motion to capture a frame scan of the tip at a distinct depth. The direction vector of the beam $\mathbf{v}$ is estimated by at least two center points from two frames.

\subsubsection{Scanning System}
\begin{figure}[!htb]
    \centering
    \subfloat[]{\includegraphics[height=1.1in ]{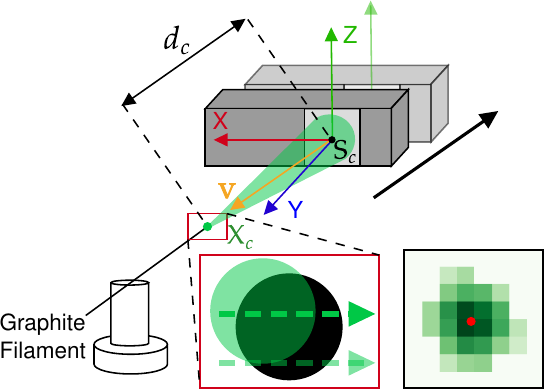}\label{fig:in-frame-motion-and-beam}} \hspace{.1in}
    \subfloat[]{\includegraphics[height=1.1in]{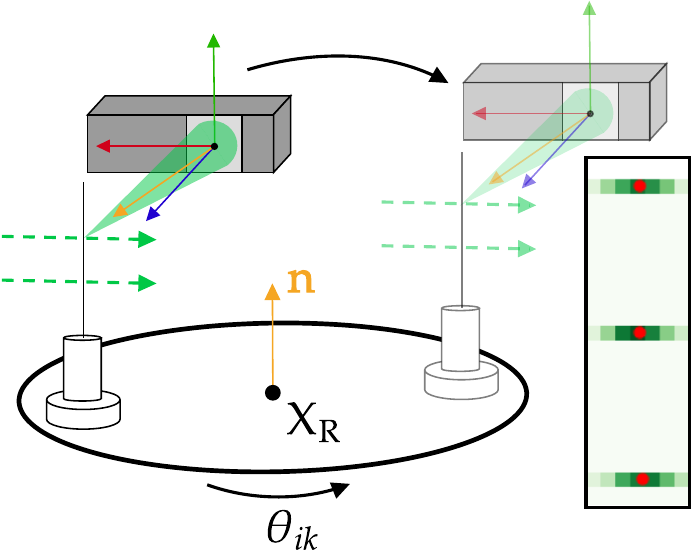}\label{fig:bw-frame-rot-platform-pose}}\vspace{.0in}
    \subfloat[]{\includegraphics[width=1.5in]{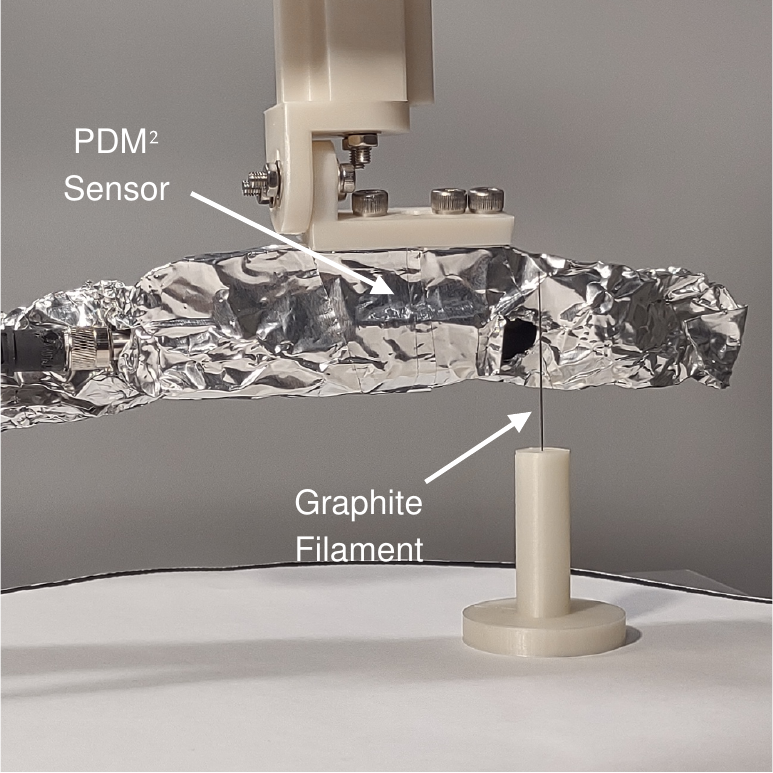}\label{fig:cali-blowup}} \hspace{0.1in}
    \subfloat[]{\includegraphics[width=1.5in]{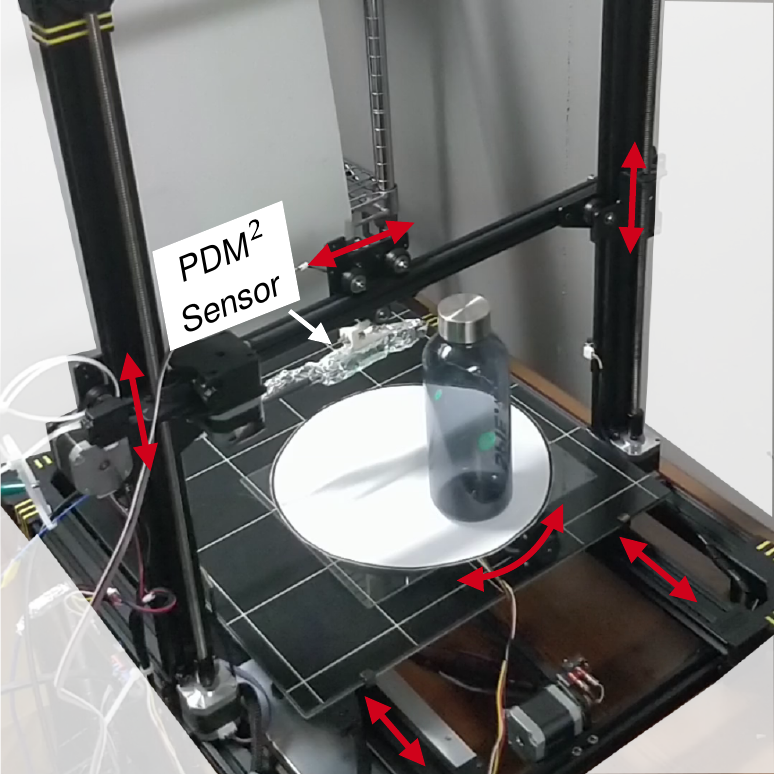}\label{fig:cali}}
    \caption{Schematics and photos of two calibration rig setup and the scanning process where the gray 3D cuboid shooting out green laser beam is the sensor: (a) The graphite filament is in horizontal configuration for sensor parameter calibration. (b) The graphite filament is in a vertical configuration for the calibration of the scanning system. In (a) and (b), in-frame and between-frame motions are colored with green dashed and black solid arrows, respectively. 
    (c) Close-up photo of the sensor and the rig on the turntable during calibration. The sensor is wrapped in aluminum foil for electromagnetic shielding.  
    (d) An overview photo of the object scanning system with a water bottle as the sample scanned object on the turntable.} \label{fig:obj-and-procedure}
\end{figure}

As shown in Fig. \ref{fig:bw-frame-rot-platform-pose},  the rotation axis $\mathbf{n}$ and the center $\mathrm{X_R}$ of the turntable are also calibrated by using the rig placed on the turntable. The graphite filament is held vertically by the rig and is to be scanned across the sidewall (or edge).  In-frame motion is applied to obtain a frame scan of the filament edge. Similarly to the previous step, the center points of the edge $\mathrm{X}_{ij}$ are determined from a frame, and their corresponding raw measurements $(d_{ij},\mathrm{S}_{ij})$ are recorded. A representative frame is shown in the black box in the bottom right of Fig. \ref{fig:bw-frame-rot-platform-pose}, and the center points of the edges are marked with red dots.
During this step, both translational movements of the sensor and rotational motion (with angle $\theta_{ik}$) of the rig mounted on the turntable are introduced as the between-frame motion. Subsequently, another iteration of in-frame motion captures a frame scan of the edge from the different sensor-rig configuration caused by the between-frame motion. Raw measurements $(d_{kj},\mathrm{S}_{kj})$ of the new edge center points of the frame are recorded to pair with their pre-rotation counterparts. Then, the rotation axis $\mathbf{n}$ and the center $\mathrm{X}_{\mathrm{R}}$ of the turntable are estimated based on at least two edge frame scans, each containing raw measurements of no less than two edge center points.

\subsection{Problem Definition}
The object scanning system is calibrated by solving a 2 degrees-of-freedom (DoF) sensor parameter calibration problem and then a 4-DoF scanning system parameters calibration problem.

\begin{Def}[Sensor Parameter Calibration]\label{def:def_sensor_calib}
Given the sensor depth measurements and position readings of the tip center points $(d_{c}, \mathrm{S}_{c})$, estimate the sensor parameter $\mathbf{v}$, which is the direction vector of the laser and ultrasound beams.
\end{Def}

\begin{Def}[Scanning System Parameters Calibration]\label{def:def_system_calib}
Given sensor parameter $\mathbf{v}$, turntable rotation angle readings $\theta_{ik}$ between a pair of edges $\mathbf{E}_i$ and $\mathbf{E}_k$, and their edge center points' sensor depth measurements and position readings $(d_{ij}, \mathrm{S}_{ij})$ and $(d_{kj}, \mathrm{S}_{kj})$, estimate the scanning system parameters $\mathbf{n}$ and $\mathrm{X_R}$, which are the rotation axis and the center point of the turntable, respectively.
\end{Def}

\subsection{Calibration Algorithm}\label{subsec:Algorithm}
The calibration algorithm is a three-step process. First, it estimates the sensor parameter $\mathbf{v}$. Second, it estimates the scanning system parameter $\mathbf{n}$ and $\mathrm{X_R}$. Lastly, it optimizes all parameters with an MLE method.

\subsubsection{Sensor Parameter $\mathbf{v}$}
Prior to diving into the beam direction estimation scheme, the procedure of retrieving a center point from sensor depth $d_{c}$ and position $\mathrm{S}_{c}$ of the $c$-th frame is demonstrated, based on the linear-stage readings.
In Fig. \ref{fig:in-frame-motion-and-beam}, it can be observed that in the scan of the $c$-th frame, the laser beam emitted from the sensor originates from the position  $\mathrm{S}_{c}$  and travels a distance $d_{c}$ along the direction vector $\mathbf{v}$. Its trajectory continues until it intersects with the point $\mathrm{X}_{c}$ on the graphite filament.
Consequently, $\mathrm{X}_{c}$ should satisfy
\begin{equation}\label{eq:point_measurement}
     \mathrm{X}_{c} = \mathrm{S}_{c} + d_{c}\mathbf{v}
\end{equation}

For the calibration of sensor parameters, the direction of the laser beam $\mathbf{v}$ is estimated by analyzing the center points of the tip obtained from multiple frame scans, such as the red dot located in the bottom right black box of Fig.~\ref{fig:in-frame-motion-and-beam}. Let $\mathcal{I}_1$ represent the set of frame indices gathered during this step. Since the same point $\mathrm{X}_\mathrm{C}$ on the filament tip is captured as center points by all frames from various depths, the ideal expectation is that $\mathrm{X}_{c} = \mathrm{S}_{c} + d_{c}\mathbf{v} = \mathrm{X}_\mathrm{C} $ for all frame indices $c\in\mathcal{I}_1$.
This implies that during the tip scanning process, all sensor positions $\mathrm{S}_{c}$ are in a straight line and parallel to the vector $\mathbf{v}$.
Therefore,
\begin{equation} \label{eq:beam_direction}
    (\mathrm{S}_{c}-\mathrm{\bar{S}}_c)\times\mathbf{v} = \mathbf{0},
\end{equation} 
where $\mathrm{\bar{S}}_c = \frac{1}{|\mathcal{I}_1|}\sum\limits_{c\in\mathcal{I}_1}\mathrm{S}_{c}$ is the averaged tip center points' corresponding sensor position, and `$\times$' is the cross product. By stacking \eqref{eq:beam_direction} for all points $\mathrm{S}_{c}$ with $i\in\mathcal{I}_1$, the least squares estimation of the beam direction $\mathbf{v}$ is obtained using the singular value decomposition (SVD).

\subsubsection{Scanning System Parameters $\mathbf{n}$ and $\mathrm{X_R}$}

During calibration of the scanning system parameters, the rotation axis $\mathbf{n}$ and the center $\mathrm{X_R}$ of the turntable are estimated by rotation motion based on the coordinates of edge center points observed in multiple frame scans (depicted as red dots in the black box below right of Fig.~\ref{fig:bw-frame-rot-platform-pose}).
First, the edges are estimated on the basis of the given edge center points. Next, the rotation motion is elaborated and the utilization of these edges to derive both $\mathbf{n}$ and $\mathrm{X_R}$ from the observed motion is demonstrated.

At least two measured center points are used to estimate an edge, which are determined by the previously estimated $\mathbf{v}$ and the raw measurements acquired by following the derivation process described in equation \eqref{eq:beam_direction}.
Let $\mathbf{E}_i$ denote the filament scanned in the i-th frame, and $\mathrm{X}_{ij}$ represent the center point of the j-th edge in $\mathbf{E}_i$. The set of point indices collected in this step can be denoted as $\mathcal{I}_2$. The direction vector $\mathbf{q}_i$ of $\mathbf{E}_i$ is parallel to the vector connecting any two points on the edge.
\begin{equation} \label{eq:edge_direction}
    (\mathrm{X}_{ij}-\mathrm{\bar{X}}_{i})\times\mathbf{q}_i = \mathbf{0}
\end{equation} 
where $\mathrm{\bar{X}}_i = \frac{1}{|\mathcal{I}_2|}\sum\limits_{j\in\mathcal{I}_2}\mathrm{X}_{ij}$ is the averaged edge center point on $\mathbf{E}_i$. The moment vector $\mathbf{m}_i$ of $\mathbf{E}_i$ is calculated by 
\begin{equation} \label{eq:moment}
\mathbf{m}_{i} = \mathrm{\bar{X}}_i \times \mathbf{q}_i
\end{equation} 
following the conventions in \cite{mason2001mechanics}. By stacking \eqref{eq:edge_direction} for all edge points $\mathrm{X}_{ij}$ in $\mathbf{E}_i$, the least squares estimate of the direction vector $\mathbf{q}_i$ is obtained using SVD. The moment vector $\mathbf{m}_i$ of the edge $\mathbf{E}_i$ is determined by the solution provided by \eqref{eq:moment}, which completes the estimation of $\mathbf{E}_i = [[\mathbf{q}_i]_\times \textrm{  }  \mathbf{m}_i]$. The edge $\mathbf{E}_k$ is estimated in the same manner from its raw measurements $(d_{kj}, \mathrm{S}_{kj})$.

The rotational movement of the turntable $\mathbf{T}_\mathrm{R}$ between two successive frame scans can be broken down into three distinct stages: translating the origin of $\{\mathbf{0}\}$ to the center of the turntable, then rotating motion, and finally translating the origin back. The rotation motion $\mathbf{T}_{\mathrm{R}}$ can be written as
\begin{equation} \label{eq:RotationMotion}
\begin{split}
\mathbf{T}_\mathrm{R} &= 
\begin{bmatrix}\mathbf{I} & \mathrm{X}_\mathrm{R} \\ \mathbf{0} & 1 \end{bmatrix}\begin{bmatrix}\mathbf{R}_{\mathbf{n}}(\theta) & \mathbf{0} \\ \mathbf{0} & 1 \end{bmatrix}\begin{bmatrix}\mathbf{I} & -\mathrm{X}_\mathrm{R} \\ \mathbf{0} & 1 \end{bmatrix} \\ 
&= \begin{bmatrix}\mathbf{R}_{\mathbf{n}}(\theta) & (\mathbf{I}_{3\times 3}-\mathbf{R}_{\mathbf{n}}(\theta))\mathrm{X}_\mathrm{R} \\ \mathbf{0} & 1 \end{bmatrix},
\end{split}
\end{equation} 
where $\mathbf{R}_\mathbf{n}(\theta) = \mathbf{I}_{3\times 3}+\sin\theta[\mathbf{n}]_\times+(1-\cos\theta)[\mathbf{n}]_\times^2$ is the Rodrigues' formula for axis-angle rotation, $\mathbf{n}$ is the normal vector of turntable surface, and $\theta$ is the rotation angle of the motion. Additionally, $\mathrm{X}_\mathrm{R}$ is the center of the turntable.

Using a minimum of two edges and their rotational motion, the scanning system parameters $\mathbf{n}$ and $\mathrm{X}_\mathrm{R}$ can be estimated. 
Let $\mathbf{E}_k$ represent the corresponding edge counterpart of $\mathbf{E}_i$ after applying the rotation motion $\mathbf{T}_\mathrm{R}$. The direction vector $\mathbf{q}_{i}$ of the edge $\mathbf{E}_i$ is parallel to its counterpart $\mathbf{q}_{k}$ after rotation.
\begin{equation} \label{eq:RotationStageR}
\mathbf{R}_\mathbf{n}(\theta)\mathbf{q}_{i} \times \mathbf{q}_{k} = \mathbf{0}.
\end{equation} 
Furthermore, a point $\mathrm{X}_{ij}$ on the edge $\mathbf{E}_i$ lies on the transformed edge $\mathbf{E}_{k}$ after rotation
\begin{equation} \label{eq:RotationStageX}
\mathbf{E}_{k}(\mathbf{T}_\mathrm{R}\mathbf{X}_{i}) = \mathbf{0},
\end{equation} 
where $\mathbf{X}_i = \begin{bmatrix}\ldots, \begin{bmatrix}\mathrm{X}_{ij}\\1\end{bmatrix}, \ldots \end{bmatrix}$ are all the edge points on $\mathbf{E}_i$. 
Using SVD,
the normal vector $\mathbf{n}$ of the turntable and the center position $\mathrm{X}_\mathrm{R}$ can be determined by combining \eqref{eq:RotationStageR} and \eqref{eq:RotationStageX}.

\subsubsection{MLE Estimation}

Using the least-squares estimations for $\mathbf{v}$, $\mathbf{n}$ and $\mathrm{X}_\mathrm{R}$ derived from the preceding steps, we re-estimate all parameters using MLE to improve accuracy by assuming Gaussian noise.

Suppose that $\mathbf{E}_k$ is the counterpart of edge $\mathbf{E}_i$ after applying the rotation motion $\mathbf{T}_\mathrm{R}$. The collection of raw measurements from the two frames is denoted as $\mathcal{X}_{ik} = [\theta_{ik},\mathbf{d}_{i}^\mathsf{T},\mathbf{S}_{i}^\mathsf{T},\mathbf{d}_{k}^\mathsf{T},\mathbf{S}_{k}^\mathsf{T}]^\mathsf{T}$. Here $\theta_{ik}$ is the degree of rotation of the turntable between the i-th and k-th frame. $\mathbf{d}_{i} = [\ldots,d_{ij},\ldots]^\mathsf{T}$ and $\mathbf{S}_{i} = [\ldots,\mathrm{S}_{ij},\ldots]^\mathsf{T}$ are depth and sensor position readings of edge center points on edge $\mathbf{E}_i$, $\mathbf{d}_{k}$ and $\mathbf{S}_{k}$ are those for edge $\mathbf{E}_k$.

The MLE optimization problem is formulated as follows.
\begin{equation} \label{eq:MLE}
\begin{split}
& \min_{\mathbf{v}, \mathbf{n}, \mathrm{X}_\mathrm{R}} \sum_{i\neq k}\left\|\mathcal{X}_{ik}-f_{ik}(\mathbf{v}, \mathbf{n}, \mathrm{X}_\mathrm{R})\right\|^2_{\Sigma_{ik}} + \sum_{c}\left\|\mathrm{S}_{c}-f_{c}(\mathbf{v})\right\|^2_{\Sigma_{S_c}}
\end{split}
\end{equation}
where $\left\|\cdot\right\|^2_{\Sigma_{ik}}$ is Mahalanobis distance with covariance matrix $\Sigma_{ik}$ of $\mathcal{X}_{ik}$, and $\Sigma_{ik} = \mbox{Diag}(\sigma_{ik}, \Sigma_{\mathrm{S}i},\Sigma_{\mathrm{d}i},\Sigma_{\mathrm{S}k},\Sigma_{\mathrm{d}k})$. The covariance matrices $\Sigma_{\mathrm{S}i} = \mbox{Diag}(\ldots,\Sigma_{ij},\ldots)$ and $\Sigma_{\mathrm{d}i} = \mbox{Diag}(\ldots,\sigma_{ij},\ldots)$ are obtained from Sec.~\ref{subsec:Measurements}, and the same applies for $\sigma_{ik}$, $\Sigma_{\mathrm{S}k}$, $\Sigma_{\mathrm{d}k}$ and $\Sigma_{S_c}$. The prediction function $f_{ik}(\mathbf{v}, \mathbf{n}, \mathrm{X}_\mathrm{R})$ is derived from \eqref{eq:edge_direction}, \eqref{eq:RotationStageR} and \eqref{eq:RotationStageX} and $f_{c}$ is derived from \eqref{eq:beam_direction}. The MLE problem is solved by the Levenberg-Marquardt algorithm.

This MLE optimization can be simplified when the PDM$^2$ sensor is mounted on a pre-calibrated scanning system like a robot manipulator. In this case only the sensor parameter needs to be refined as
\begin{equation} \label{eq:MLE_v}
\begin{split}
& \min_{\mathbf{v}} \sum_{c}\left\|\mathrm{S}_{c}-f_{c}(\mathbf{v})\right\|^2_{\Sigma_{S_c}}
\end{split}.
\end{equation}

\section{Experimental Results}\label{sec:exp}

We have fabricated the new PDM$^2$ sensor and assembled the object scanning system to verify the three main contributions: sensor design, signal processing algorithms, and calibration. 
The experiments are organized as follows. We begin with the estimation of ToF in Sec.~\ref{ssc:Tof-exp}. The result of ToF determinants ranging performance, which is validated in Sec.~\ref{ssc:ranging-exp}. Segmented signal sequences are fed into the BOSS classifier to allow us to verify the material / interior structure of the scanned targets in Sec.~\ref{ssc:material-exap}. These algorithms are implemented using  Matlab\texttrademark~R2021a. The overall calibration performance is validated in Sec.~\ref{ssc:cali-exp}. Finally, we perform contour reconstruction of common household objects in Sec.~\ref{ssc:reconstruction-exp} as a demonstration of overall sensor capability.

\subsection{ToF Auto estimation}\label{ssc:Tof-exp}
\begin{figure}[!htb]
    \centering
    \subfloat[]{\includegraphics[width=1.6in ]{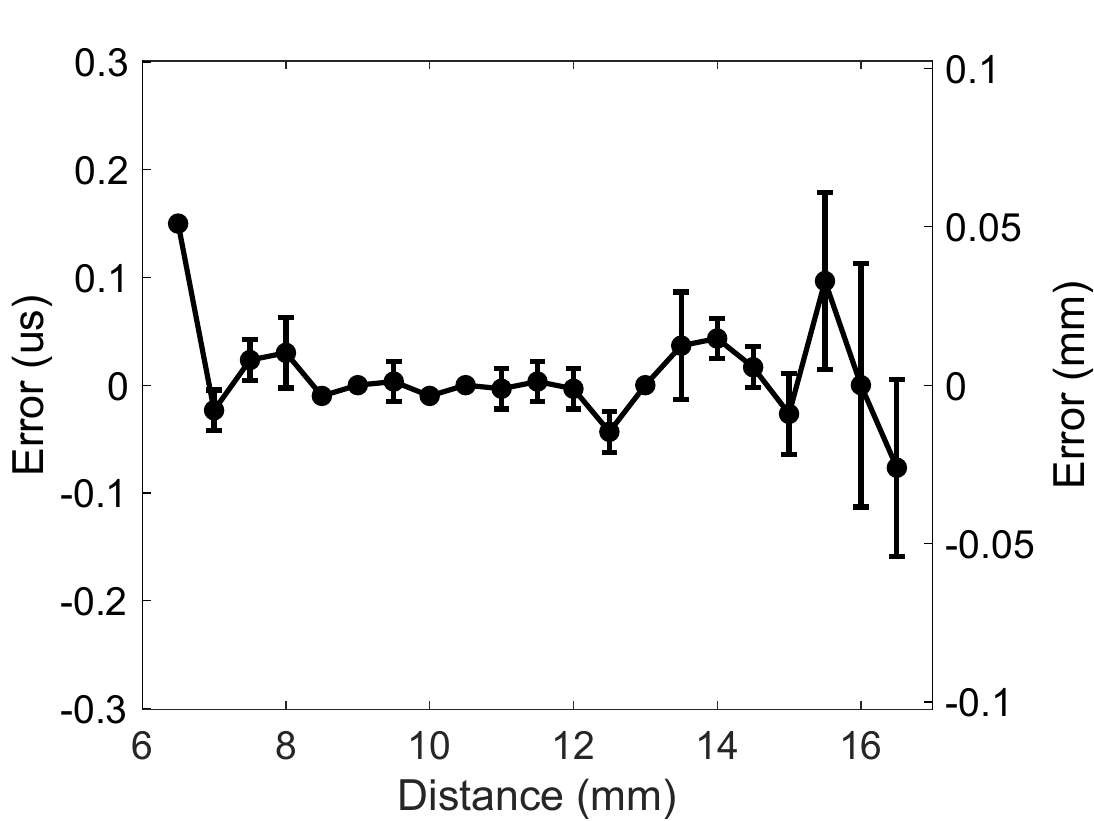}\label{fig:exp:ToF-OA}} \hspace{.0in}
    \subfloat[]{\includegraphics[width=1.6in]{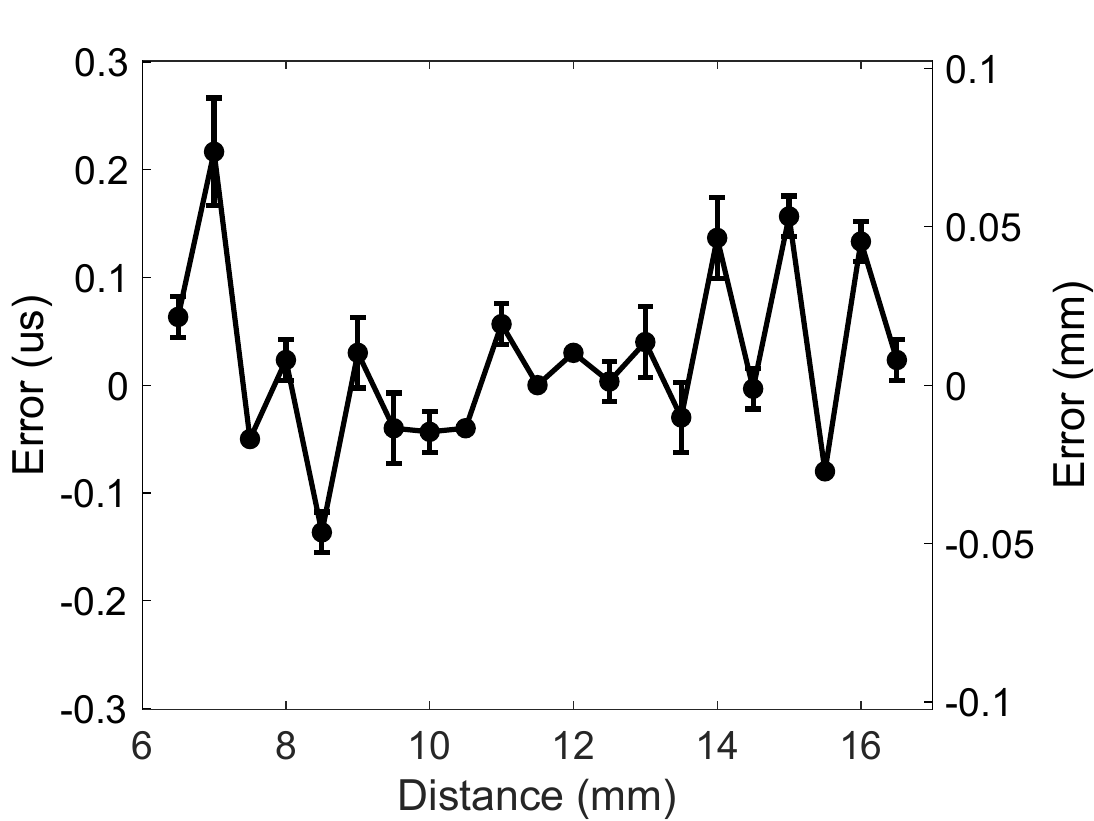}\label{fig:exp:ToF-US}}\vspace{.0in}
    \caption{Errors of the automatically-acquired ToFs  for both (a) US and (b) OA modalities.} \label{fig:exp:ToF-auto-acuqisition}
\end{figure}

The accuracy of ToF auto-estimation is tested in a ranging experiment. An aluminum block with optimal OA generation and acoustic reflection is used as the target to provide dual-modality signals. The sensor-target distance is adjusted with a step length of 0.50 mm within the range $\mathcal{D}_m= [6.50,\, 16.50]$ mm. At each step, three data are collected. The ranging errors of the automatically acquired ToFs of the OA and US signals are obtained by comparing the ToFs to the corresponding manually labeled ground-truth results. Comparisons are shown in Fig. \ref{fig:exp:ToF-auto-acuqisition}, which are less than 0.3 $\mu$s or 0.1 mm throughout the distance range and even less within the Depth of Focus $\mathcal{D}_f=[8.0,\, 13.0]$ mm for both modalities. These error values outside the Depth of Focus $\mathcal{D}_f$ tend to increase, mainly due to the relatively weak signal-to-noise ratio. The accuracy shown in the results is well within the requirement of precise grasping, which indicates that our automatic ToF estimation is successful in both US and OA modalities.

\subsection{Distance Ranging}\label{ssc:ranging-exp}

With the validated automatic ToF estimation method, we are ready to test the overall range performance for both modalities.

\subsubsection{US Ranging}
\begin{figure}[!htbp]
    \centering
    \subfloat[]{\includegraphics[height=1.26in]{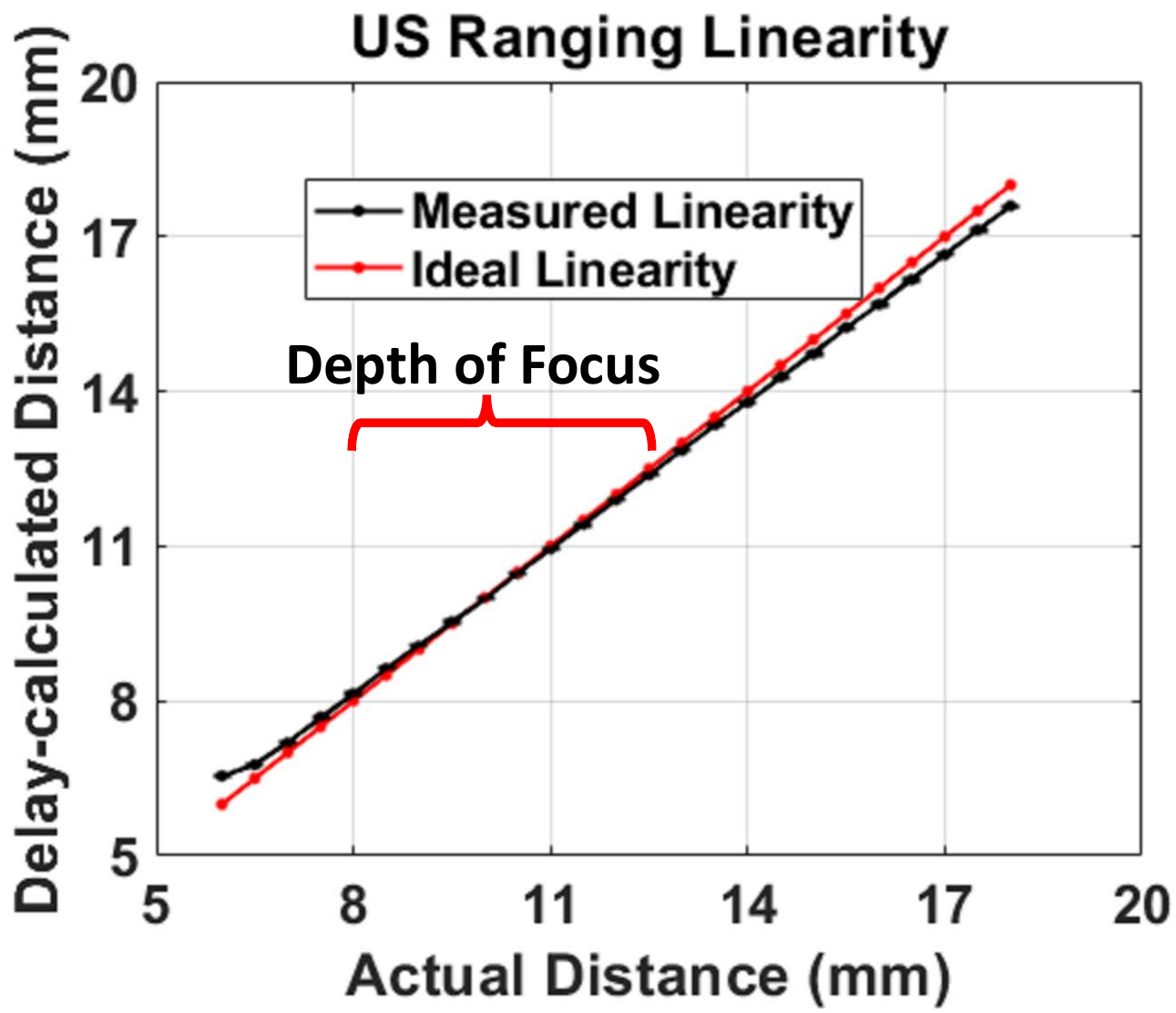}\label{fig:exp:US-ranging-exp}}
    \hspace*{.1in}
    \subfloat[]{\includegraphics[height=1.26in]{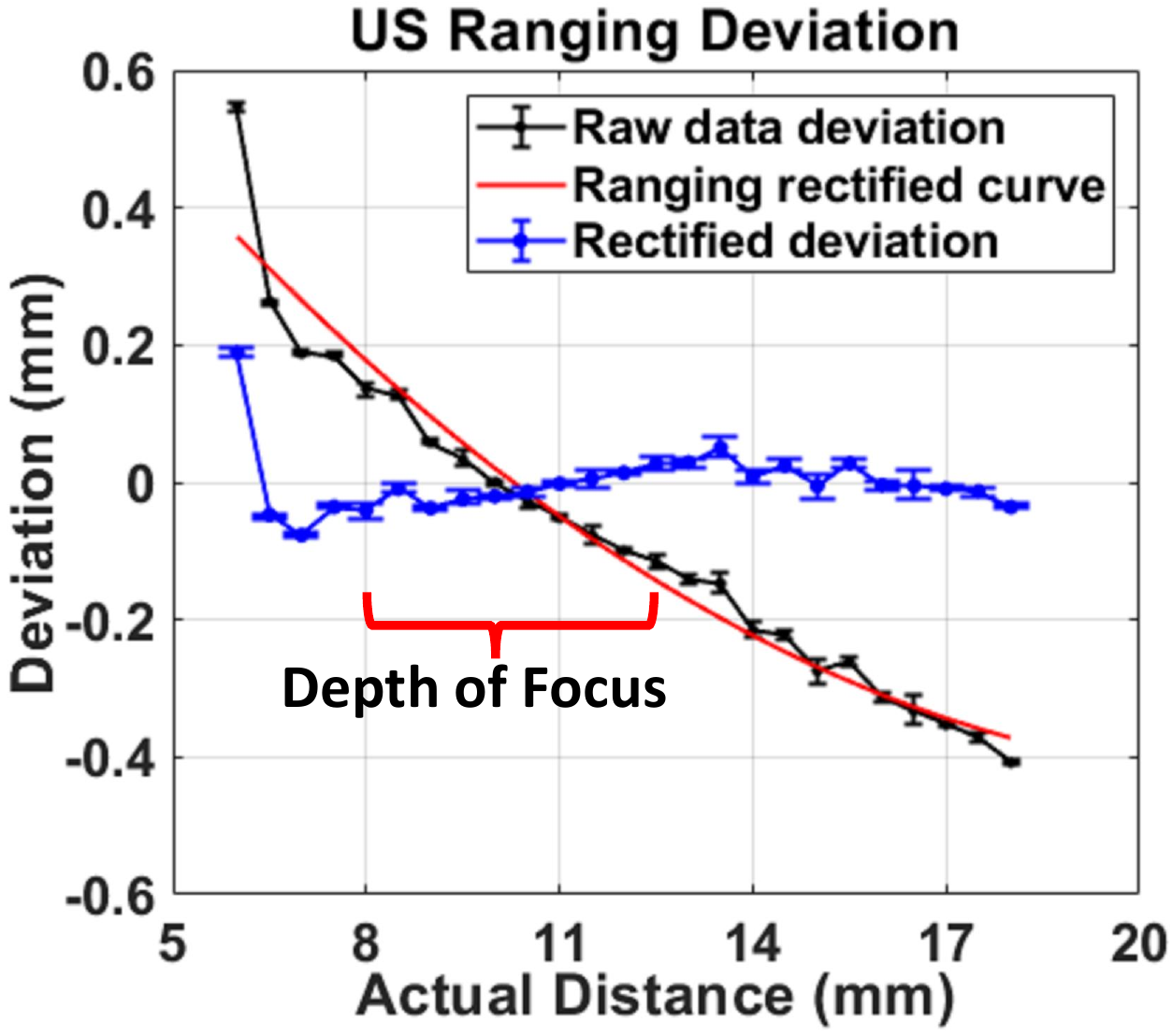} \label{fig:exp:US-residual}}\\
    \subfloat[]{\includegraphics[width=1.26in]{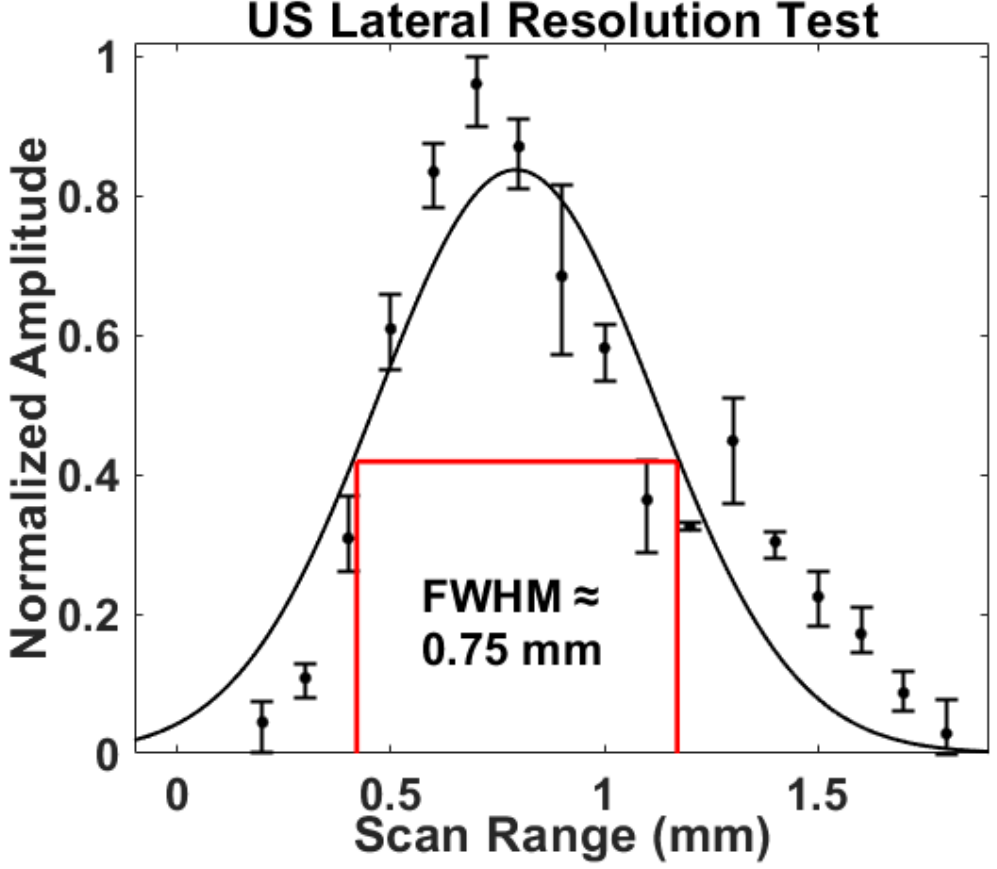}\label{fig:exp:US-lateral-resolution}}
    \caption{(a) Comparison between the US measured (in black) and actual (in red) distances. (b) Deviation of the US measured distance from the real distance. (c) US lateral resolution around 0.75 mm determined by the minimal acoustic focal diameter at d = 10.0 mm.}
    \label{fig:exp:US-ranging-retification}
\end{figure}

The US range performance of the new PDM$^2$ sensor is tested in the setup described in Fig. \ref{fig:schematic-design} by using a 1.00 mm thick glass slide as the target. The distance $d$ between the parabolic mirror and the glass slide is decreased from 18.00 mm to 6.00 mm with a step length of 0.50 mm. The measured distance vs. the real distance and their deviations are shown in Figs. \ref{fig:exp:US-ranging-exp} and \ref{fig:exp:US-residual}, respectively. The raw ranging deviations are up to 0.60 mm when the target is far out of the Depth of Focus. To minimize the deviation for out-of-depth focus measurements and improve the working distance range, the range rectification algorithm in Sec.~\ref{ssc:ranging-rect-alg} is applied which shown as the red curve in Fig.~\ref{fig:exp:US-residual}. After that, the rectified deviation is less than 0.20 mm, where $d$ is from 6.0 mm to 18.0 mm, as shown by the blue curve in Fig.~\ref{fig:exp:US-residual}. 

The same setup is used to measure the lateral resolution of the US ranging, except that the glass slide target is replaced by a copper wire with a diameter around 0.7 mm. After repeating lateral scans at different distances ($d$) from 6.0 to 18.0 mm, the lateral resolution of the ultrasound is determined by the minimal focal diameter of the acoustic (Fig. \ref{fig:exp:US-lateral-resolution}), indicating lateral resolution around 0.75 mm at focal length $d$ = 10.0 mm. The measured Depth of Focus is around 4.5 mm, where $d$ changes from 8.00 mm to 12.50 mm. 

\subsubsection{OA Ranging}
\begin{figure}[!htbp]
    \centering
    \subfloat[]{\includegraphics[height=1.26in]{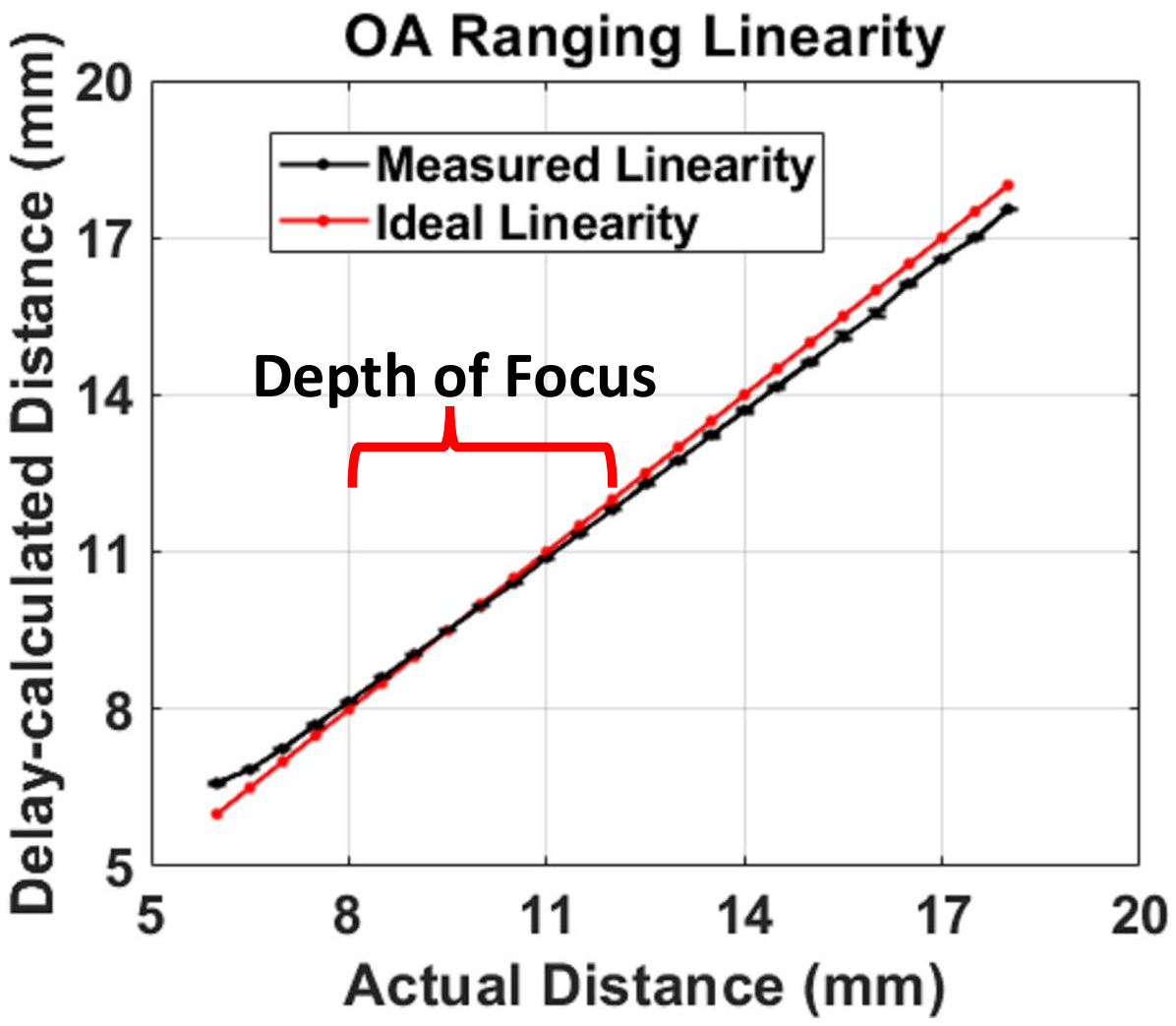}\label{fig:exp:OA-ranging}} 
    \hspace*{.1in}
    \subfloat[]{\includegraphics[height=1.26in]{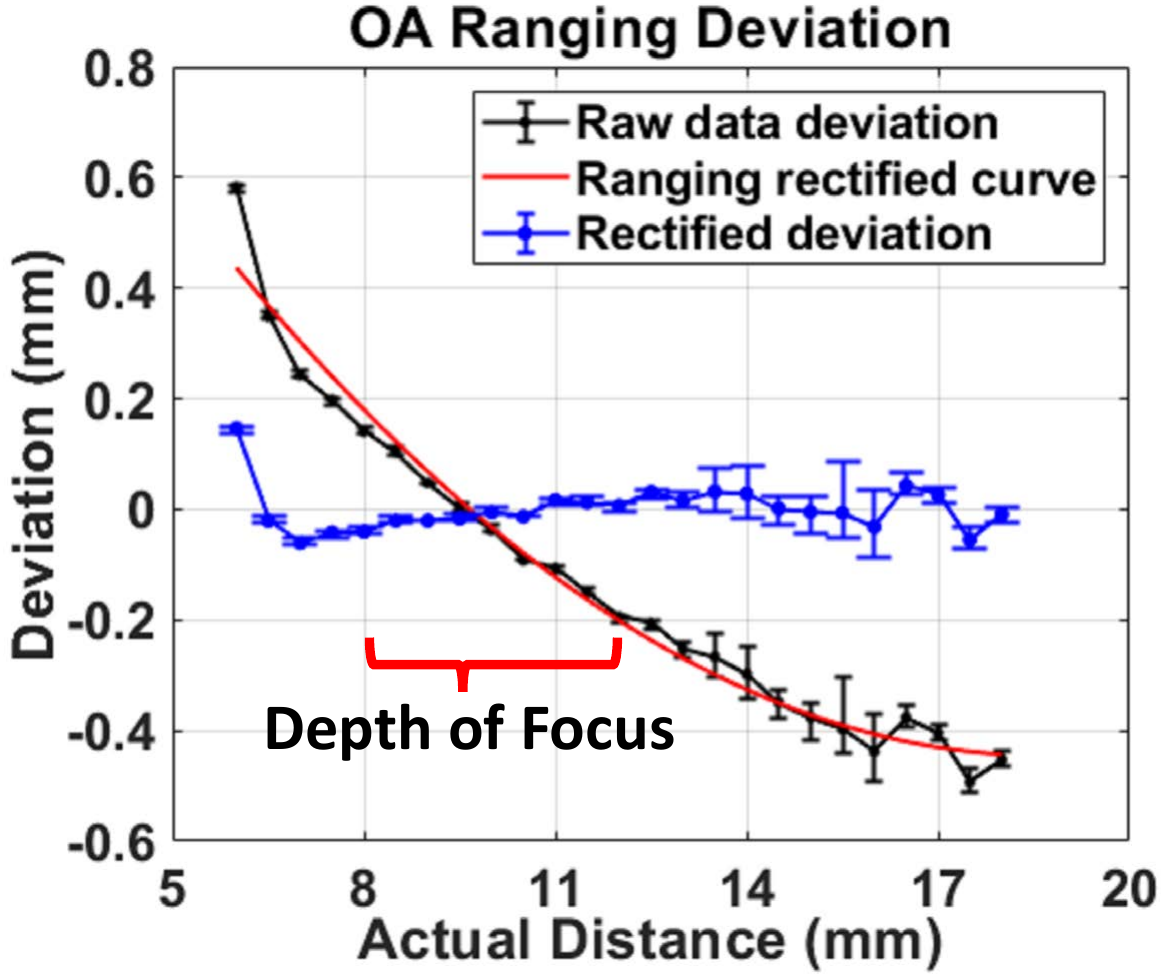}\label{fig:exp:OA-residual}}\\
    \subfloat[]{\includegraphics[width=1.26in]{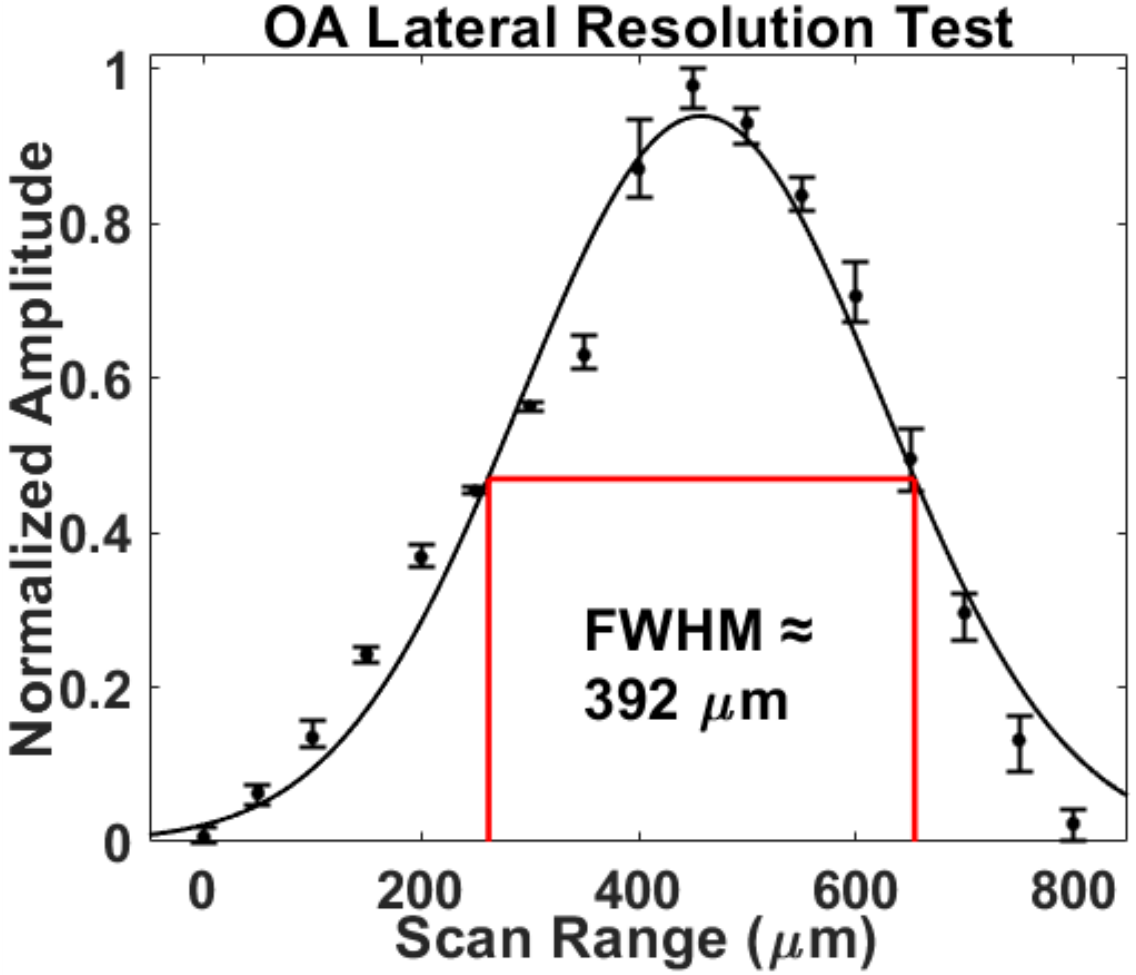}\label{fig:exp:OA-lateral-resolution}}
    \caption{(a) Comparison between the OA measured (in black) and the real (in red) distances. (b) Deviation of the measured distance of OA from the real distance. (c) OA lateral resolution of 392 $\mu$m determined by the minimal focal diameter of the OA at $d$ = 9.5 mm.}
    \label{fig:exp:OA-ranging-retification}
\end{figure}

The OA ranging performance of the new PDM$^2$ sensor is tested under the same setup as the previous US ranging using an aluminum block as target.  The measured distance vs. the real distance ($d$) and their deviations are shown in Figs. \ref{fig:exp:OA-ranging} and \ref{fig:exp:OA-residual}, respectively. Similarly, after applying the ranging rectification in Sec.~\ref{ssc:ranging-rect-alg}, the deviation is reduced to less than 0.16 mm. The same setup is used to test the OA lateral resolution, except that the aluminum block target is replaced by a tungsten wire with a diameter of around 0.4 mm. After repeating lateral scans at different distances ($d$) from 6.0 mm to 18.0 mm, the lateral resolution of the OA is determined by the minimal focal diameter of the OA (Fig. \ref{fig:exp:OA-lateral-resolution}), indicating a lateral resolution ~392 $\mu$m at the focal length ($d$ = 9.5 mm). The measured Depth of Focus is around 4.0 mm where $d$ changes from 8.00 mm to 12.00 mm.

\subsubsection{US/OA Ranging vs. Surface Angle}
The US/OA ranging accuracy may be affected by the normal direction of the target. We are interested in studying the sensitivity to the normal direction issue because it would affect how the PDM$^2$ sensor can be used.

\begin{figure}[!htbp]
    \centering
    \subfloat[]{\includegraphics[height=1.0in]{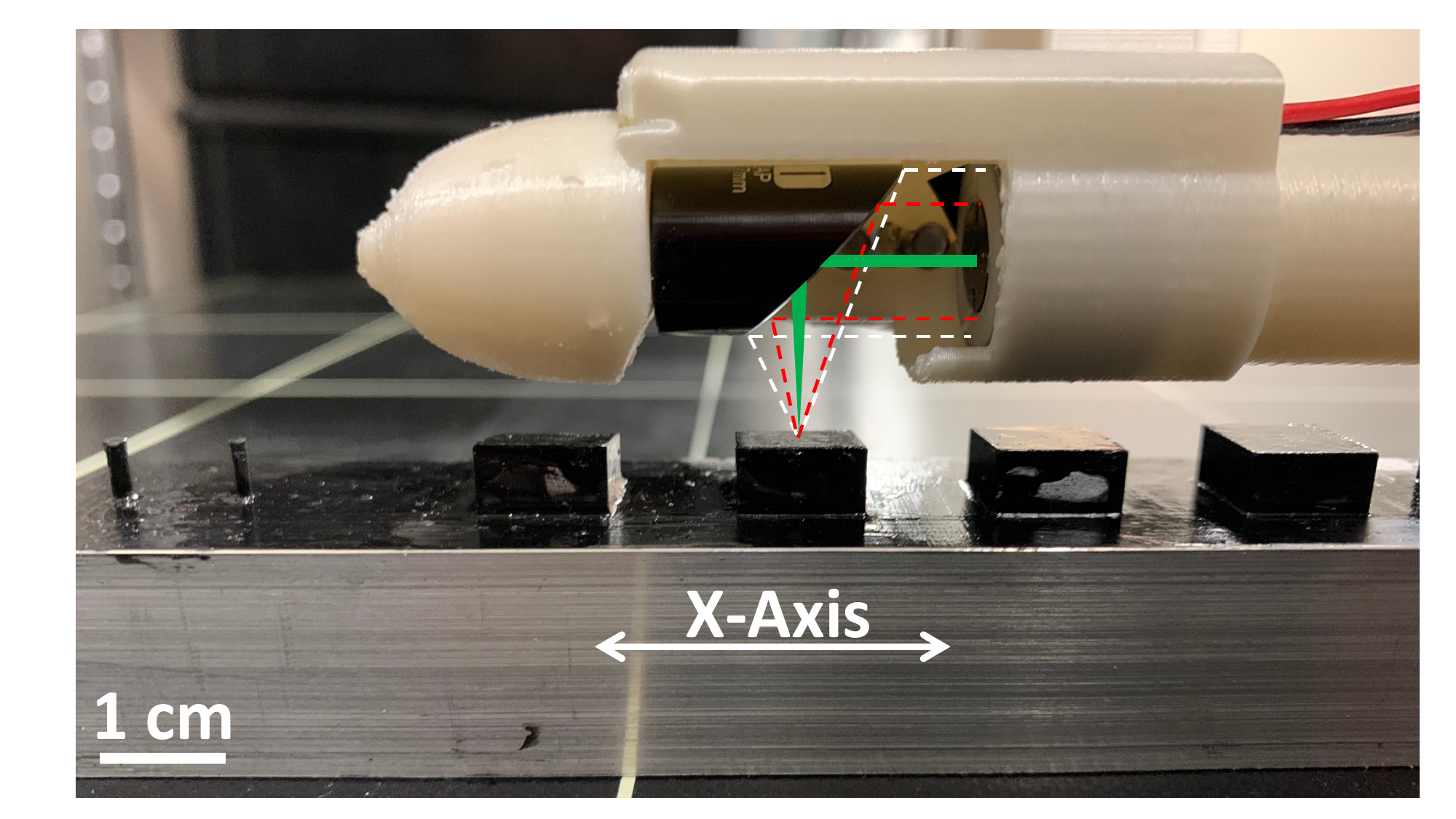}\label{fig:exp:Photos-Al-block-X}} \hspace*{.1in}
    \subfloat[]{\includegraphics[height=1.0in]{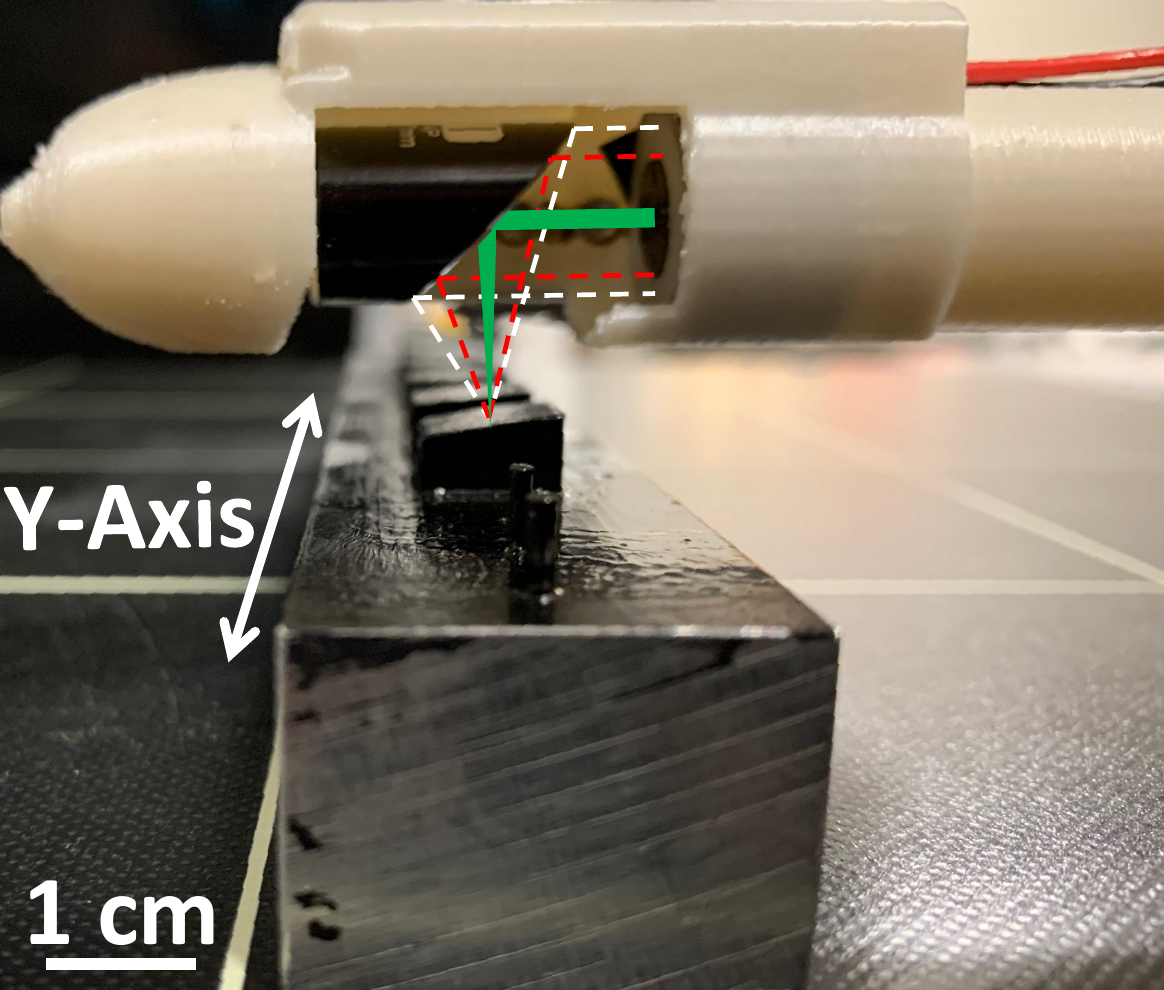}\label{fig:exp:Photos-Al-block-Y}}\\
    \subfloat[]{\includegraphics[height=1.26in]{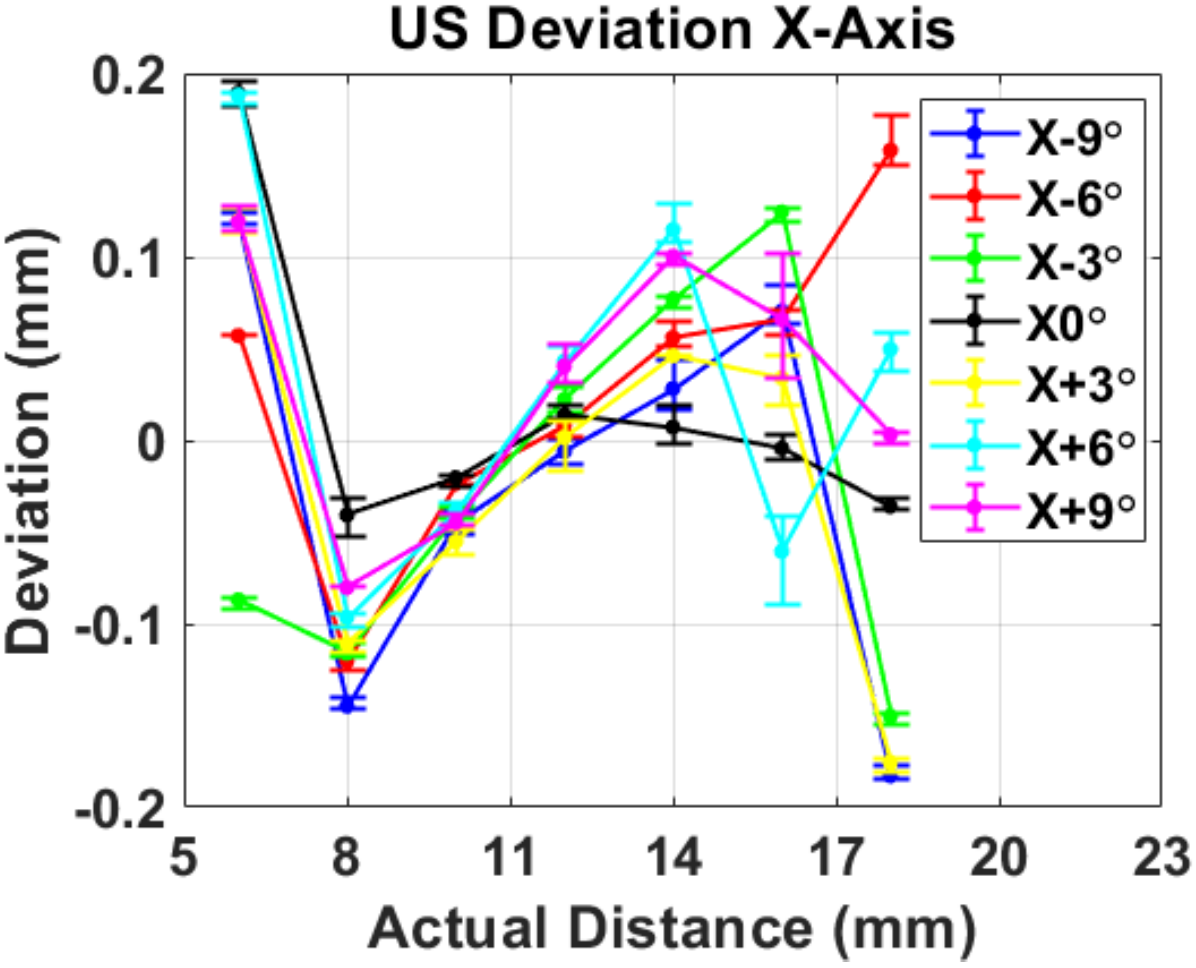}\label{fig:exp:OA-residual-X}} \hspace*{.1in}
    \subfloat[]{\includegraphics[height=1.26in]{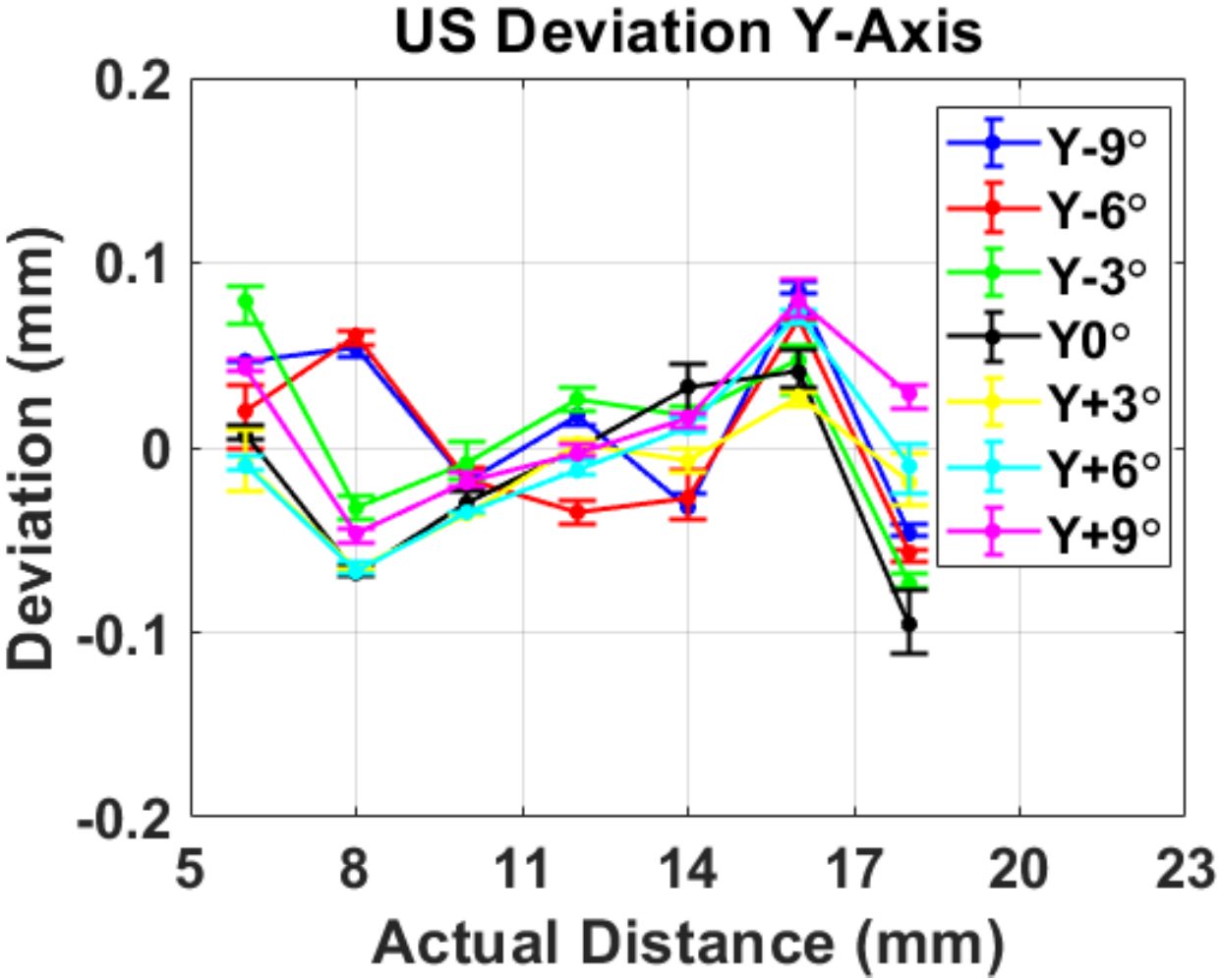}\label{fig:exp:OA-residual-Y}}\\
    \subfloat[]{\includegraphics[height=1.26in]{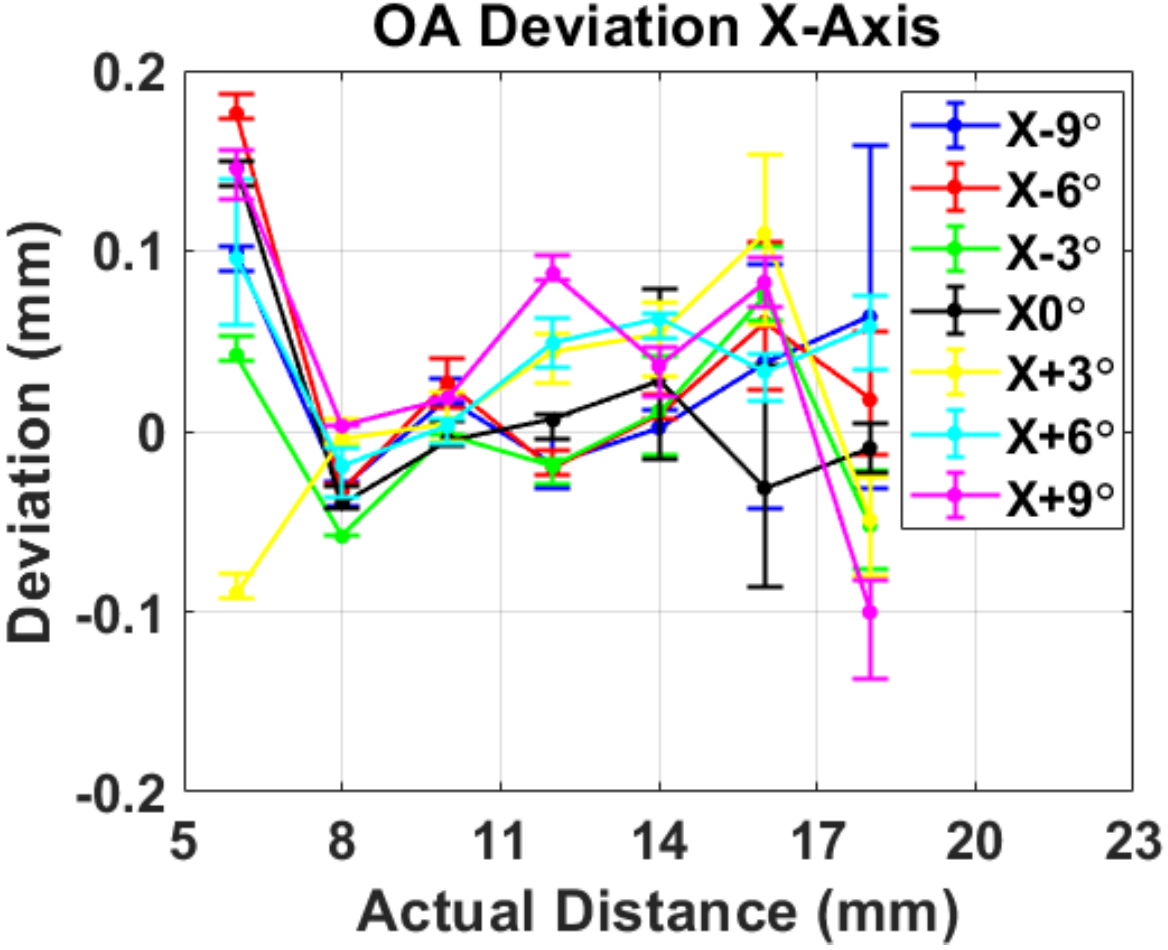}\label{fig:exp:US-residual-X}} \hspace*{.1in}
    \subfloat[]{\includegraphics[height=1.26in]{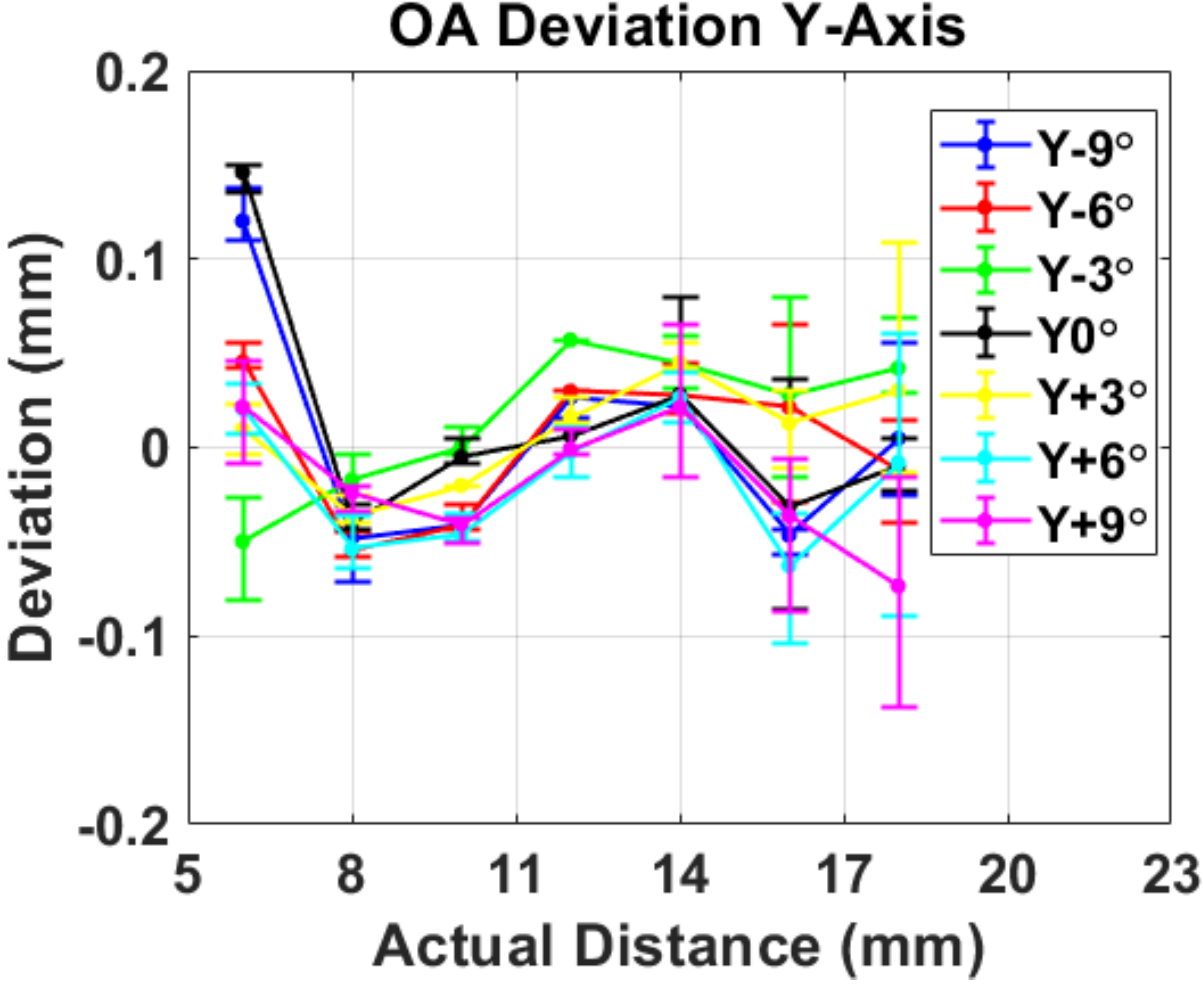}\label{fig:exp:US-residual-Y}}
    \caption{(a) and (b): Photos of the Al block with surfaces of different angles scanned on the (a) X and (b) Y axes, where the US transmission and US/OA reception paths and signals envelope are drawn using the red- and white-dashed lines, respectively. The pulsed laser beam is indicated by green color. (c) and (d): US ranging deviations vs. surface angles on the (c) X and (d) Y axes at different distances, respectively. OA ranging deviations vs. surface angles on the (e) X and (f) Y axes at different distances, respectively.}
    \label{fig:exp:OA-ranging-retification}
\end{figure}

The US/OA ranging deviations vs. target surface angle of the new PDM$^2$ sensor is tested under the setup in Figs. \ref{fig:exp:Photos-Al-block-X} and \ref{fig:exp:Photos-Al-block-Y}, using a custom-built aluminum block with surfaces precisely machined at different angles (±3º, ±6º, ±9º) as the target. The pillars at both ends are used as alignment markers to maintain the same scanning path at different sensor-target distances. To boost the OA amplitude, a thin layer of black paint is coated on the aluminum surface. The aluminum block is scanned on the X and Y axes, and the measured ranging deviations vs. angle at different distances are shown in Figs. \ref{fig:exp:OA-residual-X}, \ref{fig:exp:OA-residual-Y}, \ref{fig:exp:US-residual-X}, and \ref{fig:exp:US-residual-Y}, after the aforementioned ranging rectification. For distances from 6.00 mm to 18.00 mm, the maximum deviations US / OA are less than 0.2 mm, which indicates the robust tolerance of the PDM$^2$ sensor at different angles of the target surface. 

\subsection{Material/Structure Sensing} \label{ssc:material-exap}
Material detection and subsurface structure detection are unique capabilities of our PDM$^2$ sensor. Both are based on spectrum analysis of the US and OA signals. We have tested the capabilities here.
\subsubsection{Data Acquisition and Classification}
Both the low- and high-frequency components of the US and OA signals received by the ring PZT transducer are used for material / subsurface structure (especially thickness) sensing. Depending on the optical and mechanical properties of the target, the received signals consist mainly of either a target-induced OA signal, or a target-reflected US signal, or both, which provide distinctive features for target material/structure sensing. Data $\mathbf{o}$ are processed by the BOSS classifier for material / structure differentiation, whose performance is guaranteed by the advantages of structure-based similarity and noise resistance. The data entered into the BOSS classifier are randomly divided into two groups by a ratio of $3:1$ without overlap, which are used for training and testing, respectively. After division, the data are transformed into BOSS histograms that serve as a set of characteristics for differentiation. After 50 random trials, the confusion matrix is concluded to show the average accuracy of the correct differentiation.

\subsubsection{Material/Structure Differentiation}
\begin{figure}[!htbp]
    \centering
    \subfloat[]{\includegraphics[height=1.26in]{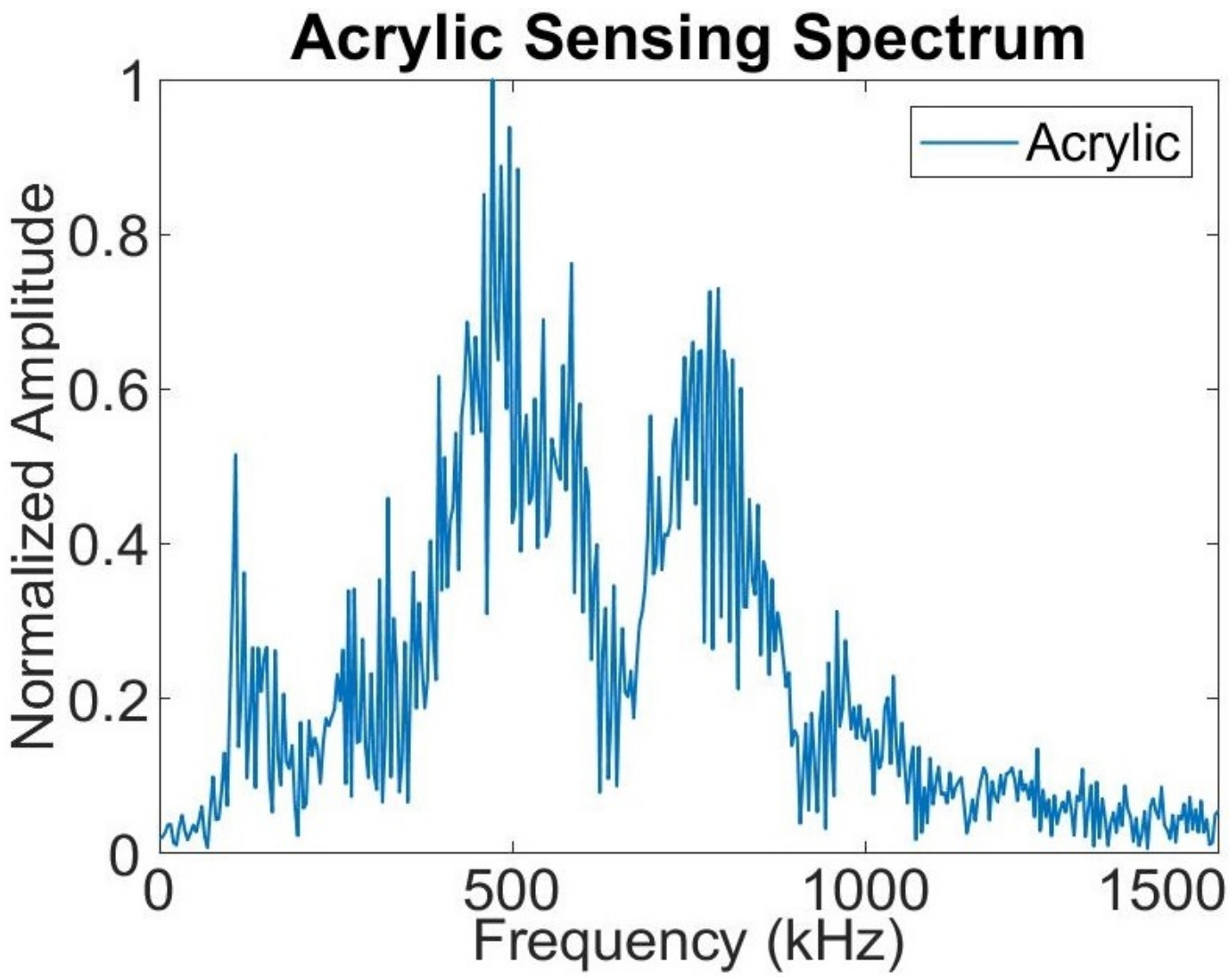}\label{fig:exp:spectrum-acrylic}} \hspace*{.1in}
    \subfloat[]{\includegraphics[height=1.26in]{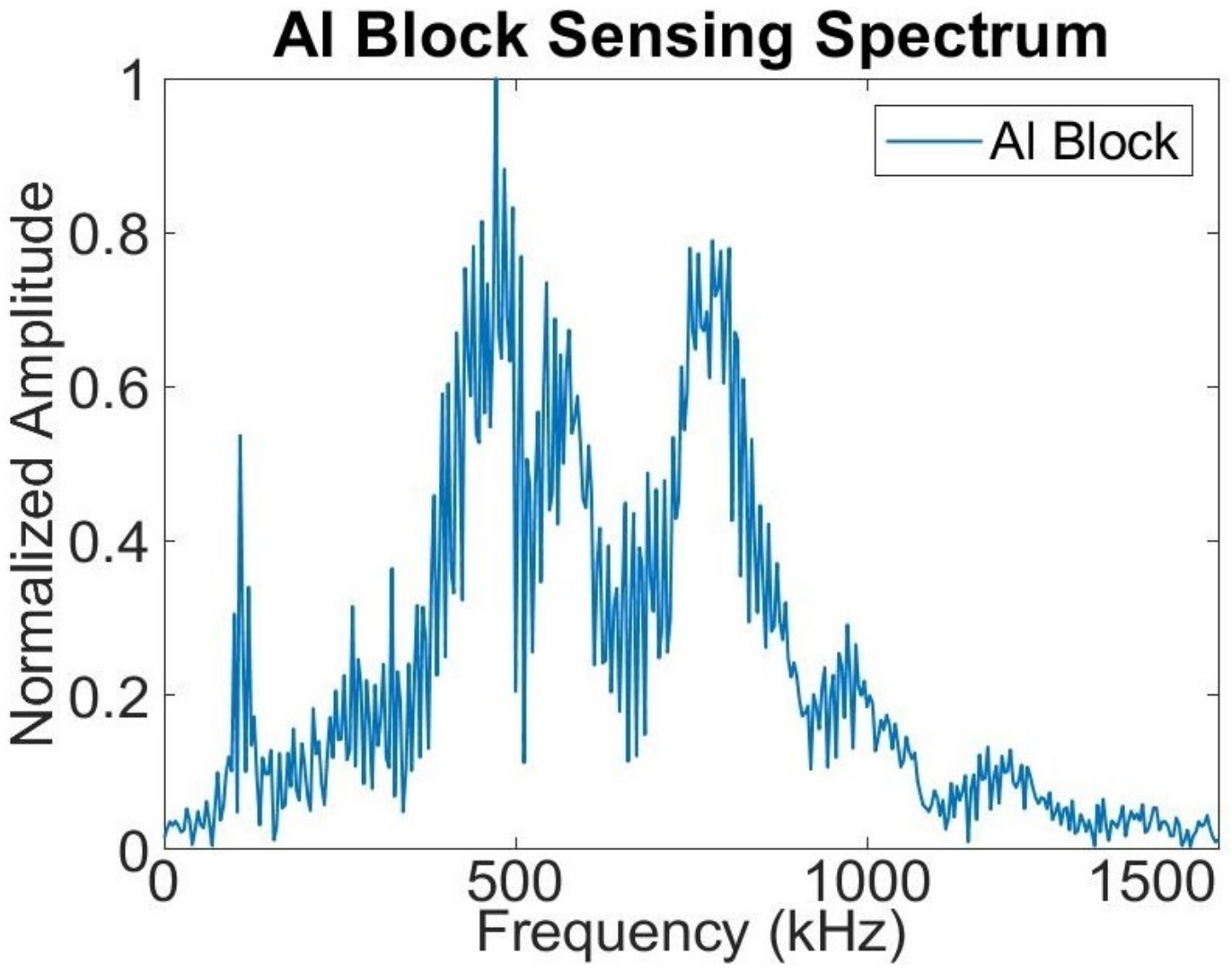}\label{fig:exp:spectrum-Al}}\\
    {\includegraphics[height=1.26in]{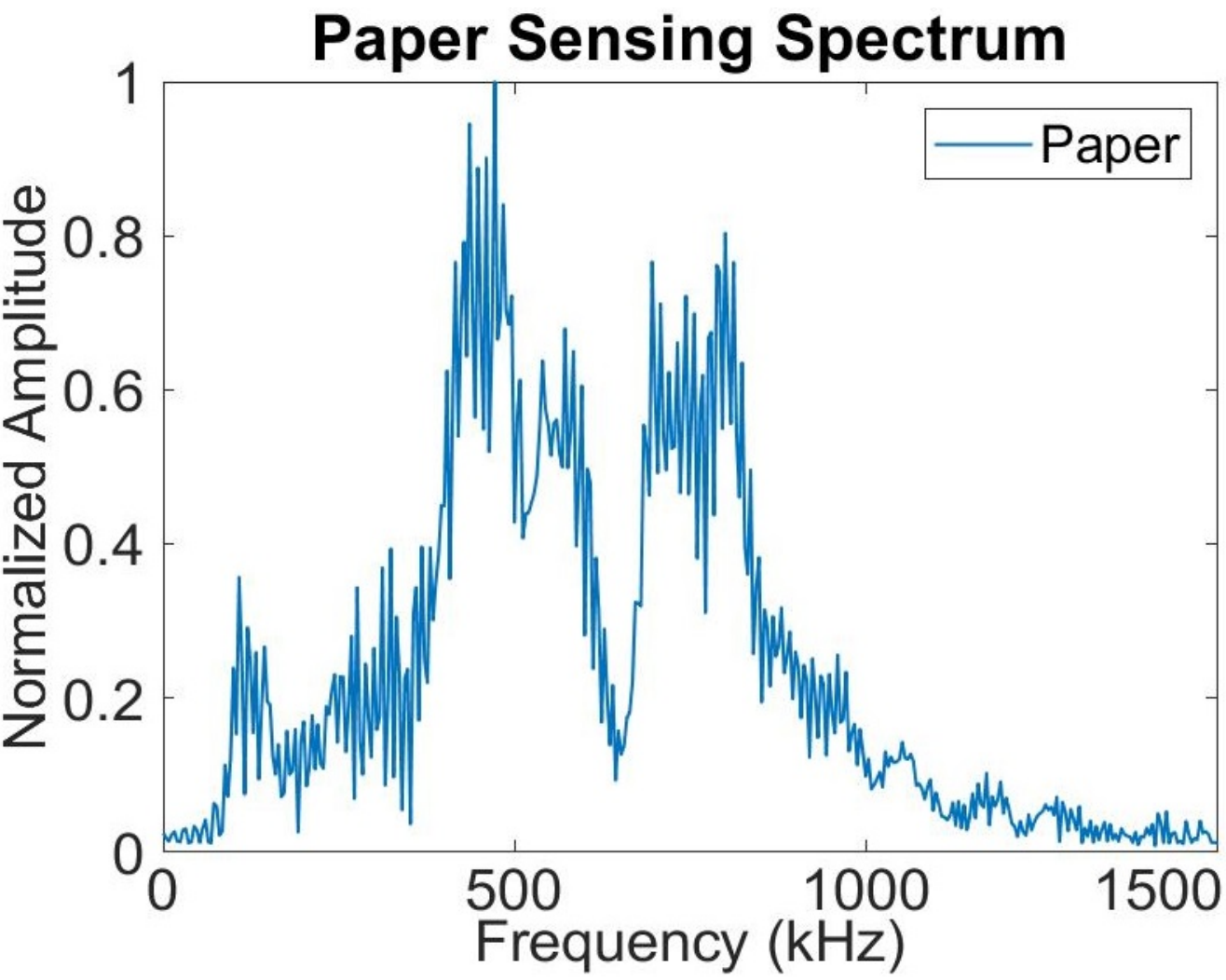}\label{fig:exp:spectrum-paper}} \hspace*{.1in}
    \subfloat[]{\includegraphics[height=1.26in]{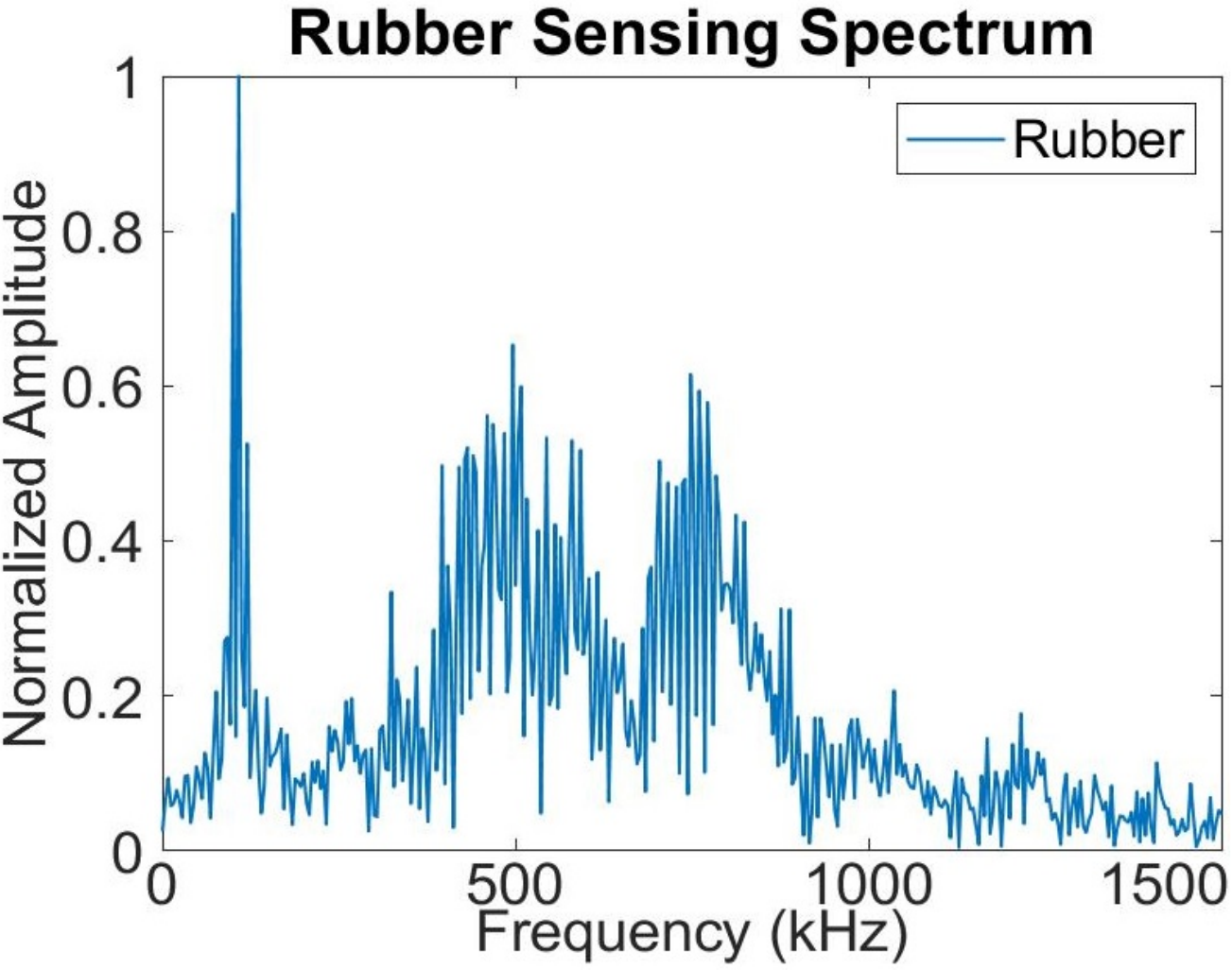}\label{fig:exp:spectrum-rubber}}\\
    \subfloat[]{\includegraphics[width=1.7in]{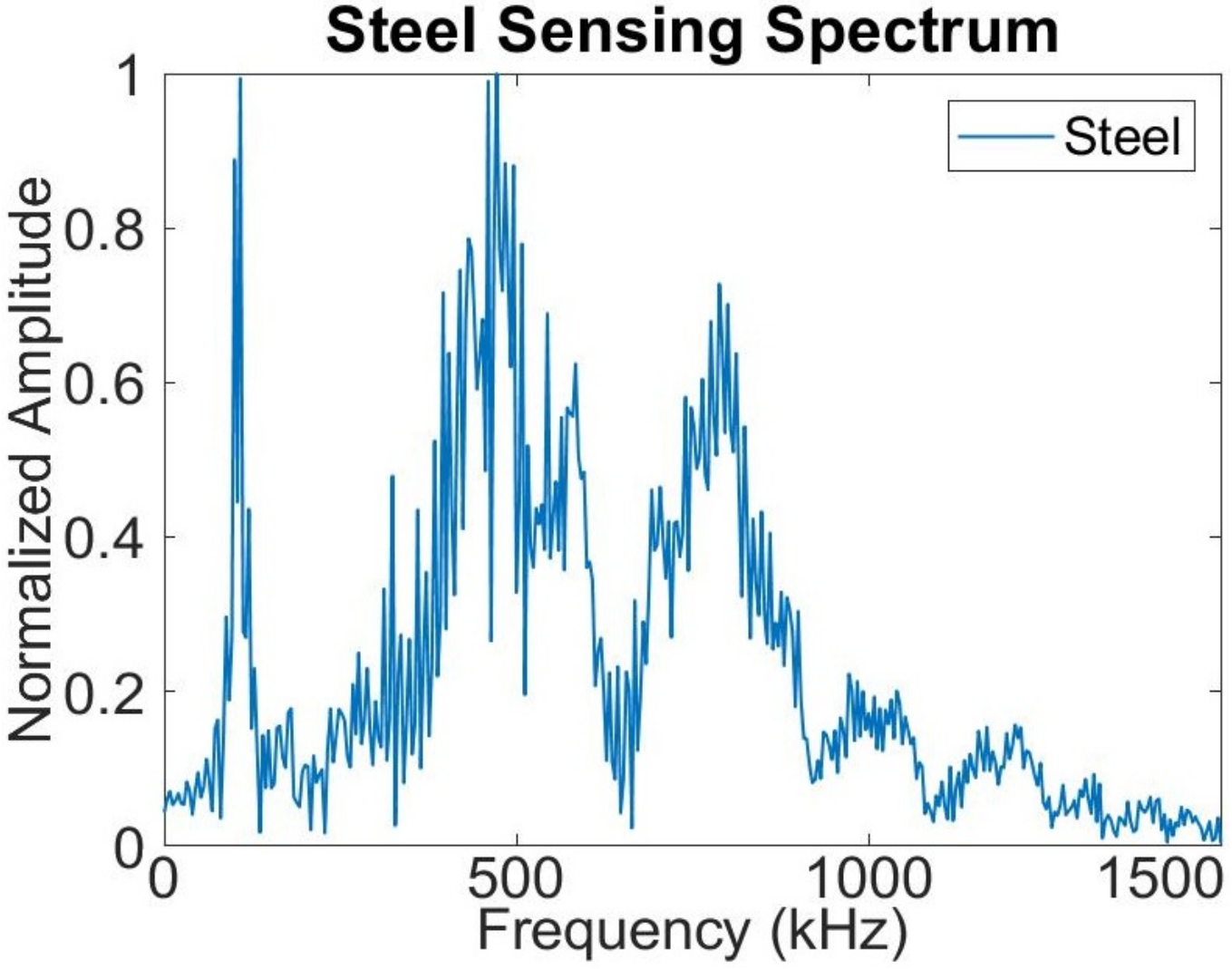}\label{fig:exp:spectrum-steel}}
    \caption{Representative PDM$^2$ acoustic spectra from five normal daily targets of (a) acrylic, (b) aluminum block, (c) paper, (d) rubber, and (e) steel.}
    \label{fig:exp:spectra-material}
\end{figure}

\begin{figure}[!htbp]
    \centering
    \subfloat[]{\includegraphics[height=1.26in]{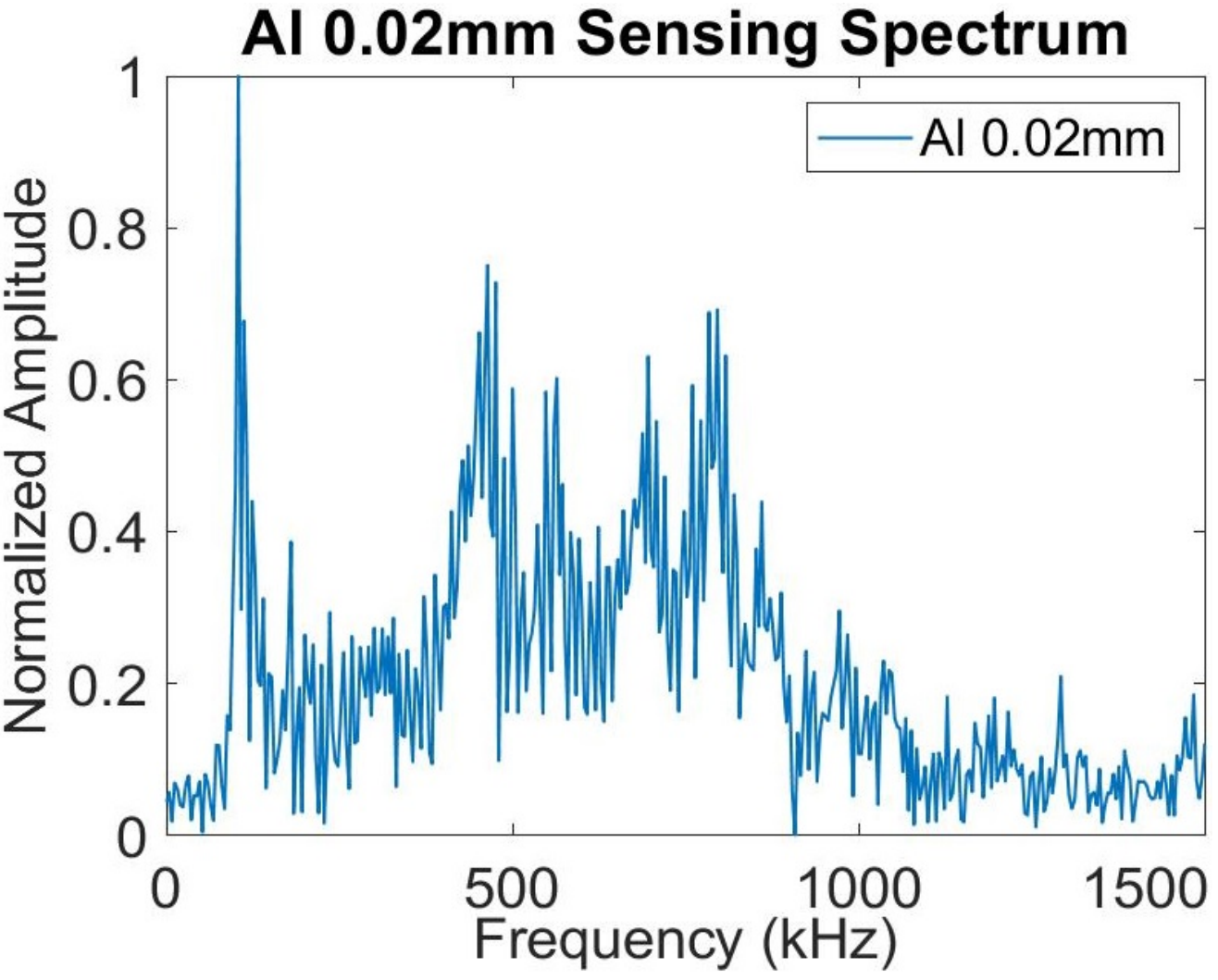}\label{fig:exp:spectrum-2}} \hspace*{.1in}
    \subfloat[]{\includegraphics[height=1.26in]{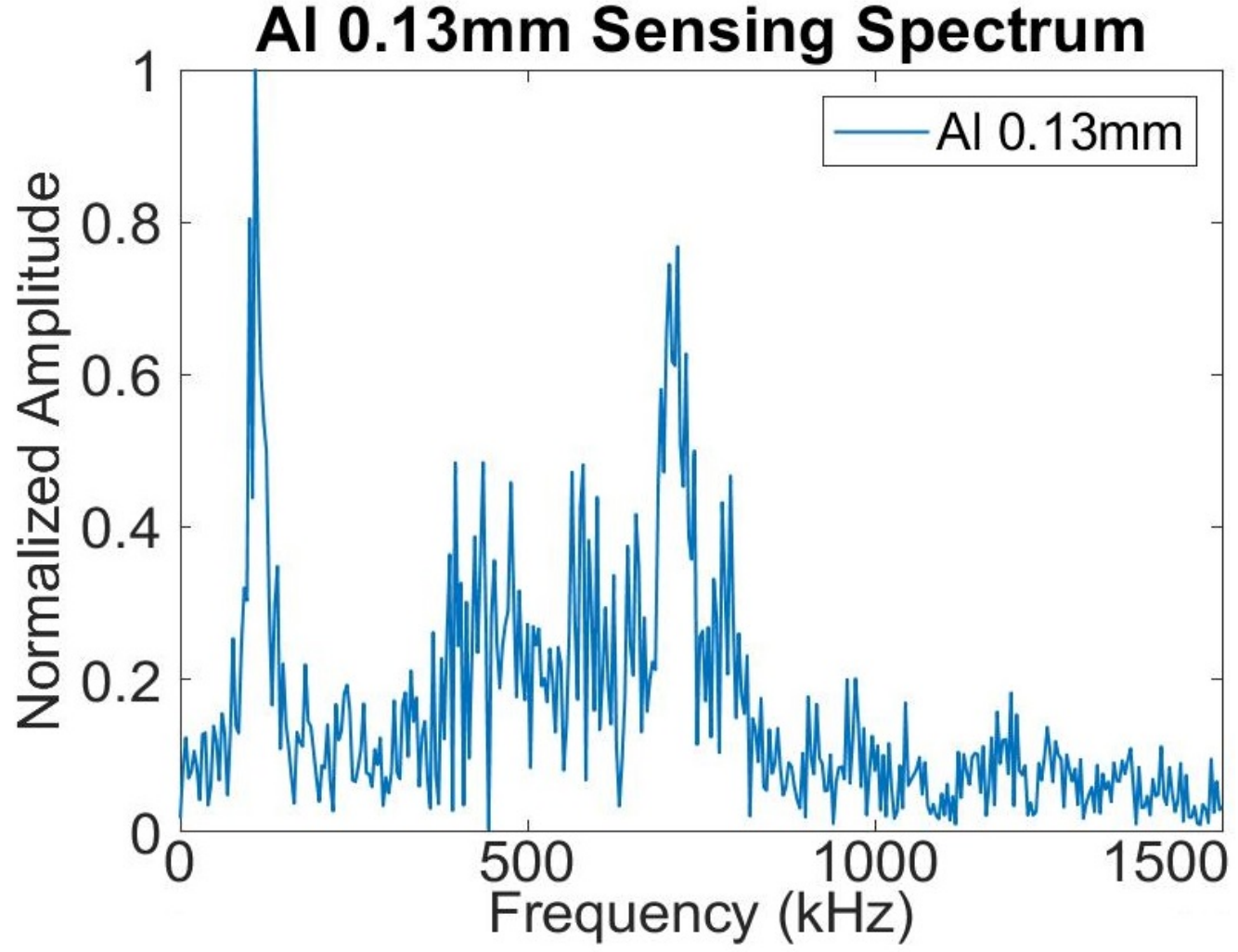}\label{fig:exp:spectrum-13}}\\
    {\includegraphics[height=1.26in]{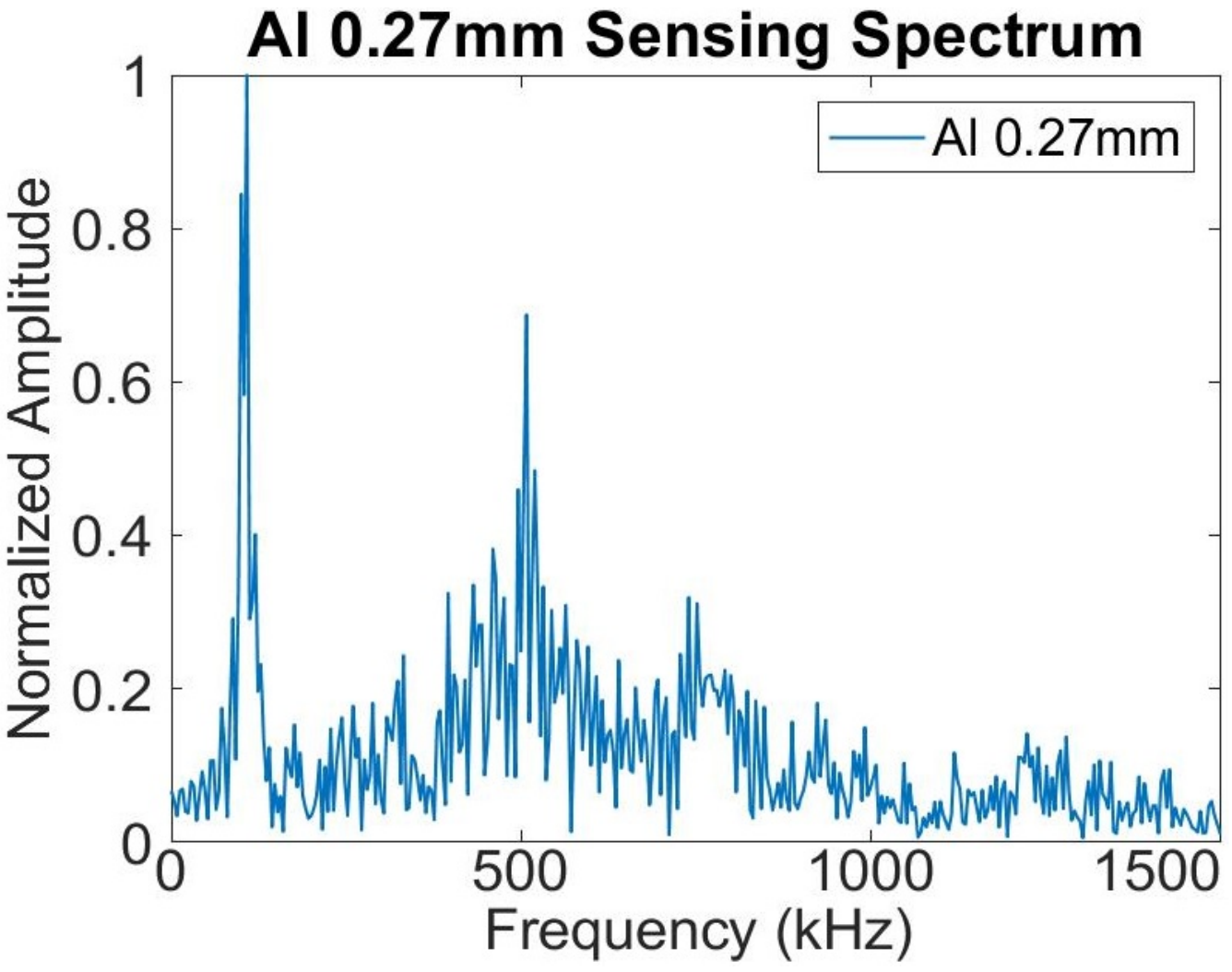}\label{fig:exp:spectrum-27}} \hspace*{.1in}
    \subfloat[]{\includegraphics[height=1.26in]{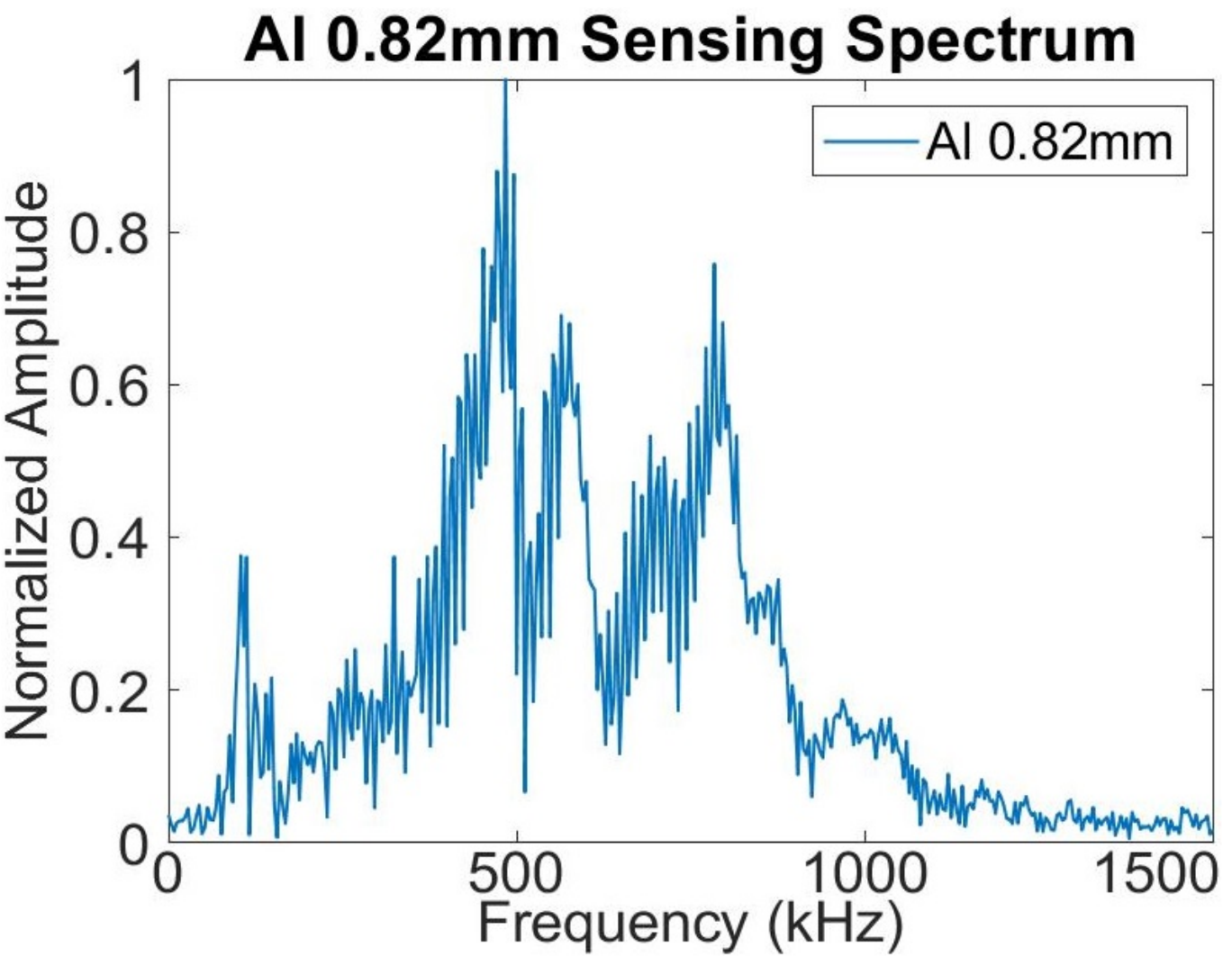}\label{fig:exp:spectrum-82}}\\
    {\includegraphics[height=1.26in]{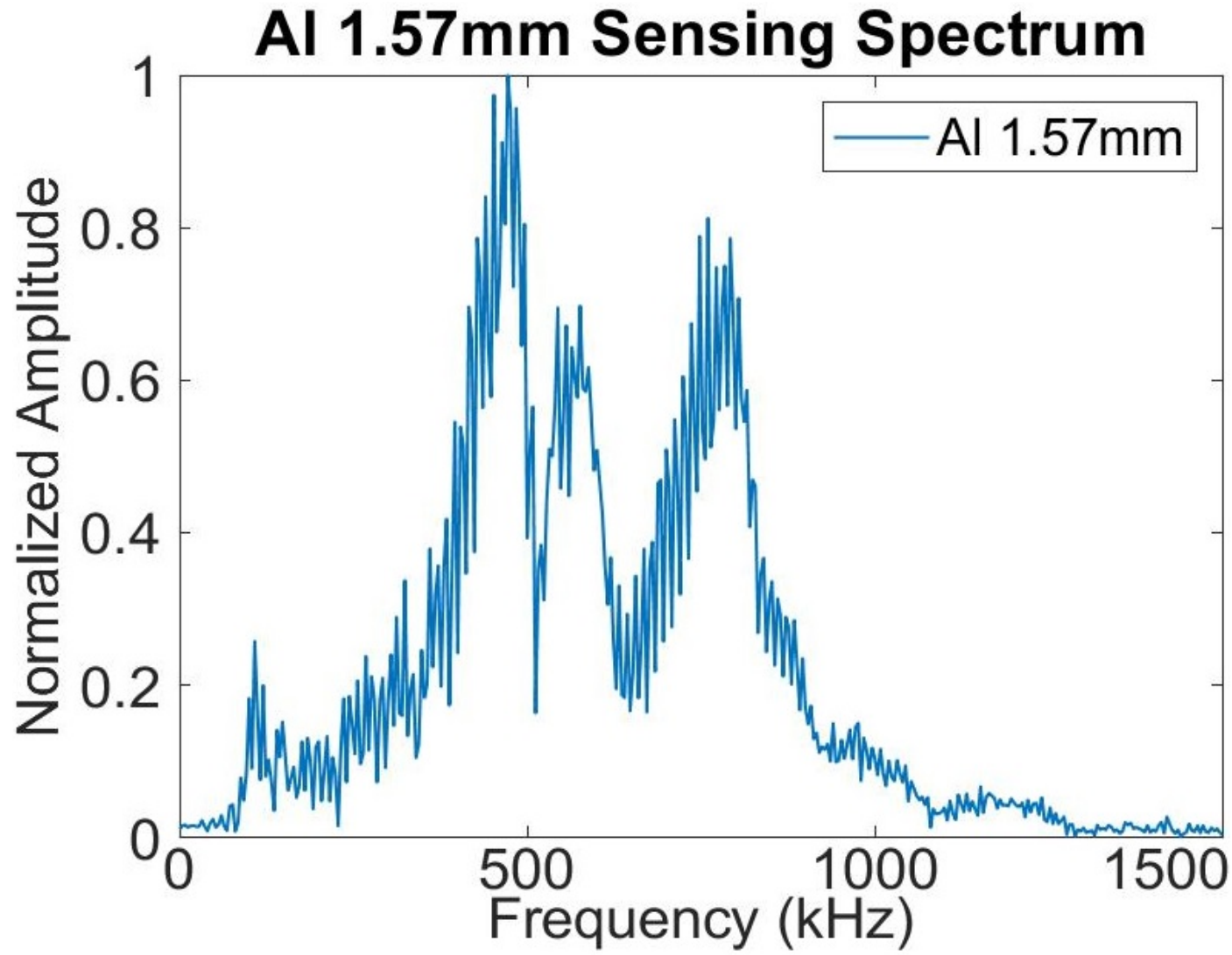}\label{fig:exp:spectrum-157}} \hspace*{.1in}
    \subfloat[]{\includegraphics[height=1.26in]{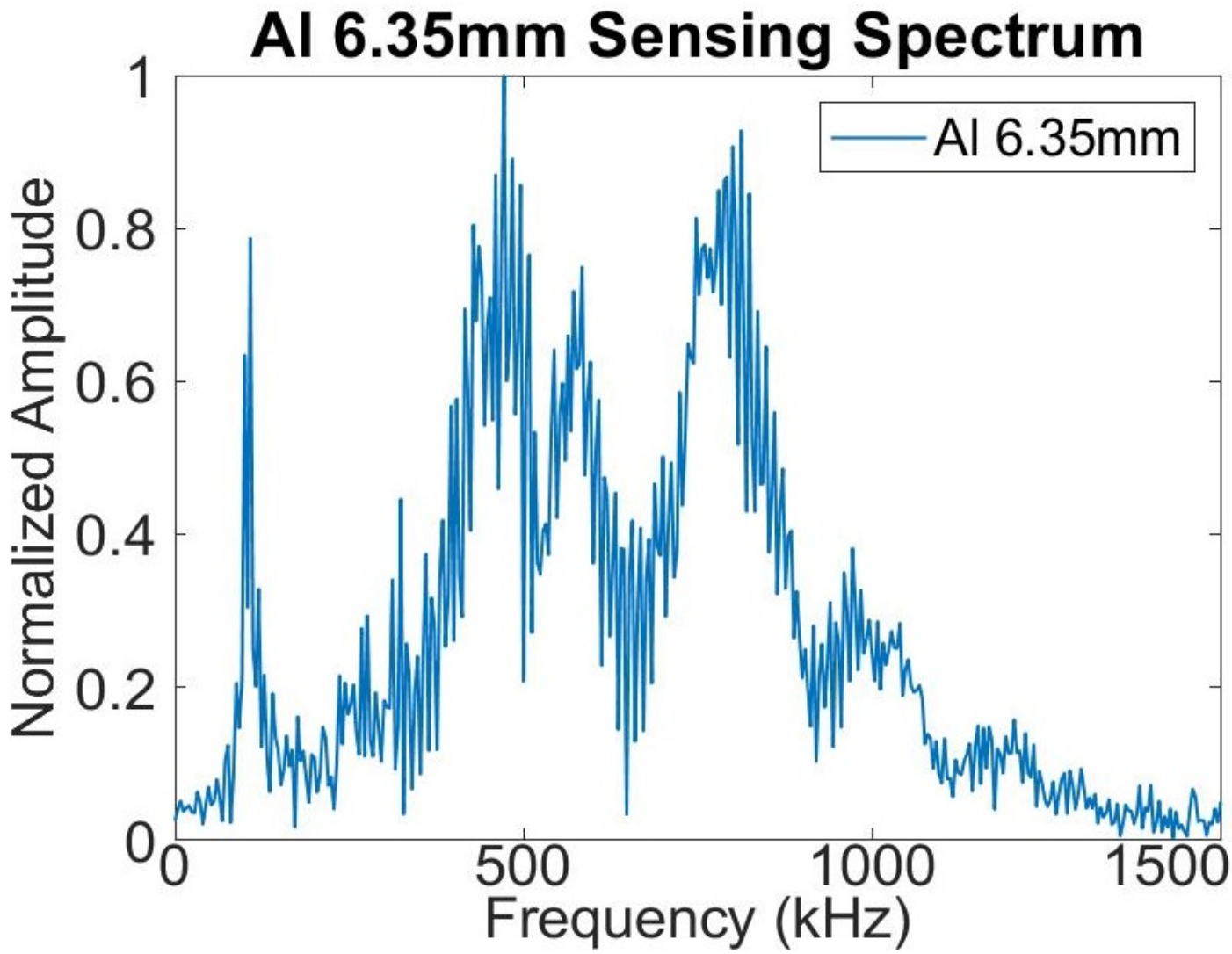}\label{fig:exp:spectrum-635}}\\
    \subfloat[]{\includegraphics[width=1.65in]{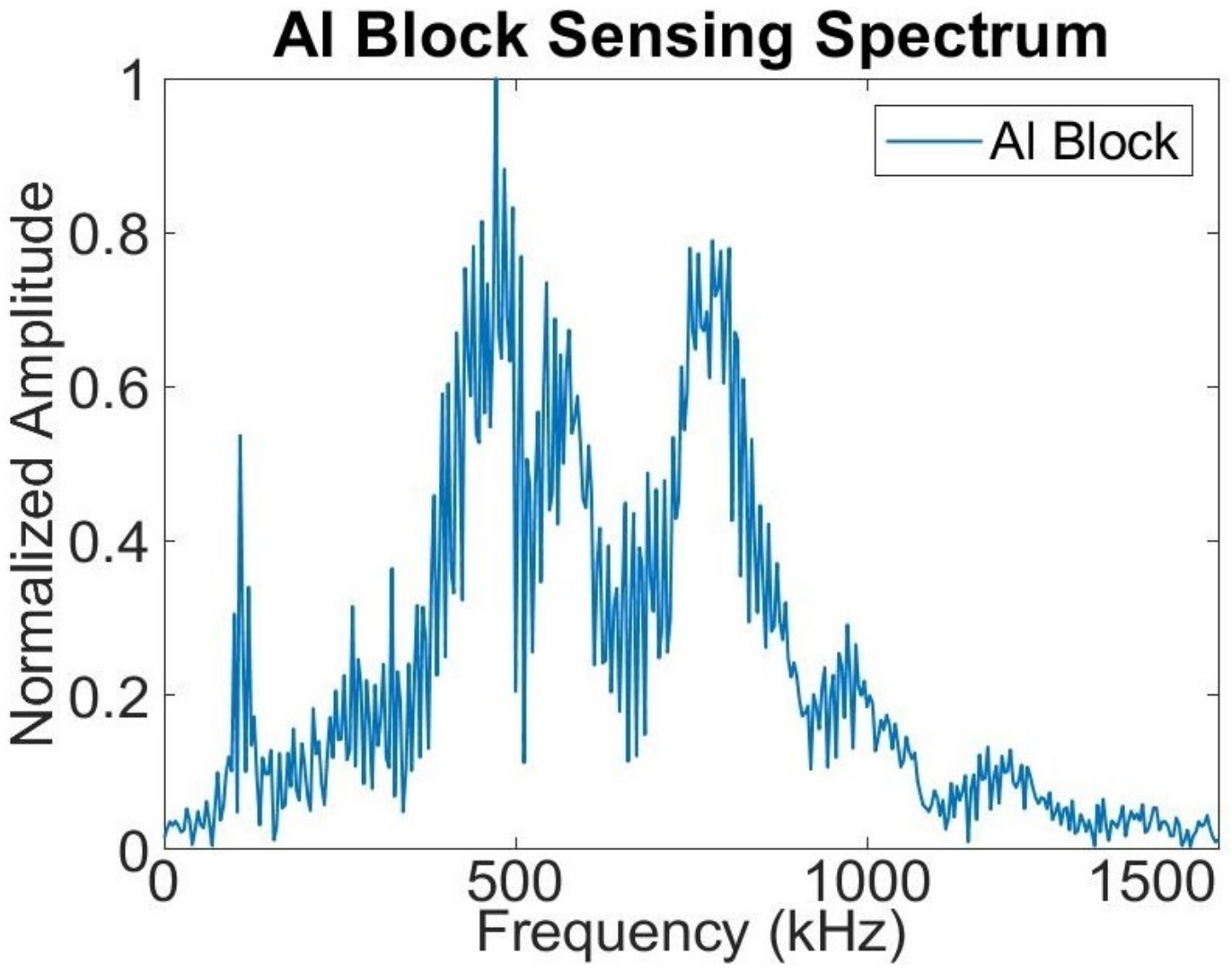}\label{fig:exp:spectrum-block}}
    \caption{Representative PDM$^2$ acoustic spectra from aluminum sheets with different thickness.}
    \label{fig:exp:spectra-thickness}
\end{figure}

\begin{figure}[!htbp]
    \centering
    \subfloat[]{\includegraphics[height=1.5in]{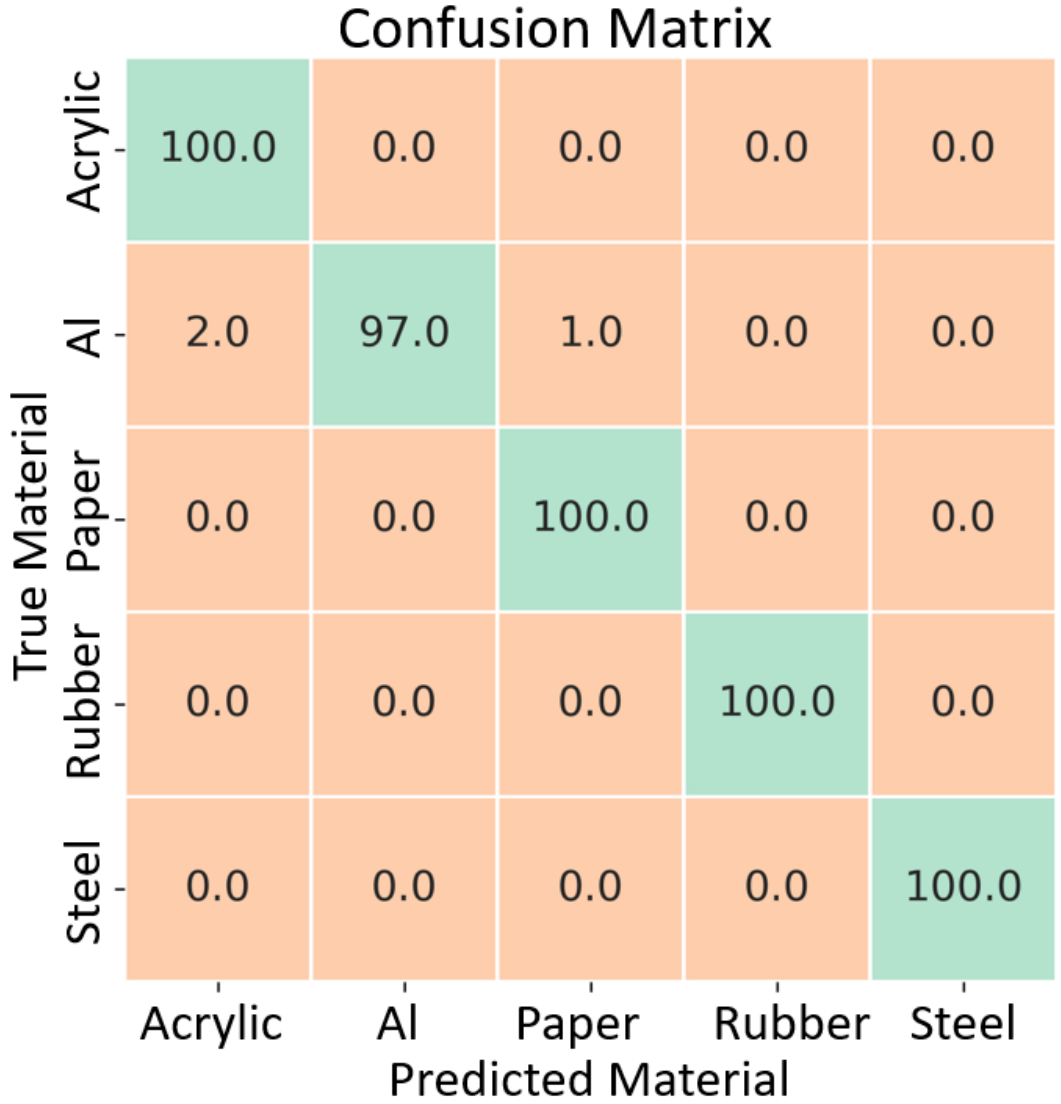}\label{fig:exp:BOSS-material}} \hspace*{.1in}
    \subfloat[]{\includegraphics[height=1.5in]{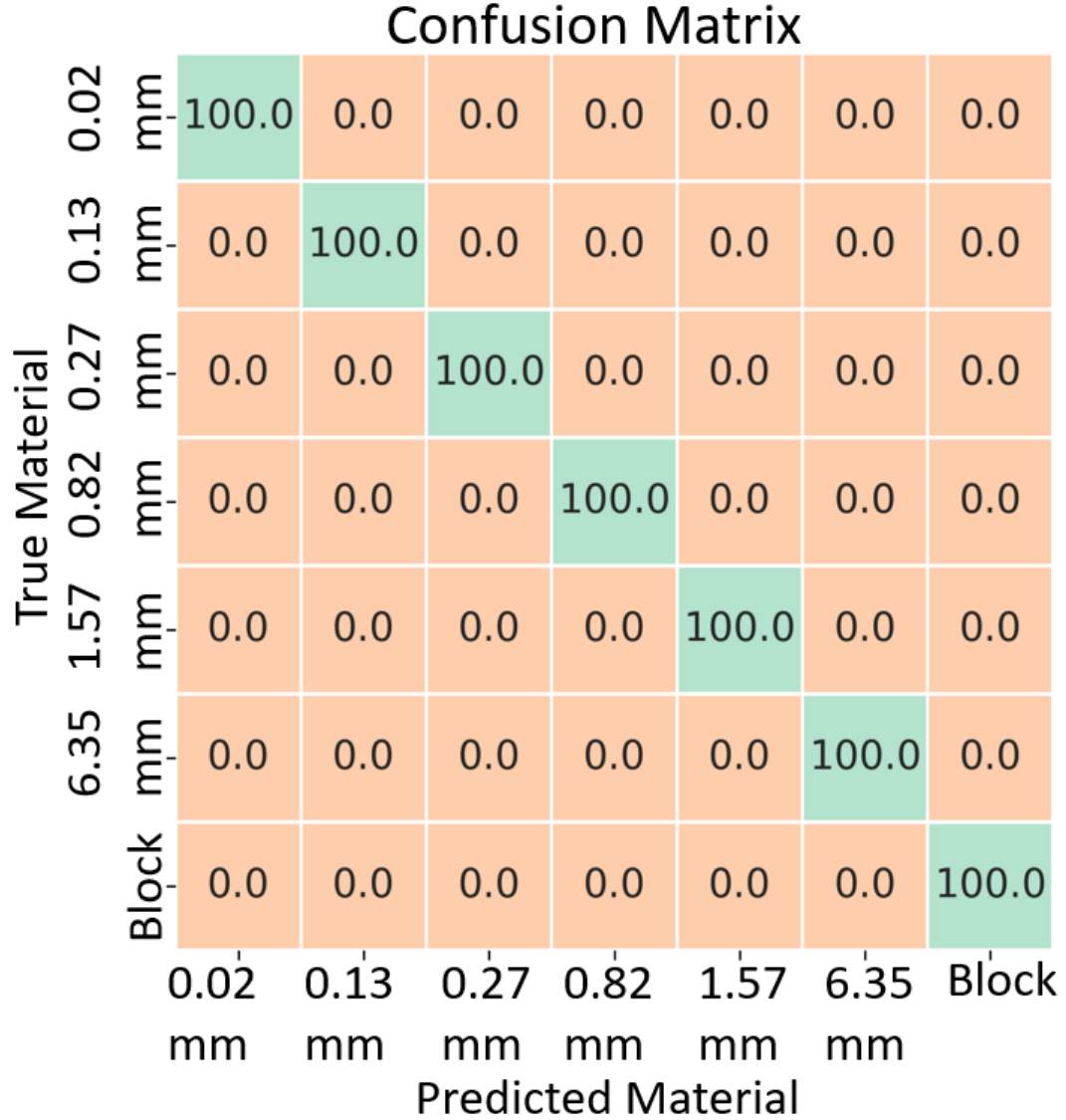}\label{fig:exp:BOSS-thickness}}
    \caption{Material and structure detection results: the averaged confusion matrix of (a) five normal daily targets and (b) aluminum sheets with different thickness.}
    \label{fig:exp:BOSS}
\end{figure}

The first group of targets represents normal daily objects, including 1.6-mm thick acrylic, aluminum block, thick paper from white milk box, black rubber, steel and aluminum sheets with different thicknesses are used as targets under the configuration depicted in Fig. \ref{fig:schematic-design}. Acrylic and paper were coated with black ink to avoid confusion with the challenging targets in the next section. The captured waveforms, including the US and / or OA signals, are similar to those in Fig. \ref{fig:representative-waveform-US-OA}, which carry the distinctive spectrum features of the targets. The representative PDM$^2$ acoustic spectra are shown in Figs. \ref{fig:exp:spectra-material} and \ref{fig:exp:spectra-thickness}. The confusion matrices output by the BOSS classifier (Fig. \ref{fig:exp:BOSS}) indicate $\geqslant$ 97\% the general precision of the material differentiation and 100\% the general precision of the thickness classification. Overall performance is satisfactory.

\subsubsection{Challenging Targets}
\begin{figure}[!htbp]
    \centering
    \subfloat[]{\includegraphics[height=.63in]{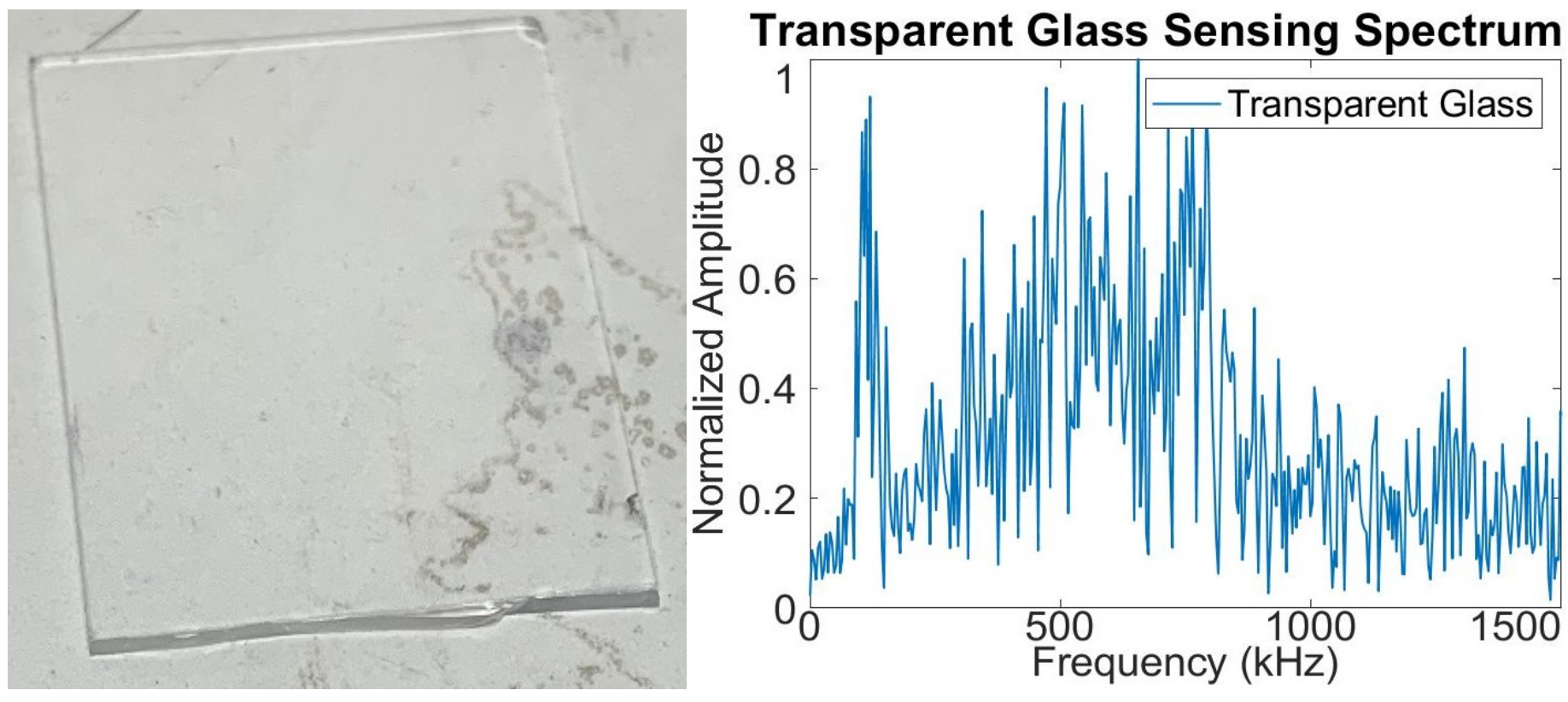}\label{fig:exp:spectrum-a}} 
    \subfloat[]{\includegraphics[height=.63in]{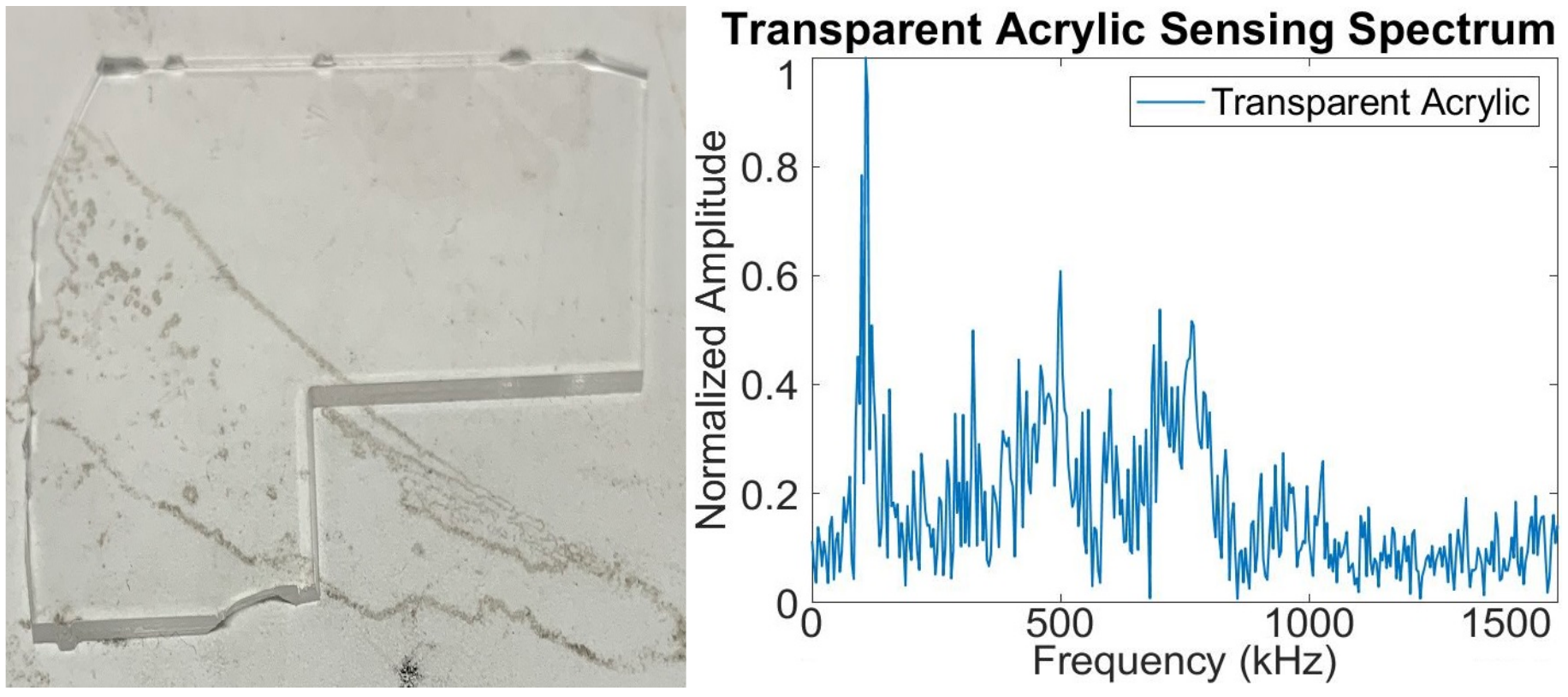}\label{fig:exp:spectrum-b}}
    
    \subfloat[]{\includegraphics[height=.63in]{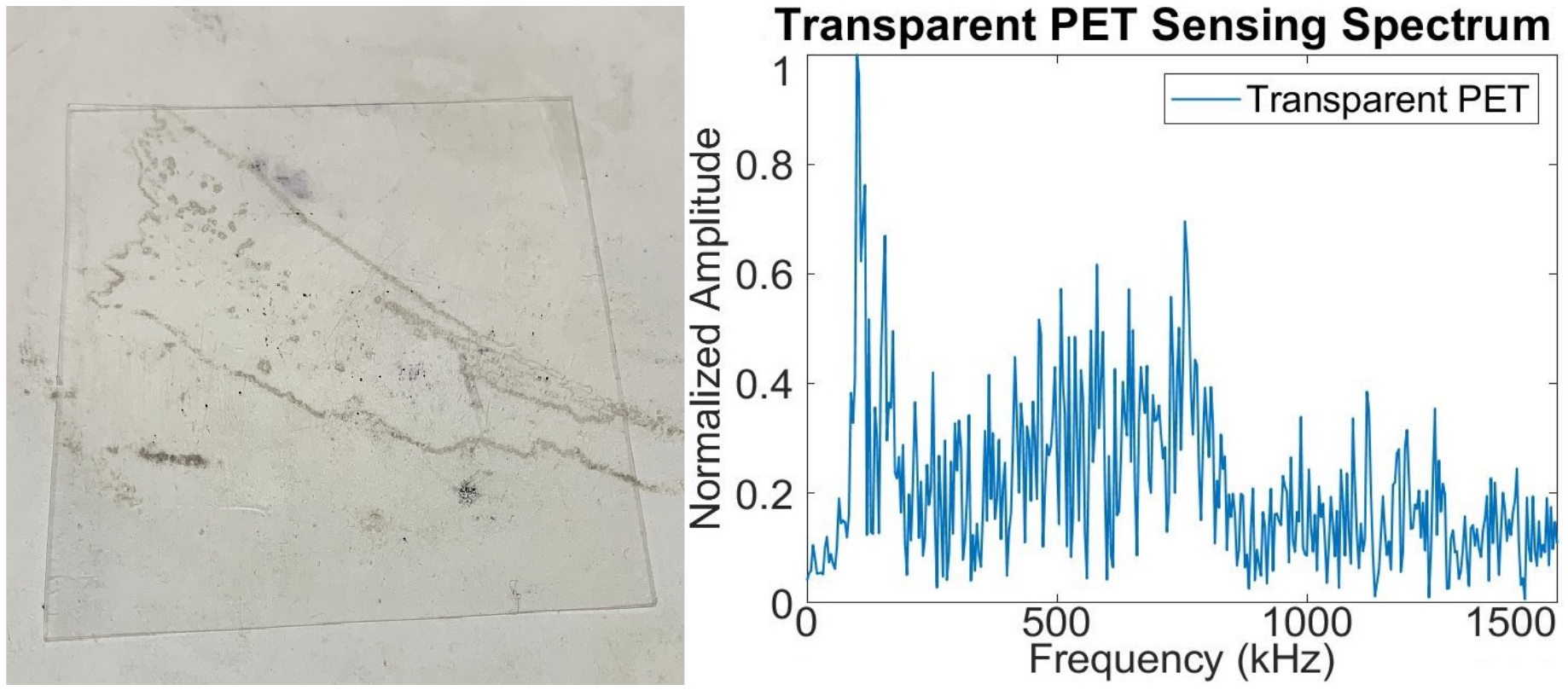}\label{fig:exp:spectrum-c}} 
    \subfloat[]{\includegraphics[height=.63in]{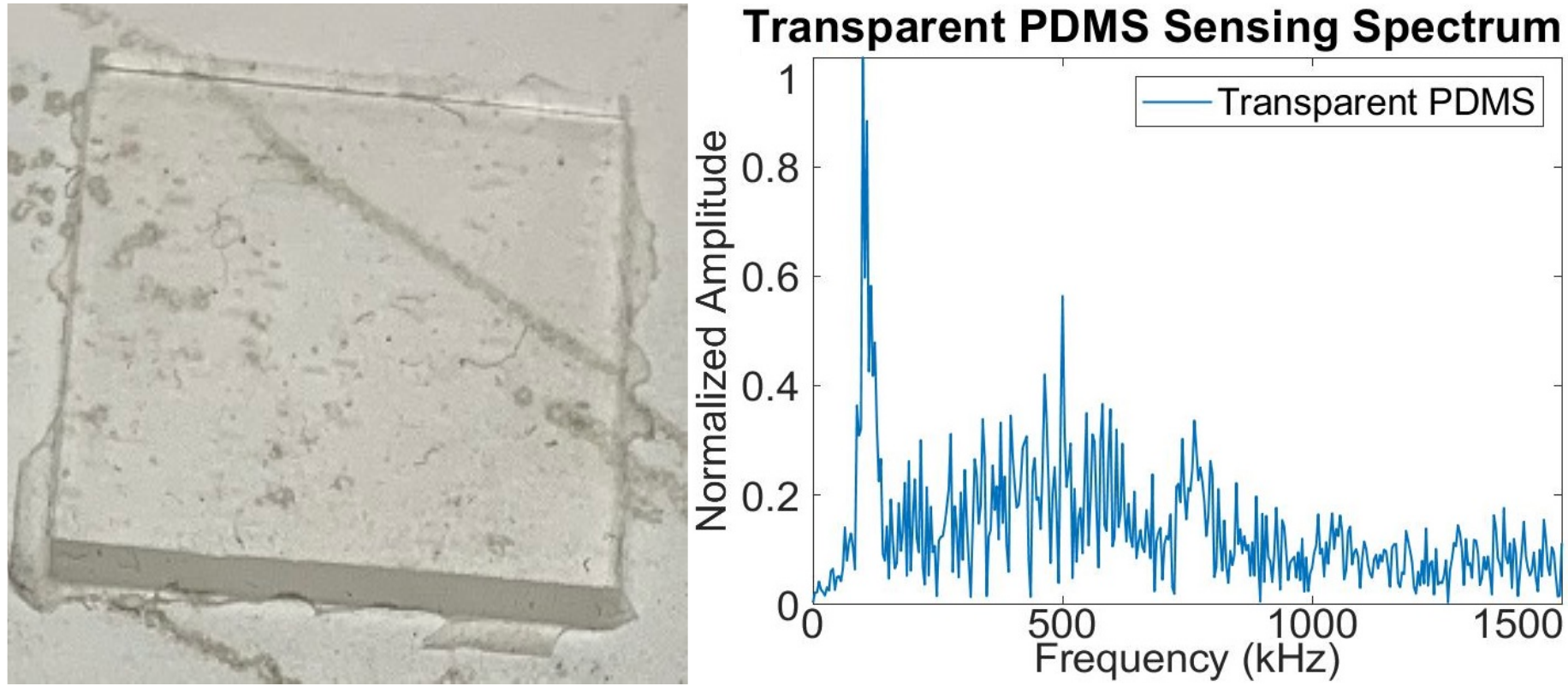}\label{fig:exp:spectrum-d}}
    
    \subfloat[]{\includegraphics[height=.63in]{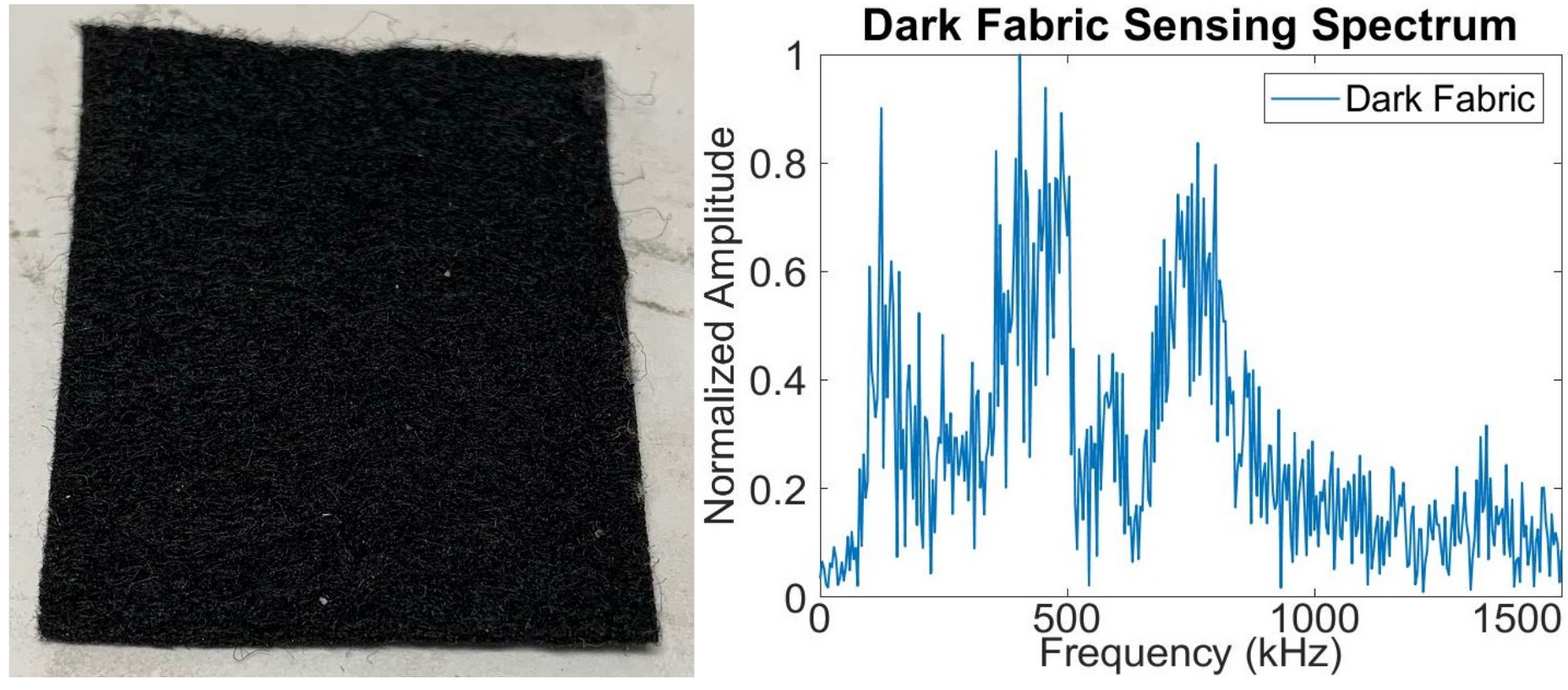}\label{fig:exp:spectrum-e}} 
    \subfloat[]{\includegraphics[height=.63in]{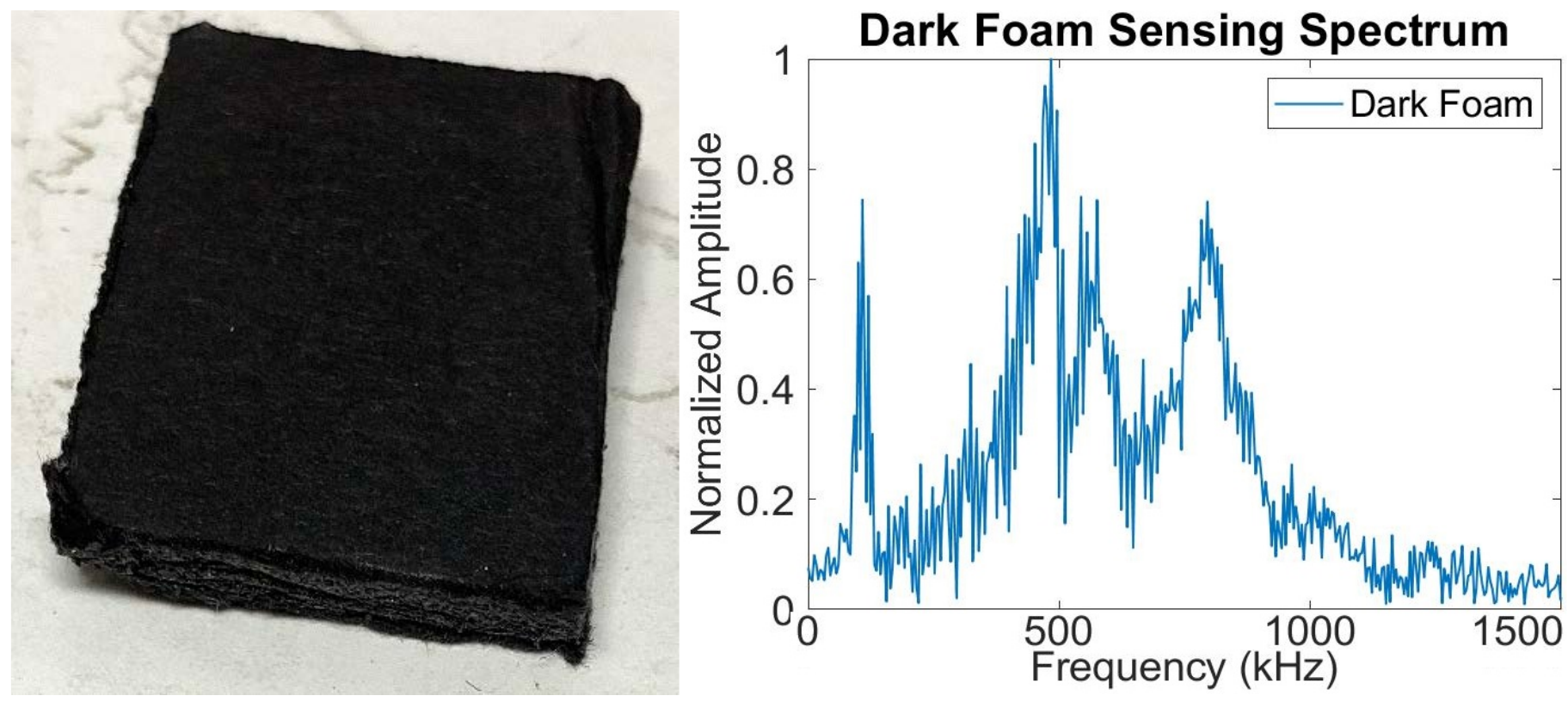}\label{fig:exp:spectrum-f}}
    
    \subfloat[]{\includegraphics[height=.63in]{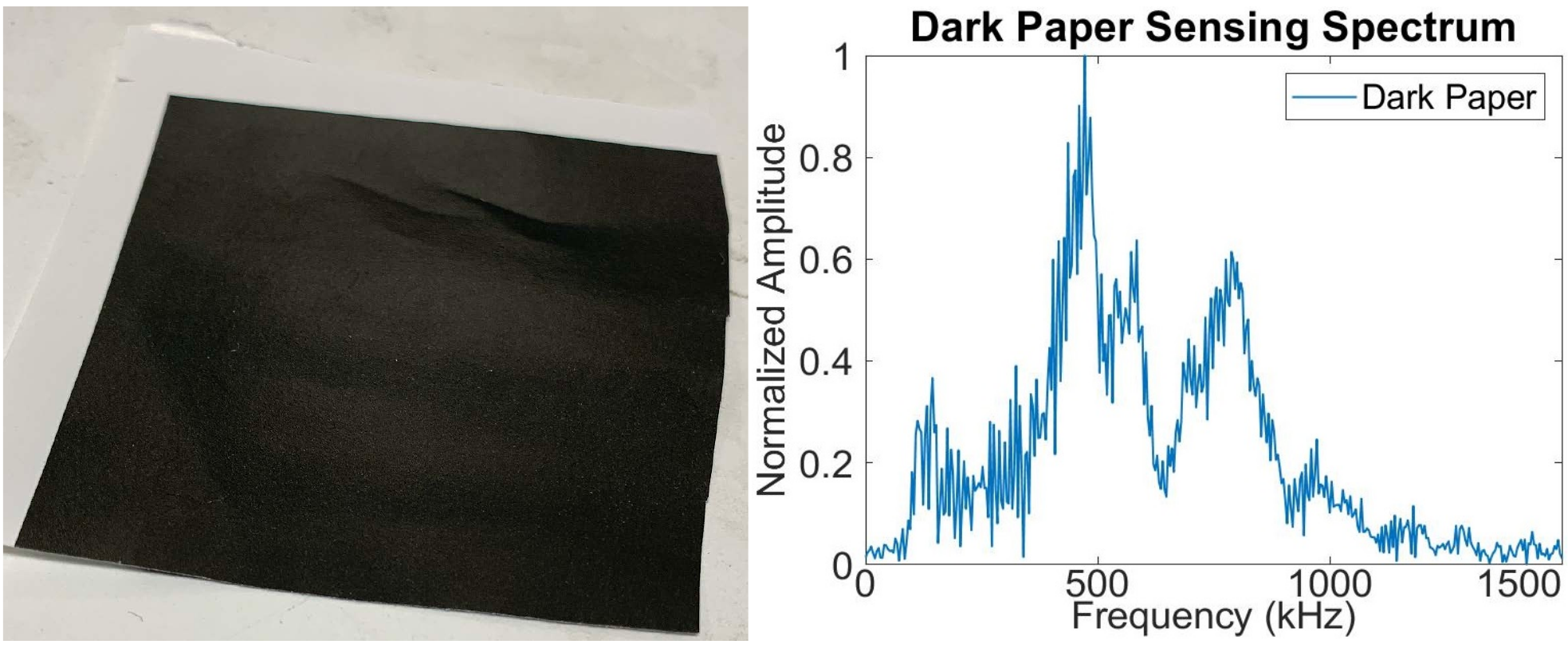}\label{fig:exp:spectrum-g}} 
    \subfloat[]{\includegraphics[height=.63in]{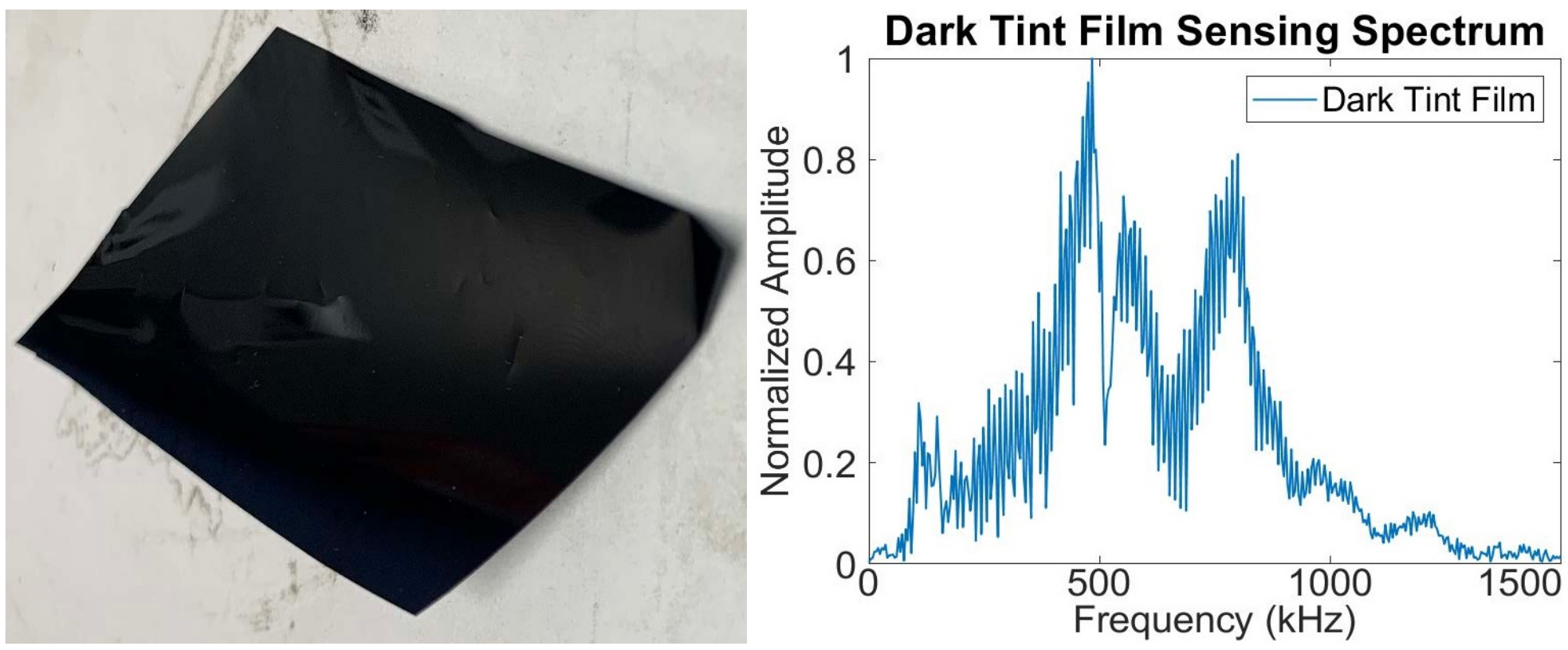}\label{fig:exp:spectrum-h}}
    
    \subfloat[]{\includegraphics[width=1.8in]{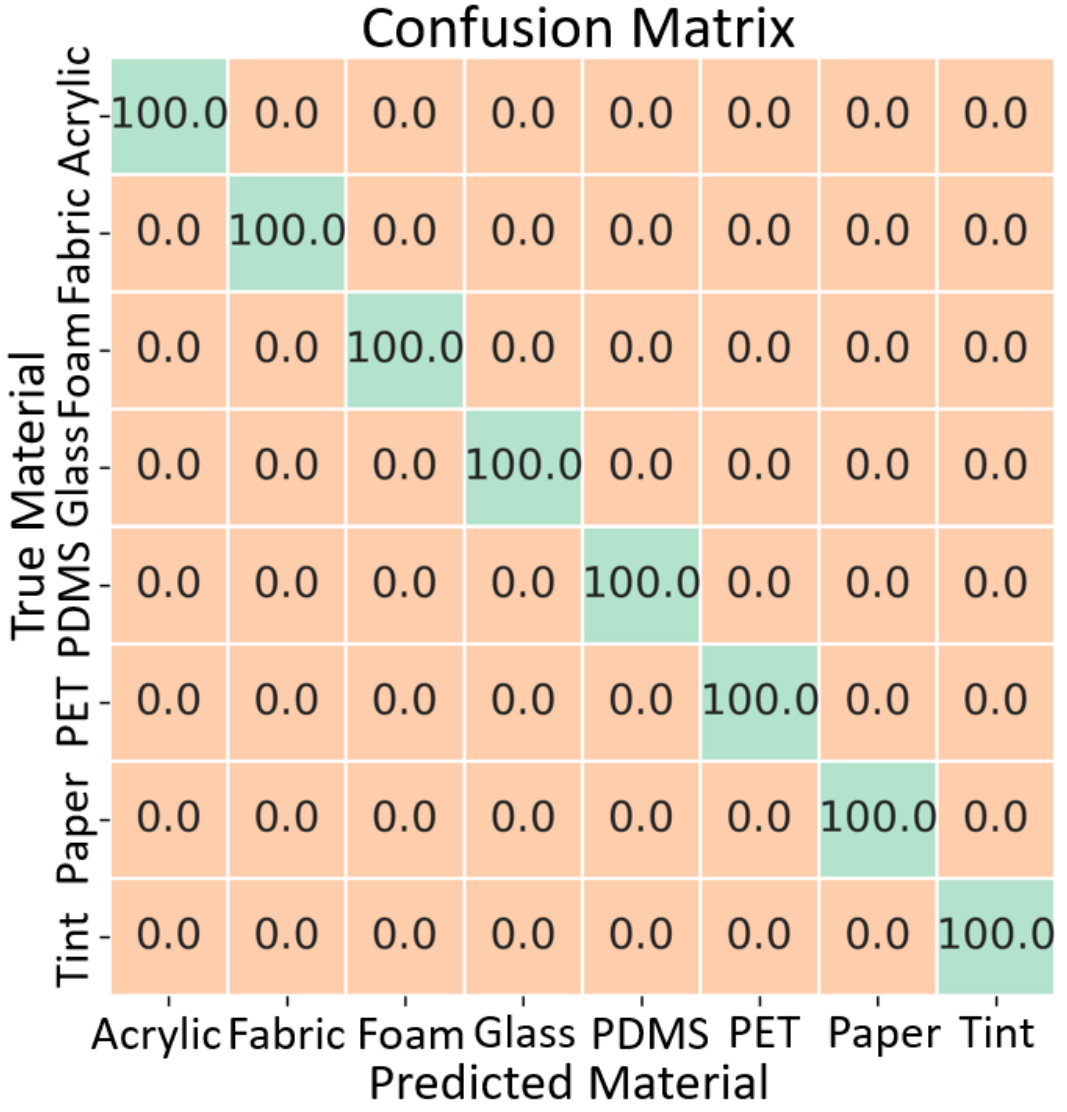}\label{fig:exp:spectrum-i}}
    \caption{PDM$^2$ material and struture sensing performance under OACTs: (a-d) optically-transparent targets of glass, acrylic, PET, PDMS with thicknesses around 1.0 mm, 1.60 mm, 0.11 mm, and 1.5 mm, respectively, and (e-h) low acoustic reflectivity targets consisting of dark\&thin/porous targets of fabric, foam, paper, window tint film with thicknesses around 2.0 mm, 8.0 mm, 0.10 mm, 0.06 mm, respectively. (i) The average confusion matrix of the eight OACTs.}
    \label{fig:exp:OACTs}
\end{figure}

To further investigate the sensing capabilities of the new PDM$^2$ sensor, eight optically and acoustically challenged targets (OACTs) are tested (Fig. \ref{fig:exp:OACTs}), including four optically transparent targets of glass, acrylic, polyethylene terephthalate (PET), polydimethylsiloxane (PDMS) (Figs. \ref{fig:exp:spectrum-a}, \ref{fig:exp:spectrum-b}, \ref{fig:exp:spectrum-c}, and \ref{fig:exp:spectrum-d}) with little optical absorptivity and therefore weak OA response, and four low acoustic reflectivity targets including dark\&thin/porous targets of fabric, foam, thin paper, 95\%-transmittance window tint film (Figs. \ref{fig:exp:spectrum-e}, \ref{fig:exp:spectrum-f}, \ref{fig:exp:spectrum-g}, and \ref{fig:exp:spectrum-h}). To compensate for the difference in target thickness, the height of the target is adjusted until the top of the target is at the focus of the parabolic mirror using the scanning configuration in Fig. \ref{fig:schematic-design}. The confusion matrix given by the BOSS classifier indicates an overall accuracy of 100\% for all eight OACTs (Fig. \ref{fig:exp:spectrum-i}). The exciting capabilities have not been seen in other sensors. 

\subsection{Calibration Experiments}\label{ssc:cali-exp}
After ranging and signal processing algorithm have been validated, we now test our calibration algorithm in experiments.
\subsubsection{Calibration Experiment Data and Results}
The calibration dataset is collected following the procedure described in Sec.~\ref{subsec:Procedure}. In the dataset, 4 tip center points are extracted from 70 tip point measurements, and 14 edge center points are extracted from 88 edge point measurements in two edge frame scans,  between which the rotation angle is $\theta_{01}=\pi$. Using the collected data set, the calibration parameters are estimated with the algorithm mentioned in Sec.~\ref{subsec:Algorithm}, and their values are shown in Tab.~\ref{table:paramEstimation}.
\begin{table}[!htbp]
	\centering
	\caption{Estimated Calibration Parameters}
	\label{table:paramEstimation}
	\begin{tabular}{c l}
		\hline
		$\mathbf{v}$  & [~~ 0.0656; ~~~ 0.9955; \ -0.0678] \\
		$\mathbf{n}$  & [~ -0.0007; ~~~ 0.0022; \ ~0.9999] \\
		$\mathrm{X}_{\mathrm{R}}$ (mm) & [~~~235.21; \  ~~~288.17; \ ~~~~0.00] \\\hline
	\end{tabular}
\end{table}

\subsubsection{Calibration Result Validation}
The estimated calibration parameters and the codirectional alignment of the OA and US beams are validated by reconstructing an aluminum block with a known shape. The reconstructed points are compared with the ground-truth measurements from the Vernier caliper. The error metric $e = d(\mathrm{X}_{ij},\mathbf{E}_i)$ is defined based on the Euclidean distance function $d$ between a reconstructed point $\mathrm{X}_{ij}$ and its corresponding contour line of ground truth $\mathbf{E}_i$. 

\begin{figure}[!ht] \vspace{-0.1in}
    \centering
    \subfloat[]{\raisebox{1mm}{\includegraphics[height=0.85in ]{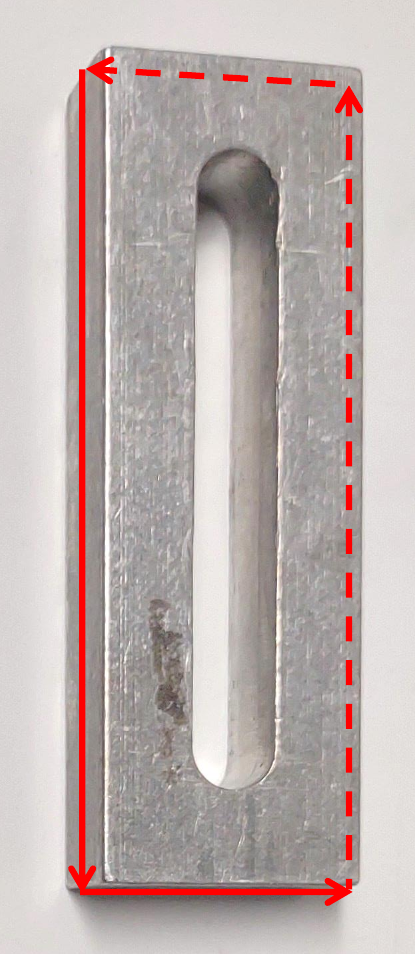}}\label{fig_AlSlot}} \hspace{.3in}
    \subfloat[]{\includegraphics[height=0.9in]{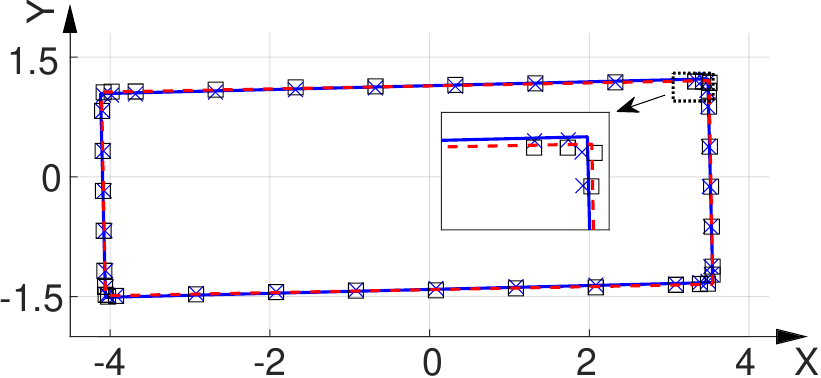}\label{fig_AlSlot_c}}
    \caption{Aluminum block contour reconstruction for validating calibration process. (a) Aluminum block with scanning path. (b) Reconstructed Contour.  The unit of X-Y axes is centimeter. Symbols `$\square$' and `$\times$' represent OA and US scanned points, respectively. The red and blue boxes represent the OA and US fitted contours, respectively.} \label{fig:validation}
\end{figure}

The reconstruction error and standard deviation of OA and US are $\left(0.06 \pm 0.06\right)$ and $\left(0.15 \pm 0.11\right)$ mm, respectively. The reconstruction result using both modalities is shown in Fig. \ref{fig:validation}.

The results show that both the OA and the US reconstruction achieved a satisfactory level of accuracy. These results verify the calibration result and indicate that the two co-directional signal beams are aligned well.

\subsection{Object Contour Scanning and Reconstruction}\label{ssc:reconstruction-exp}

The combined ranging and calibration performance can be tested in the reconstruction of the object contour based on PDM$^2$ sensor scans using our object scanning system. We have tested the capabilities of the PDM$^2$ sensor on six household items.

\begin{figure}[!ht]
    \centering
    \subfloat[]{\includegraphics[width=0.55in ]{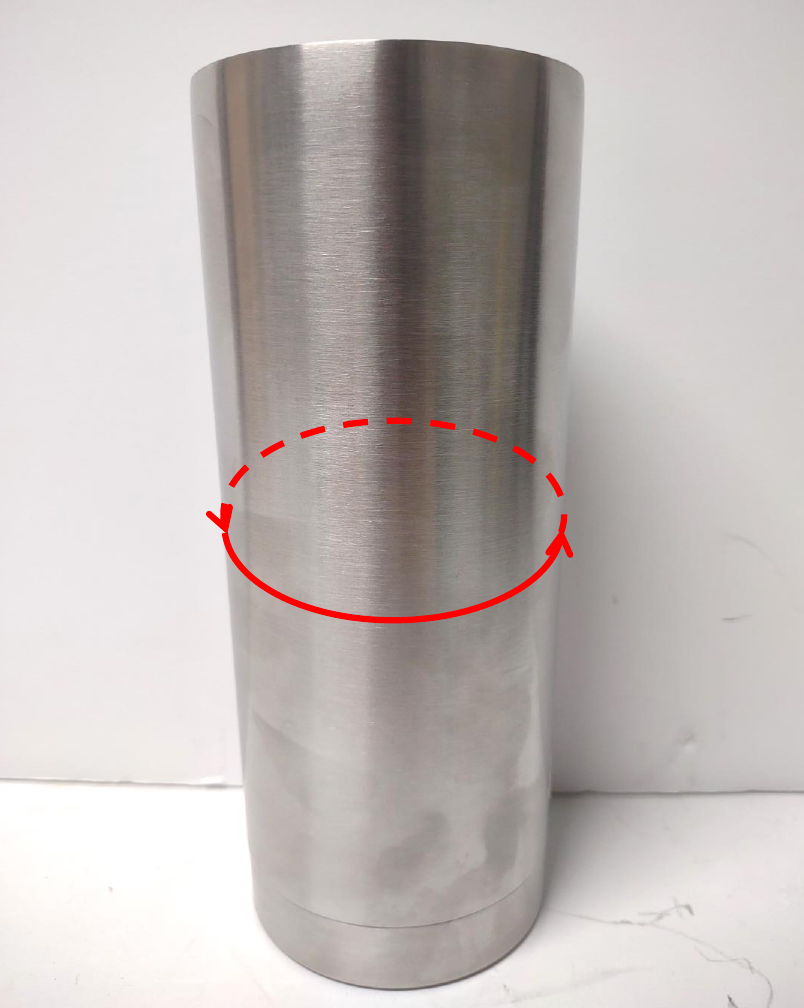}\label{fig_SteelBottle}} 
    \subfloat[]{\includegraphics[width=0.55in ]{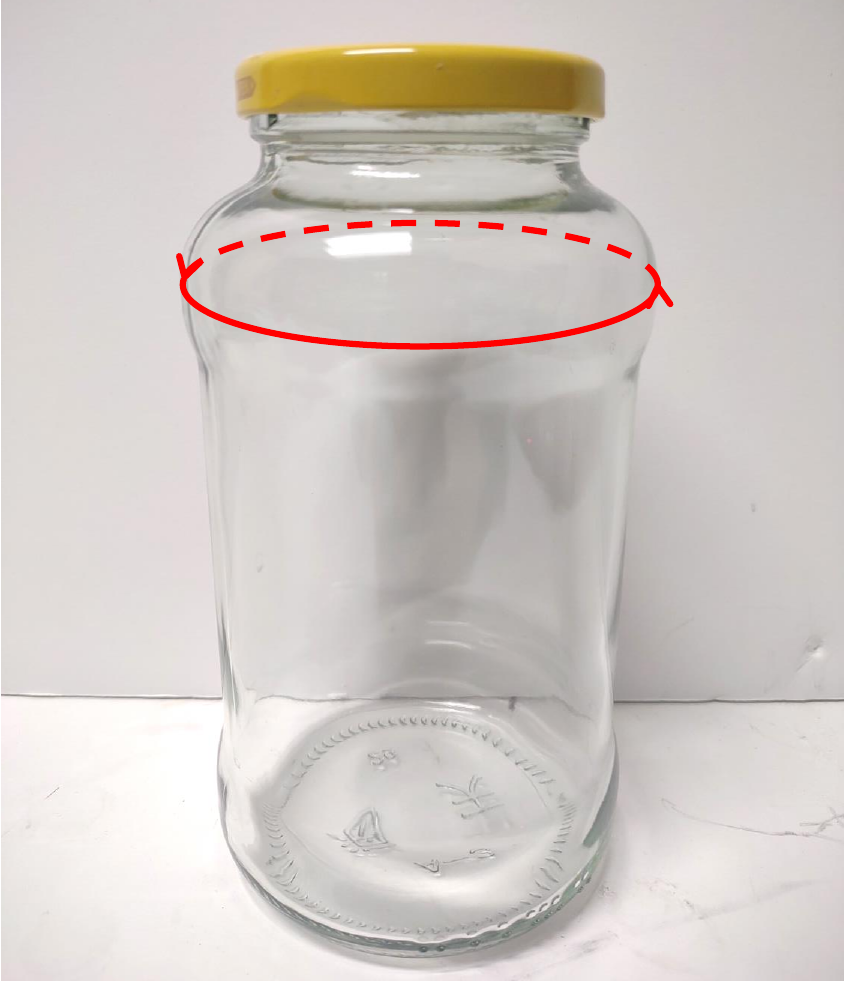}\label{fig_GlassBottle}} 
    \subfloat[]{\includegraphics[width=0.55in ]{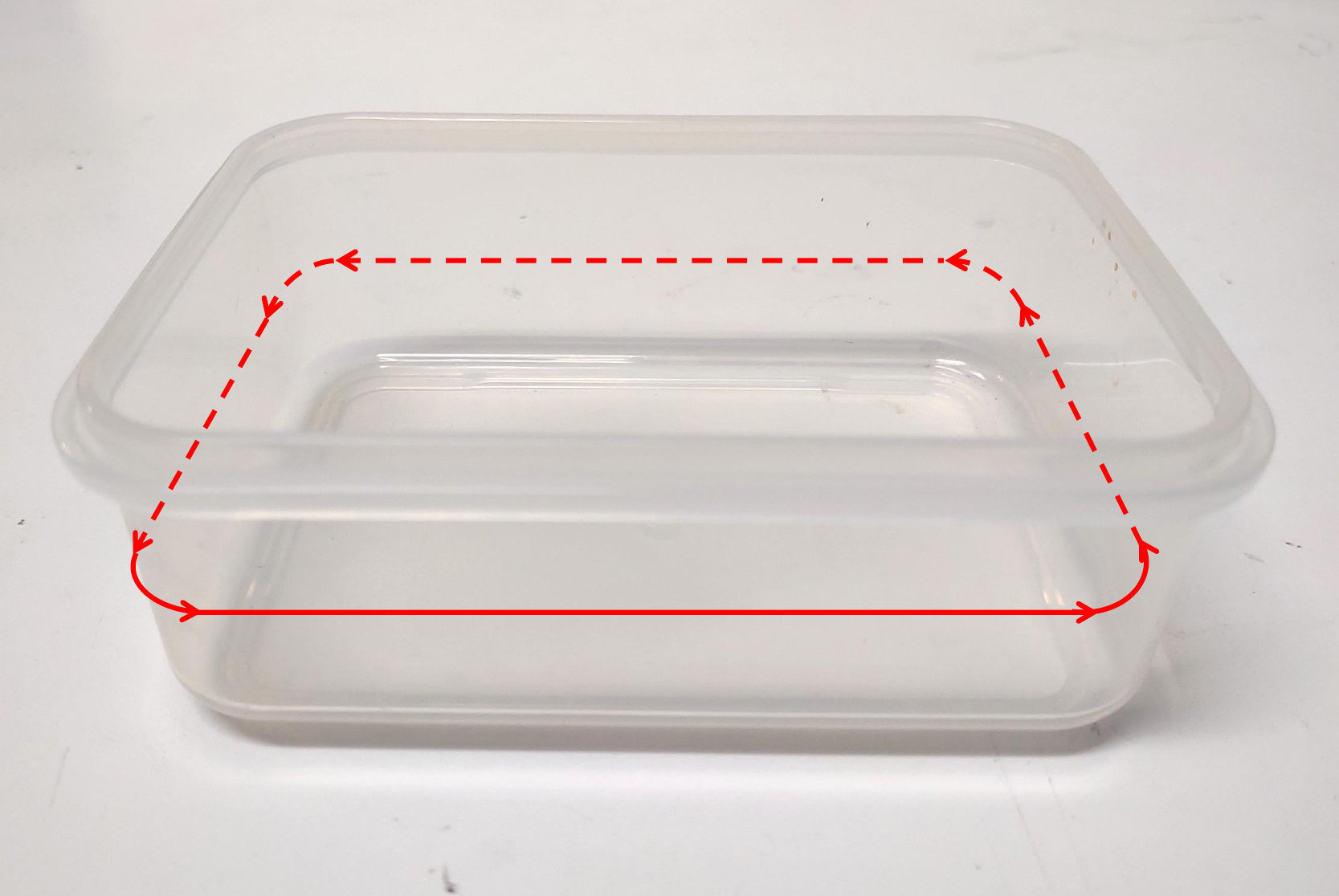}\label{fig_PlasticBox}} 
    \subfloat[]{\includegraphics[width=0.55in ]{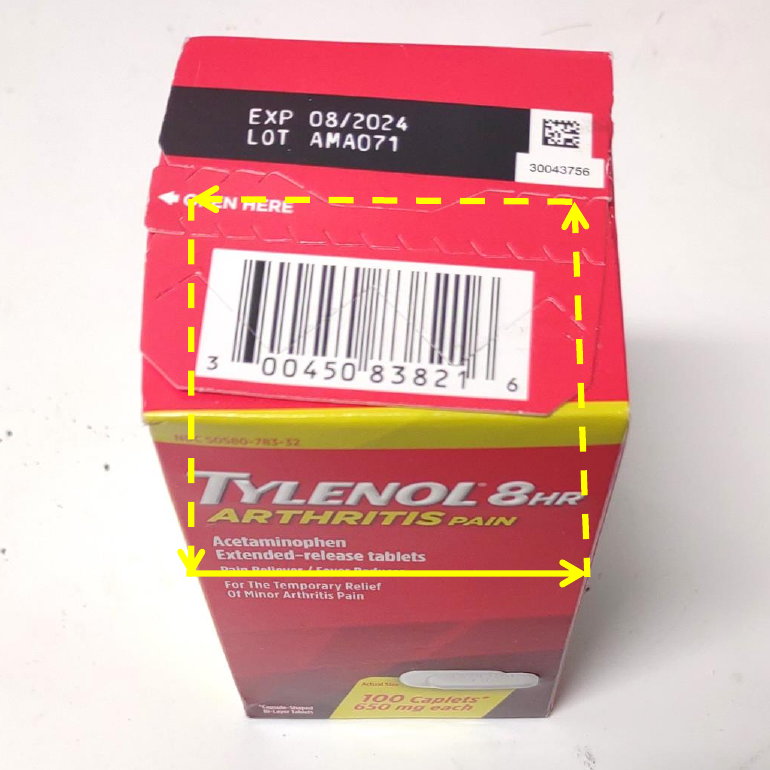}\label{fig_procedure_1}}
    \subfloat[]{\includegraphics[width=0.55in ]{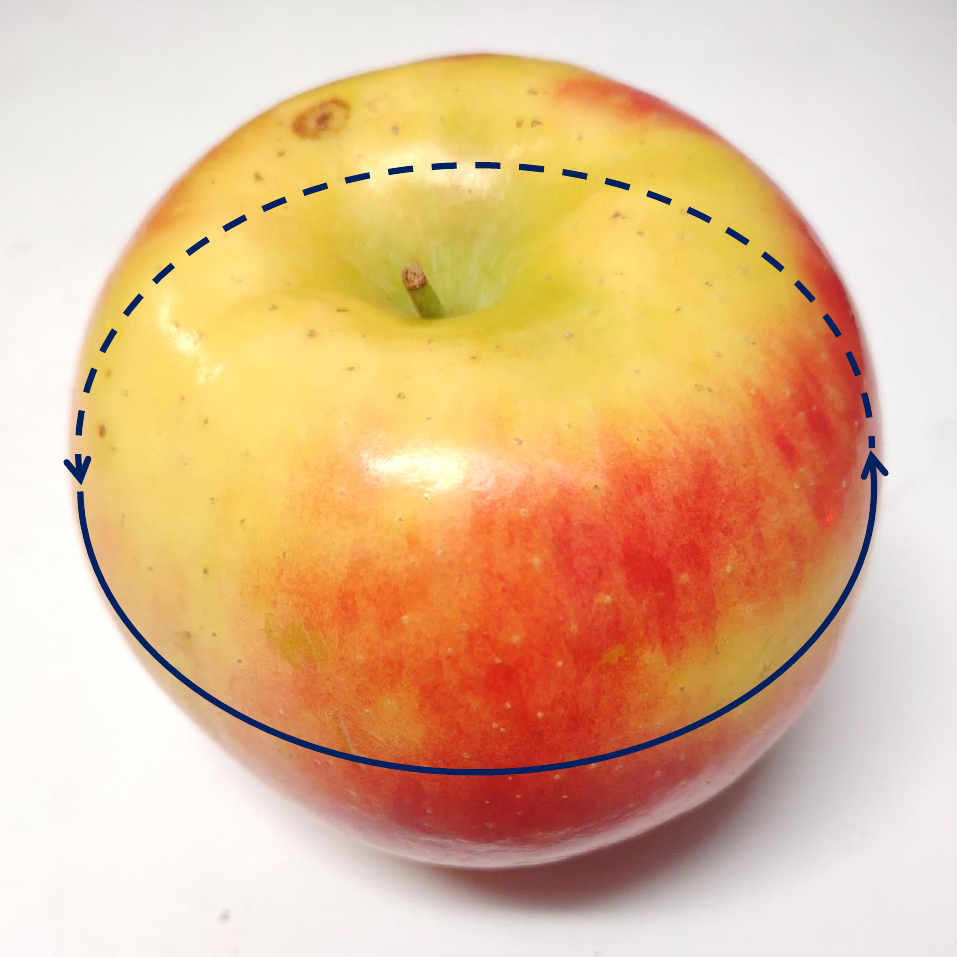}\label{fig_Apple}}
    \subfloat[]{\includegraphics[width=0.55in ]{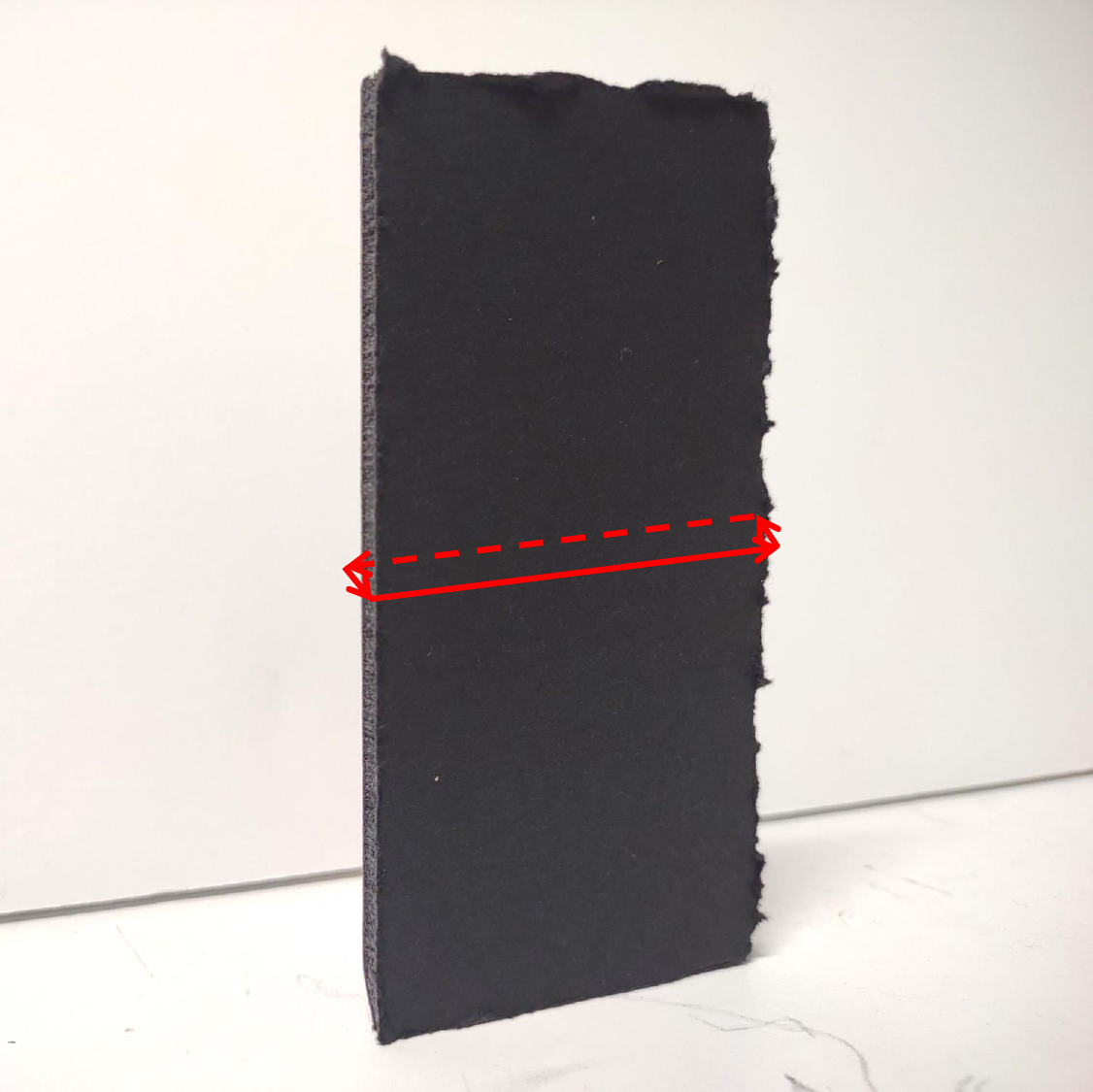}\label{fig_Foam}} \\
    
    \subfloat[]{\includegraphics[width=0.55in]{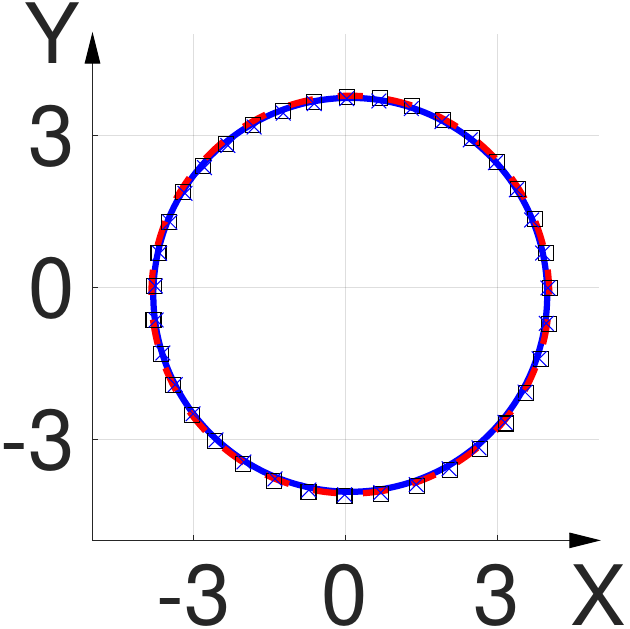}\label{fig_PlasticBox_c}}    
    \subfloat[]{\includegraphics[width=0.55in]{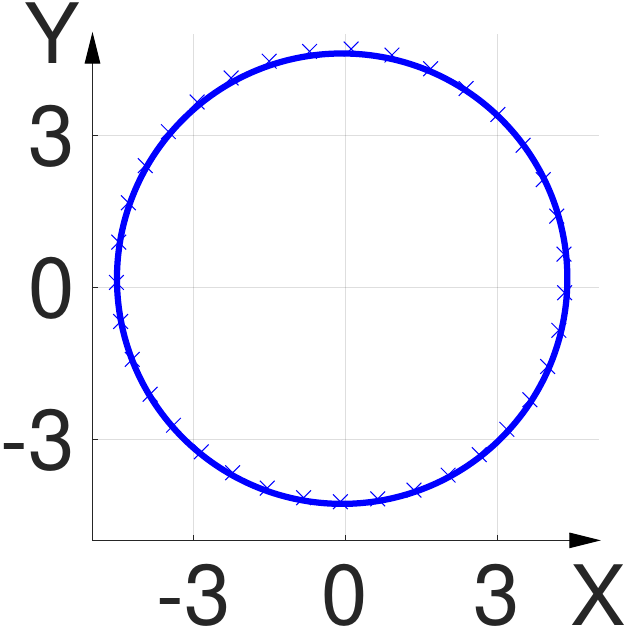}\label{fig_PlasticBox_c}}
    \subfloat[]{\includegraphics[width=0.55in]{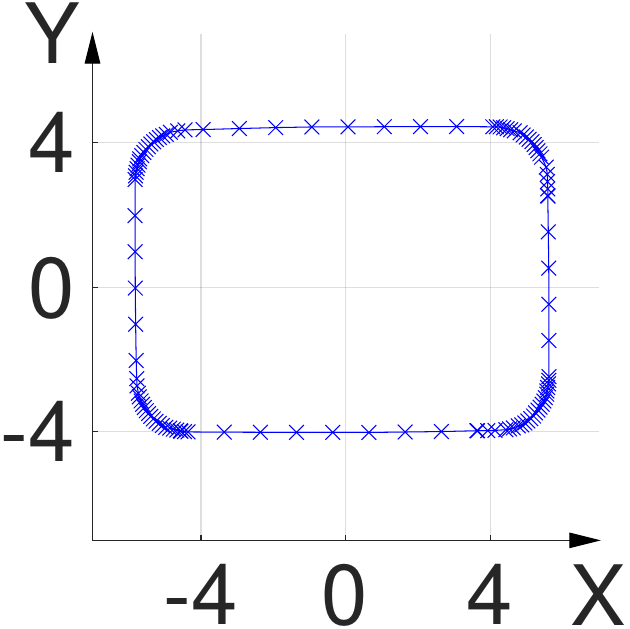}\label{fig_PlasticBox_c}}
    \subfloat[]{\includegraphics[width=0.55in]{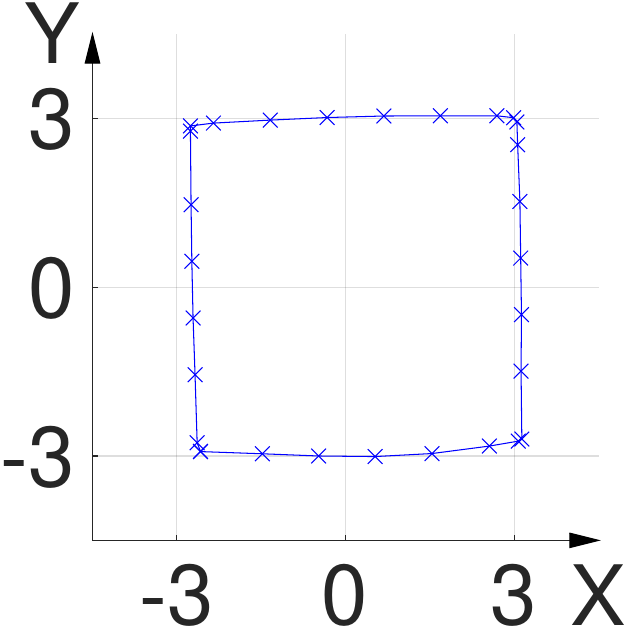}\label{fig_PaperBox_c}}
    \subfloat[]{\includegraphics[width=0.55in]{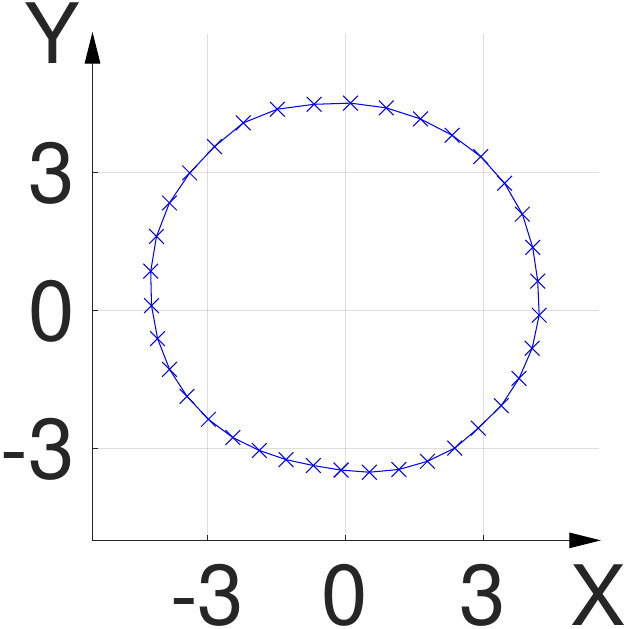}\label{fig_Apple_c}}
    \subfloat[]{\includegraphics[width=0.55in]{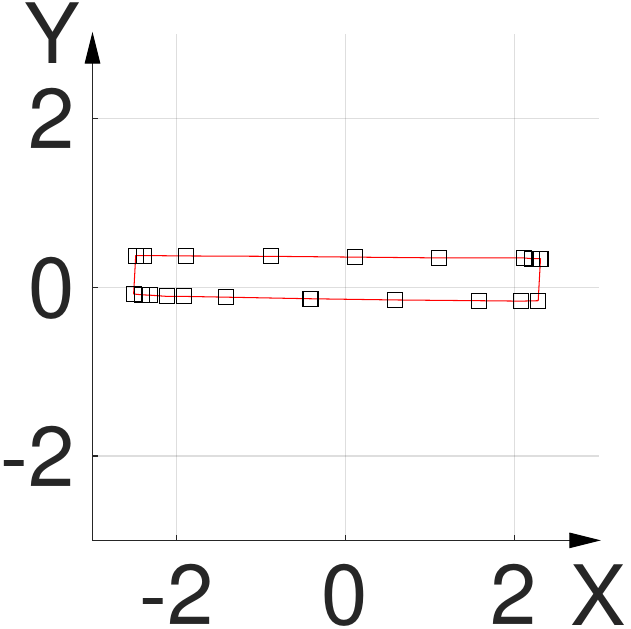}\label{fig_Foam_c}}
    \caption{Photographs and the reconstructed contours of the six common household objects. The first row shows objects with scanning paths. The second row illustrates their reconstructed contours. The unit of the $X$-$Y$ axes is the centimeter. The symbols `$\square$' and '$\times$' represent the points scanned with OA and US, respectively. The red and blue lines represent the OA and US connected lines between adjacent points, respectively.} \label{fig:obj3}
\end{figure}

Photos of six objects (within 27 × 27 × 43 cm$^3$) and their reconstructed contours are shown in Fig. \ref{fig:obj3}. The scanning path is drawn as the arrow curve in each photo. For contours, the OA and US scanned points are marked by ‘$\Box$’ and ’×’, respectively, while the OA and US modalities are indicated by the red and blue lines between adjacent points, respectively.

Among the six objects in Fig. \ref{fig:obj3}, only the steel bottle has responses in both modalities, the others only respond to US or OA scans alone. More specifically,
it is not possible to detect OA signals from glass bottles, plastic boxes, paper boxes, or apples primarily because of their high level of optical transmittance or low optical absorptivity. It is difficult to detect US signals from the black foam because of its porous structure, which results in weak acoustic reflectivity. For most cases, at least one-modality signals can be detected and used to reconstruct the contours of objects, again showing that our PDM$^2$ sensor has good adaptivity for a variety of objects.

\section{Conclusion and Futurework}\label{sec:conlusion}
In this paper, we reported a new PDM$^2$ sensor and algorithm design for object ranging and material \& interior structure detection to provide perception assistance for robotic grasping. Dual-modality ranging and material/structure sensing capabilities have been achieved using both US and OA modalities. We designed signal processing algorithms to automatically recognize ToFs in both modalities and developed ranging rectification alorithm to more nonlinearity. We also designed a sensor and scanning system calibration algorithm to help deploy the new sensor. To verify our design, a prototype PDM$^2$ sensor and an object scanning system have been constructed. We tested the new sensor with both common household items and OACTs and our sensor and algorithm design achieved satisfactory ranging and detection performances.  In conclusion, the new PDM$^2$ sensor provides a practical and powerful perception solution to assist robotic grasping of unknown objects. 

In the future, we plan to further optimize the sensor and algorithm design to improve the resistance to environmental noise and simplify installation and calibration. New results will be reported in future publications.

\section*{}
{\small
\section*{Acknowledgment}
The authors would like to thank Ken Goldberg, Xiaoyu Duan, Shuangyu Xie, and Aaron Kingery for their input and feedbacks.

\bibliographystyle{IEEEtran}
\bibliography{bib/TASE}
}



\end{document}